\documentclass{article} 
\usepackage{arxiv-iclr,times}

\usepackage[hidelinks]{hyperref}
\usepackage{url}

\usepackage{booktabs}       
\usepackage{xcolor}         
\usepackage{multirow}
\usepackage{enumitem}
\usepackage{extarrows}
\usepackage{algorithm}
\usepackage{algpseudocode}
\usepackage{graphicx}
\usepackage{subfigure}
\usepackage{caption}
\usepackage{wrapfig}
\usepackage{nicefrac}
\usepackage{makecell}

\usepackage{amsmath}
\usepackage{amsfonts}
\usepackage{amsthm}
\usepackage{amssymb}
\usepackage{mathtools}

\theoremstyle{definition}

\newcommand{\E}[2]{\mathbb{E}_{#1}\!\left[#2\right]}
\newcommand{\Egiven}[3]{\mathbb{E}_{#1}\!\left[\left.\kern-\nulldelimiterspace#2\,\right|\,#3\right]}
\newcommand{\var}[2]{\mathrm{Var}_{#1}\!\left[#2\right]}
\newcommand{\prob}[1]{\mathbb{P}\!\left(#1\right)}

\newcommand{\states}{\mathcal{S}}
\newcommand{\actions}{\mathcal{A}}
\newcommand{\state}[1]{\mathbf{s}_{#1}}
\newcommand{\action}[1]{\mathbf{a}_{#1}}
\newcommand{\mstate}[1]{\hat{\mathbf{s}}_{#1}}
\newcommand{\maction}[1]{\hat{\mathbf{a}}_{#1}}

\newcommand{\dynamicsfunction}{T}
\newcommand{\dynamics}[3]{\dynamicsfunction\!\left(\left.\kern-\nulldelimiterspace#3 \,\right|\, #1, #2\right)}

\newcommand{\qpi}[2]{Q^{#1}\!\left(#2\right)}

\newcommand{\qmodel}[2]{\hat{Q}^{#1}\!\left(#2\right)}
\newcommand{\qm}[1]{\hat{Q}^{#1}}
\newcommand{\Rhat}[1]{\hat{R}_{#1}}
\newcommand{\rhat}[1]{\hat{r}_{#1}}

\usepackage{pgf}
\usepackage{pgfplots}
\pgfplotsset{compat=1.17}
\usepackage{tikz}
\usepackage{amsmath}

\usetikzlibrary{arrows,shapes,shapes.geometric,automata,petri,positioning,calc}

\pgfmathdeclarefunction{pdf}{2}{%
  \pgfmathparse{1/(#2*sqrt(2*pi))*exp(-((x-#1)^2)/(2*#2^2))}%
}

\tikzset{
    observe/.style={
        circle,
        thick,
        draw=black, fill=none,
        minimum size=10mm,
        scale=1.3
    },
    state/.style={
        regular polygon, regular polygon sides=4,
        circle,
        thick,
        draw=black, fill=none,
        minimum size=9mm,
        scale=1.3
    },
    reward/.style={
        regular polygon, regular polygon sides=4,
        circle,
        thick,
        draw=black, fill=none,
        minimum size=9mm,
        scale=1.3
    },
    action/.style={
        thick,
        minimum size=9mm,
        scale=1.3
    },
    forward/.style={
        very thick,
        draw=black
    },
    adding/.style={
        very thick,
        dashed
    },
    sampling/.style={
        very thick,
        loosely dashdotted
    },
}

\pgfmathdeclarefunction{gauss}{2}{%
  \pgfmathparse{1/(#2*sqrt(2*pi))*exp(-((x-#1)^2)/(2*#2^2))}%
}

\newcommand{\acronym}{CBOP}
\newcommand{\eh}{\mathbb{E}[h]}
 
\newcommand\Tstrut{\rule{0pt}{2.6ex}}         

\title{Conservative Bayesian Model-Based Value Expansion for Offline Policy Optimization}

\author{
    Jihwan Jeong\textsuperscript{\rm 1,3}\thanks{Equal contribution},
    Xiaoyu Wang\textsuperscript{\rm 1}\footnotemark[1],
    Michael Gimelfarb\textsuperscript{\rm 1,3},
    Hyunwoo Kim\textsuperscript{\rm 2}\thanks{Corresponding authors},
    Baher Abdulhai\textsuperscript{\rm 1} \\
    \textbf{\& Scott Sanner}\textsuperscript{\rm 1,3}\footnotemark[2]\\
    \textsuperscript{\rm 1}University of Toronto,
    \textsuperscript{\rm 2}LG AI Research,
    \textsuperscript{\rm 3}Vector Institute \\
    \texttt{\{jihwan.jeong,cnxiaoyu.wang,mike.gimelfarb\}@mail.utoronto.ca,} \\
    \texttt{hwkim@lgresearch.ai, baher.abdulhai@utoronto.ca,} \\ \texttt{ssanner@mie.utoronto.ca}
}

\iclrfinalcopy 
\begin{document}

\maketitle

\begin{abstract}
Offline reinforcement learning (RL) addresses the problem of learning a performant policy from a fixed batch of data collected by following some behavior policy. Model-based approaches are particularly appealing in the offline setting since they can extract more learning signals from the logged dataset by learning a model of the environment. However, the performance of existing model-based approaches falls short of model-free counterparts, due to the compounding of estimation errors in the learned model. Driven by this observation, we argue that it is critical for a model-based method to understand when to trust the model and when to rely on model-free estimates, and how to act conservatively w.r.t. both. To this end, we derive an elegant and simple methodology called conservative Bayesian model-based value expansion for offline policy optimization (\acronym), that trades off model-free and model-based estimates during the policy evaluation step according to their epistemic uncertainties, and facilitates conservatism by taking a lower bound on the Bayesian posterior value estimate. On the standard D4RL continuous control tasks, we find that our method significantly outperforms previous model-based approaches: e.g., MOPO by $116.4$\%, MOReL by $23.2$\% and COMBO by $23.7$\%. Further, \acronym~achieves state-of-the-art performance on $11$ out of $18$ benchmark datasets while doing on par on the remaining datasets.
\end{abstract}

\section{Introduction} \label{sec:introduction}

Fueled by recent advances in supervised and unsupervised learning, there has been a great surge of interest in data-driven approaches to reinforcement learning (RL), known as \emph{offline RL} \citep{levine2020offline}. In offline RL, an RL agent must learn a good policy entirely from a logged dataset of past interactions, without access to the real environment. This paradigm of learning is particularly useful in applications where it is prohibited or too costly to conduct online trial-and-error explorations (e.g., due to safety concerns), such as autonomous driving \citep{yu2018auto}, 
robotics \citep{kalashnikov18qtopt}, and operations research \citep{boute2021inventory}.

However, because of the absence of online interactions with the environment that give correcting signals to the learner, direct applications of \emph{online} off-policy algorithms have been shown to fail in the \emph{offline} setting \citep{fujimoto2019bcq,kumar2019bear,wu2019brac,kumar2020cql}. This is mainly ascribed to the distribution shift between the learned policy and the \emph{behavior policy} (data-logging policy) during training.  For example, in $Q$-learning based algorithms, the distribution shift in the policy can incur uncontrolled overestimation bias in the learned value function. Specifically, positive biases in the $Q$ function for out-of-distribution (OOD) actions can be picked up during policy maximization, which leads to further deviation of the learned policy from the behavior policy, resulting in a vicious cycle of value overestimation. Hence, the design of offline RL algorithms revolves around how to counter the adverse impacts of the distribution shift while achieving improvements over the data-logging policy.

In this work, we consider model-based (MB) approaches since they allow better use of a given dataset and can provide better generalization capability \citep{yu2020mopo,kidambi2020morel,yu2021combo,argenson2021mbop}.
Typically, MB algorithms --- e.g., MOPO \citep{yu2020mopo}, MOReL \citep{kidambi2020morel}, and COMBO \citep{yu2021combo} --- adopt the Dyna-style policy optimization approach developed in online RL \citep{janner2019mbpo,sutton1990dyna}. That is, they use the learned dynamics model to generate rollouts, which are then combined with the real dataset for policy optimization. 

\begin{figure}[t!]
    \captionsetup{font={small}}
    \centering
    \subfigure{ \label{fig:intro-a}
    \includegraphics[width=0.3\textwidth]{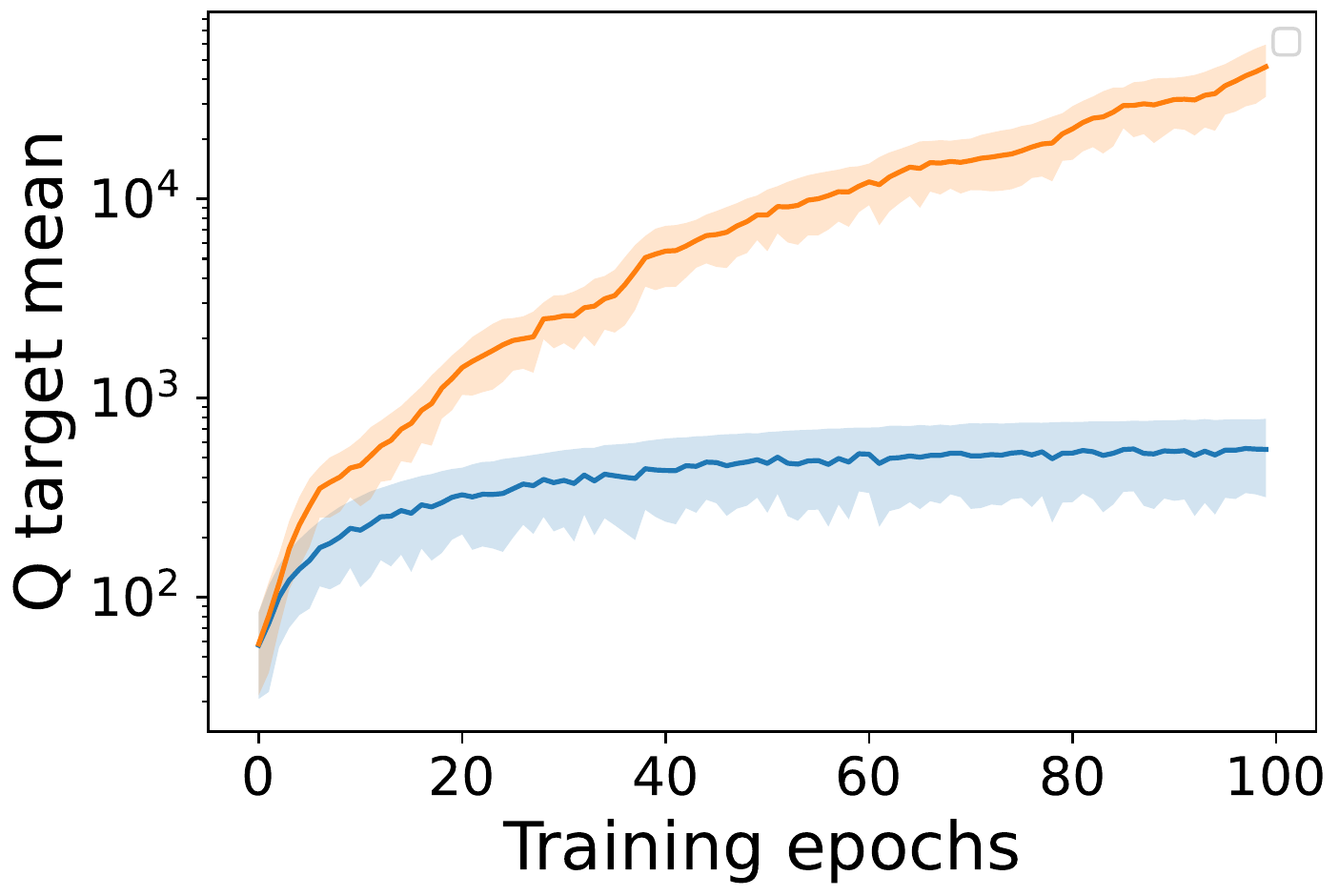}
    }
    \hspace{5pt}
    \subfigure{\label{fig:intro-b}  \includegraphics[width=0.3\textwidth]{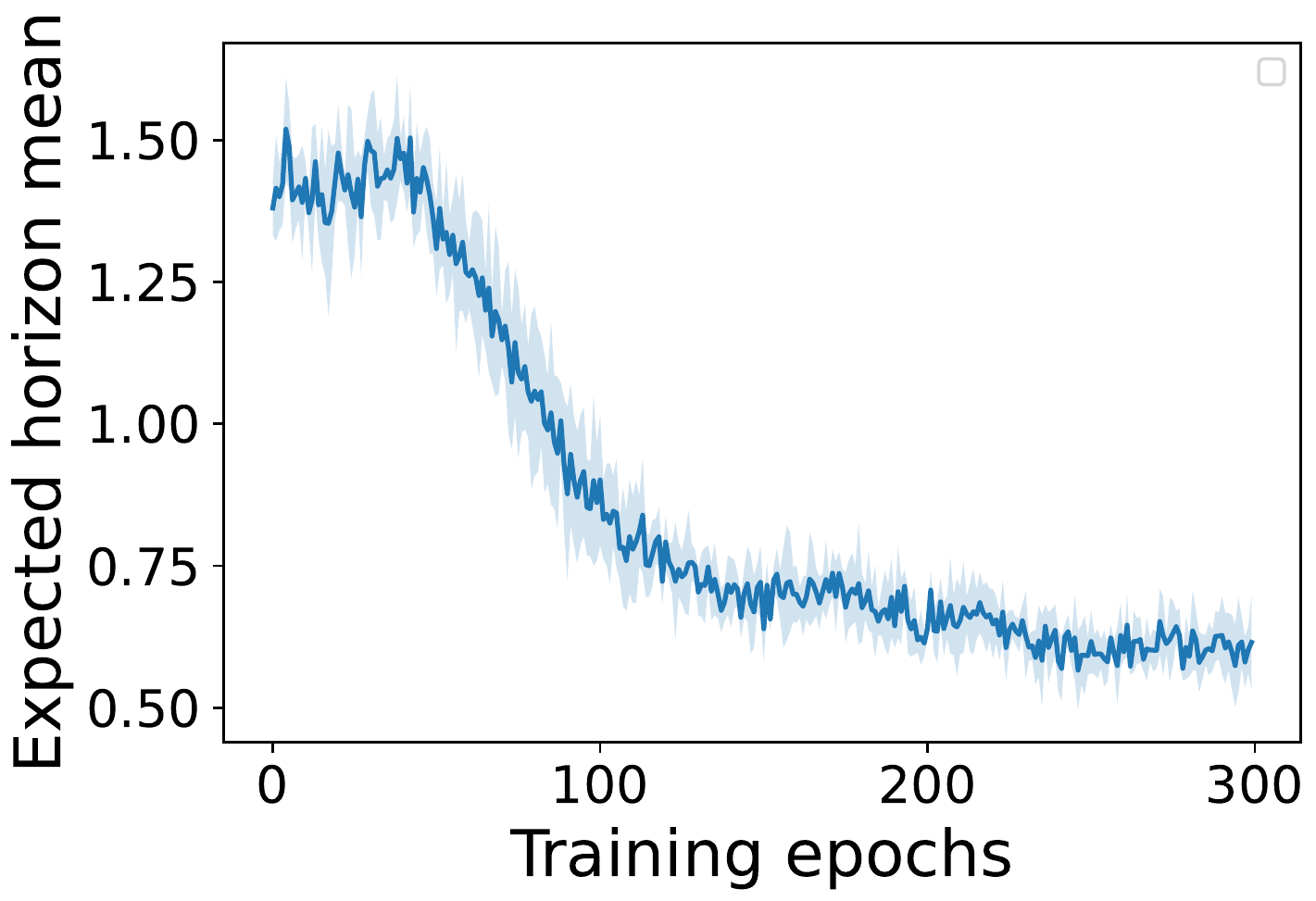}}
    \vspace{-0pt}
    \caption{
    Prevention of value overestimation \& adaptive reliance on model-based value predictions. 
    (\textit{Left}) We leverage the full posterior over the target values to prevent value overestimation during offline policy learning (blue). Without conservatism incorporated, the target value diverges (orange).  (\textit{Right}) We can automatically adjust the level of reliance on the model-based and bootstrapped model-free value predictions based on their respective uncertainty during model-based value expansion. The `\textit{expected horizon}' ($\mathbb{E}[h]=\sum_h w_h\cdot h$, $\sum_h w_h=1$) shows an effective model-based rollout horizon during policy optimization. $\mathbb{E}[h]$ is large at the beginning, but it gradually decreases as the model-free value estimates improve over time. The figures were generated using the \textit{hopper-random} dataset from D4RL \protect\citep{fu2020d4rl}.
    }
    \vspace{-0pt}
    \label{fig:intro-fig}
\end{figure}


We hypothesize that we can make better use of the learned model by employing it for target value estimation during the policy evaluation step of the actor-critic method. 
Specifically, we can compute $h$-step TD targets through dynamics model rollouts and bootstrapped terminal $Q$ function values. 
In online RL, this MB value expansion (MVE) has been shown to provide a better value estimation of a given state \citep{feinberg2018mve}. However, the naïve application of MVE does not work in the offline setting due to model bias that can be exploited during policy learning.

Therefore, it is critical to trust the model only when it can reliably predict the future, which can be captured by the epistemic uncertainty surrounding the model predictions. To this end, 
we propose \acronym~(\textbf{C}onservative \textbf{B}ayesian MVE for \textbf{O}ffline \textbf{P}olicy Optimization) to control the reliance on the model-based and model-free value estimates according to their respective uncertainties, while mitigating the overestimation errors in the learned values. 
Unlike existing MVE approaches (e.g., \citet{buckman2018steve}), \acronym~estimates the \emph{full} posterior distribution over a target value from the $h$-step TD targets for $h=0,\dots,H$ sampled from ensembles of the state dynamics and the $Q$ function. The novelty of \acronym~lies in its ability to fully leverage this uncertainty in two related ways: (1) by deriving an adaptive weighting over different $h$-step targets informed by the posterior uncertainty; and (2) by using this weighting to derive \emph{conservative} lower confidence bounds (LCB) on the target values that mitigates value overestimation. Ultimately, this allows \acronym~to reap the benefits of MVE while significantly reducing value overestimation in the offline setting (Figure \ref{fig:intro-fig}).

We evaluate \acronym~on the D4RL benchmark of continuous control tasks \citep{fu2020d4rl}. 
The experiments show that using the conservative target value estimate significantly outperforms previous model-based approaches: e.g., MOPO by $116.4$\%, MOReL by $23.2$\% and COMBO by $23.7$\%. Further, \acronym~achieves state-of-the-art performance on $11$ out of $18$ benchmark datasets while doing on par on the remaining datasets.

\section{Background} \label{sec:background}

We study RL in the framework of \emph{Markov decision processes} (MDPs) that are characterized by a tuple $(\states,\actions,T,r,d_0,\gamma)$; here, $\mathcal{S}$ is the state space, $\mathcal{A}$ is the action space, $T\left(\mathbf{s}'|\mathbf{s},\mathbf{a}\right)$ is the transition function, $r(\mathbf{s},\mathbf{a})$ is the immediate reward function, $d_0$ is the initial state distribution, and $\gamma\in[0, 1]$ is the discount factor.
Specifically, we call the transition and reward functions the \emph{model} of the environment, which we denote as $f = (T, r)$. 
A \emph{policy} $\pi$ is a mapping from $\states$ to $\actions$, and the goal of RL is to find an optimal policy $\pi^*$ which maximizes the expected cumulative discounted reward, $\E{\state{t}, \action{t}}{\sum_{t=0}^{\infty}\gamma^t r(\state{t}, \action{t})}$, where $\state{0}\sim d_0, \state{t}\sim T(\cdot|\state{t-1}, \action{t-1})$, and $\action{t}\sim \pi^*(\cdot|\state{t})$.
Often, we summarize the quality of a policy $\pi$ by the state-action value function $\qpi{\pi}{\mathbf{s},\mathbf{a}}:=\E{\state{t},\action{t}}{\sum_{t=0}^{\infty}\gamma^t r(\state{t}, \action{t})|\state{0}=\state{}, \action{0}=\action{}}$, where $\action{t} \sim \pi(\cdot|\state{t})~\forall t>0$.

Off-policy actor-critic methods, such as SAC \citep{haarnoja2018sac} and TD3 \citep{fujimoto2018td3}, have enjoyed great successes in complex continuous control tasks in deep RL, where parameterized neural networks for the policy $\pi_\theta$ (known as actor) and the action value function $Q_\phi$ (known as critic) are maintained.
Following the framework of the \emph{generalized policy iteration} (GPI) \citep{rl-sutton}, we understand the actor-critic algorithm as iterating between (i) policy evaluation and (ii) policy improvement. Here, policy evaluation typically refers to the calculation of $Q_\phi(\state{}, \pi_\theta(\state{}))$ for the policy $\pi_\theta$, while the improvement step is often as simple as maximizing the currently evaluated $Q_\phi$; i.e., $\max_\theta \E{\state{}\sim \mathcal{D}}{Q_\phi(\state{}, \pi_\theta(\state{}))}$ \citep{fujimoto2018td3}. 

\paragraph{Policy Evaluation}
At each iteration of policy learning, we evaluate the current policy $\pi_\theta$ by minimizing the \emph{mean squared Bellman error} (MSBE) with the dataset $\mathcal{D}$ of previous state transitions:
\begin{align}
    \mathcal{L}(\phi, \mathcal{D}) = \text{MSBE} :&= \E{(\state{},\action{},r,\state{}')\sim\mathcal{D}}{\left(y(\state{},\action{}, \state{}') - Q_\phi(\state{}, \action{})\right)^2}, \label{eq:msbe}\\
    y(\state{},\action{},\state{}') &= r(\state{}, \action{}) + \gamma Q_{\phi'}(\state{}', \action{}'),~~\action{}'\sim\pi_\theta(\cdot|\state{}') \label{eq:target}
\end{align}
where $y(\state{},\action{}, \state{}')$ is the \textit{TD target} at each $(\state{},\action{})$, towards which $Q_\phi$ is regressed. A separate \textit{target} network $Q_{\phi'}$ is used in computing $y$ to stabilize learning \citep{mnih2015humanlevel}. Off-policy algorithms typically use some variations of \eqref{eq:target}, e.g., by introducing the clipped double-$Q$ trick \citep{fujimoto2018td3}, in which $\min_{j=1,2}Q_{\phi'_j}(\state{}',\action{}')$ is used instead of $Q_{\phi'}(\state{}',\action{}')$ to prevent value overestimation. 

\paragraph{Model-based Offline RL}
In the offline setting, we are given a fixed set of transitions, $\mathcal{D}$, collected by some \textit{behavior} policy $\pi_\beta$, and the aim is to learn a policy $\pi$ that is better than $\pi_\beta$.
In particular, offline \emph{model-based} (MB) approaches learn the model $\hat{f}=(\hat{T},\hat{r})$ of the environment using $\mathcal{D}$ to facilitate the learning of a good policy.
Typically, $\hat{f}$ is trained to maximize the log-likelihood of its predictions.
Though MB algorithms are often considered capable of better generalization than their \emph{model-free} (MF) counterparts by leveraging the learned model, it is risky to trust the model for OOD samples. Hence, MOPO \citep{yu2020mopo} and MOReL \citep{kidambi2020morel} construct and learn from a pessimistic MDP where the model uncertainty in the next state prediction is penalized in the reward.  Criticizing the difficulty of accurately computing well-calibrated model uncertainty, COMBO \citep{yu2021combo} extends CQL \citep{kumar2020cql} to the model-based regime by regularizing the value function on OOD samples generated via model rollouts. These methods follow the \textit{Dyna-style} policy learning where model rollouts are used to augment the offline dataset \citep{sutton1990dyna,janner2019mbpo}. 

\paragraph{Model-based Value Expansion (MVE) for Policy Optimization} 
%
An alternative to the aforementioned \textit{Dyna-style} approaches is MVE \citep{feinberg2018mve}, which is arguably better suited to seamlessly integrating the power of both MF and MB worlds. In a nutshell, MVE attempts to more accurately estimate the TD target in \eqref{eq:target} by leveraging a model of the environment, which can lead to more efficient policy iteration.
Specifically, we can use the $h$-step MVE target $\Rhat{h}(\state{},\action{},\state{}')$ for $y(\state{},\action{},\state{}')$:
\begin{align}
    \hat{y}(\state{},\action{},\state{}')&=\Rhat{h}(\state{},\action{},\state{}') := \sum_{t=0}^{h}\gamma^t \rhat{t}(\mstate{t}, \maction{t}) + \gamma^{h+1} Q_{\phi'}(\mstate{h+1}, \maction{h+1}), \label{eq:h-return}\\
    (\mstate{0},\maction{0}, \rhat{0},\mstate{1})&=(\state{},\action{}, r,\state{}'),~\mstate{t}\sim \hat{T}(\cdot|\mstate{t-1},\maction{t-1}),~\maction{t}\sim\pi_\theta(\cdot|\mstate{t}),~1\le t \le h + 1,\nonumber
\end{align}
where $\Rhat{h}(\state{},\action{},\state{}')$ is obtained by the $h$-step MB return plus the terminal value at $h+1$ ($h=0$ reduces back to MF).
In reality, errors in the learned model $\hat{f}$ compound if rolled out for a large $h$. Thus, it is standard to set $h$ to a small number.

\section{Conservative Bayesian MVE for Offline Policy Optimization}\label{s:methodology}

The major limitations of MVE when applied to offline RL are as follows:

\begin{enumerate}
    \item The model predictions $\mstate{t}$ and $\rhat{t}$ in \eqref{eq:h-return} become increasingly less accurate as $t$ increases because model errors can compound, leading to largely biased target values. This issue is exacerbated in the offline setup because we cannot obtain additional 
    experiences to reduce the model error.
    \item The most common sidestep to avoid the issue above is to use short-horizon rollouts only. However, rolling out the model for only a short horizon even when the model can be trusted could severely restrict the benefit of being model-based.
    \item Finally, when the model rollouts go outside the support of $\mathcal{D}$, $\Rhat{h}$ in \eqref{eq:h-return} can have a large overestimation bias, which will eventually be propagated into the learned $Q_\phi$ function.
\end{enumerate}


Ideally, we want to \textbf{control the reliance on the model $\hat{f}$ and the bootstrapped $Q_{\phi'}$ according to their respective epistemic uncertainty, while also preventing $Q_\phi$ from accumulating large overestimation errors}.  That is, when we can trust $\hat{f}$, we can safely roll out the model for more steps to get a better value estimation.  On the contrary, if the model is uncertain about the future it predicts, we should shorten the rollout horizon and bootstrap from $Q_{\phi'}$ early on.  
Indeed, Figure \ref{fig:intro-fig} (right) exemplifies that \acronym~relies much more on the MB rollouts at the beginning of training because the value function is just initialized. As $Q_{\phi'}$ becomes more accurate over time, \acronym~automatically reduces the weights assigned to longer MB rollouts.

Below, we present \acronym, a Bayesian take on achieving the aforementioned two goals: trading off the MF and MB value estimates based on their uncertainty while obtaining a conservative estimation of the target $\hat{y}(\state{},\action{},\state{}')$.
To this end, we first let $\qmodel{\pi}{\state{t}, \action{t}}$ denote the value of the policy $\pi$ at $(\state{t},\action{t})$ in the learned MDP defined by its dynamics $\hat{f}$; that is,
\begin{align}
    \qmodel{\pi}{\state{t}, \action{t}}=\E{\hat{f},\pi}{\sum_{k=0}^\infty \gamma^k \hat{r}(\mstate{t+k}, \maction{t+k})}, ~(\mstate{t},\maction{t})=(\state{t},\action{t}),~\maction{t+k}\sim\pi(\cdot|\mstate{t+k}). \label{eq:model-value}
\end{align}
Note that in the offline MBRL setting, we typically cannot learn $Q^\pi$ due to having only an approximation $\hat{f}$ of the model, and thus we focus instead on learning $\hat{Q}^\pi$. 

\begin{wrapfigure}[7]{r}{0.54\textwidth}
\vspace{-10mm}
\begin{minipage}{0.54\textwidth}
\begin{algorithm}[H]
\caption{Conservative Bayesian MVE} \label{alg:cbop-simple}
\begin{algorithmic}
    \State \textbf{Input:} $(\state{t},\action{t},r_t, \state{t+1}), \hat{f}, Q_{\phi'}$
    \State 1. Sample $\Rhat{h}~\forall h\le H$ using $\hat{f}$ and $Q_{\phi'}$ as in \eqref{eq:h-return}
    \State 2. Estimate $\mu_h,~\sigma_h$ according to \eqref{eq:total-expectation}, \eqref{eq:total-variance}
    \State 3. Compute the posterior $\mathcal{N}(\mu,\sigma)$ using \eqref{eq:posterior}
    \State \textbf{return} conservative value target (e.g., LCB $\mu-\psi \sigma$)
\end{algorithmic}
\end{algorithm}
\end{minipage}
\end{wrapfigure}

Although there exists a unique $\qmodel{\pi}{\state{},\action{}}$ at each $(\state{},\action{})$ given a fixed model $\hat{f}$, we cannot directly observe the value unless we infinitely roll out the model from $(\state{},\action{})$ until termination, which is computationally infeasible. Instead, we view each $\Rhat{h}$ $\forall h$ defined in \eqref{eq:h-return} as a conditionally independent (biased) noisy observation of the true underlying parameter $\qm{\pi}$.\footnote{We will omit $(\state{}, \action{}, \state{}')$ henceforth if it is clear from the context.}
From this assumption, we can construct the Bayesian posterior over $\qm{\pi}$ given the observations $\Rhat{h}$ $\forall h$.  
With the closed-form posterior distribution at hand, we can take various conservative estimates from the distribution; we use the lower confidence bound (LCB) in this work. Algorithm \ref{alg:cbop-simple} summarizes the procedure at a high-level. Please see Algorithm \ref{alg:cbop} in Appendix \ref{appendix:algo-summary} for the full description of \acronym.

\subsection{
Conservative Value Estimation via Bayesian Inference
} \label{sec:bayesian}

In this part, we formally discuss the conservative value estimation of \acronym~based on Bayesian posterior inference. Specifically, the parameter of interest is $\qm{\pi}$, and we seek its posterior estimation:
\begin{align}
    \prob{\qm{\pi}~|~\Rhat{0},\dots,\Rhat{H}} &\propto \prob{\Rhat{0},\dots,\Rhat{H}~|~\qm{\pi}}~\prob{\qm{\pi}}=\prob{\qm{\pi}} \prod_{h=0}^H \prob{\Rhat{h}~|~\qm{\pi}}, \label{eq:cond-indep-bayes}
\end{align}
where we assume that $\Rhat{h}~(h=0,\dots,H)$ are conditionally independent given $\qm{\pi}$ (see Appendix \ref{sec:appendix-assumptions} where we discuss in detail about the assumptions present in \acronym).

In this work, we model the likelihood of observations $\mathbb{P}(\Rhat{h}|\qm{\pi})$ as normally distributed with the mean $\mu_h$ and the standard deviation $\sigma_h$:
\begin{align}
    \Rhat{h}~\lvert~\qm{\pi~}\sim~ \mathcal{N}(\mu_h, \sigma_h^2), \label{eq:hreturn-gaussian}
\end{align}
since it leads to a closed-form posterior update. Furthermore, since $\hat{R}_h$ can be seen as a sum of future immediate rewards, when the MDP is ergodic and $\gamma$ is close to 1, the Gaussian assumption (approximately) holds according to the central limit theorem \citep{dearden1998bayesian}. Also, note that our Bayesian framework is not restricted to the Gaussian assumptions, and other surrogate probability distributions such as the Student-t distribution could be used instead.

For the prior, we use the improper (or uninformative) prior, $\mathbb{P}(\qm{\pi})=1$, since it is natural to assume that we lack generally applicable prior information over the target value across different environments and tasks \citep{christensen2011bayesian}.  The use of the improper prior is well justified in the Bayesian literature \citep{wasserman2010statistics,berger1985statistical}, and the particular prior we use in \acronym~corresponds to the Jeffreys prior, which has the invariant property under a change of coordinates.  
The Gaussian likelihood and the improper prior lead to a `proper' Gaussian posterior density that integrates to $1$, from which we can make various probabilistic inferences \citep{wasserman2010statistics}.

\begin{wrapfigure}[13]{r}{0.55\textwidth}
  \vspace{-3mm}
  \centering
  \includegraphics[width=\linewidth]{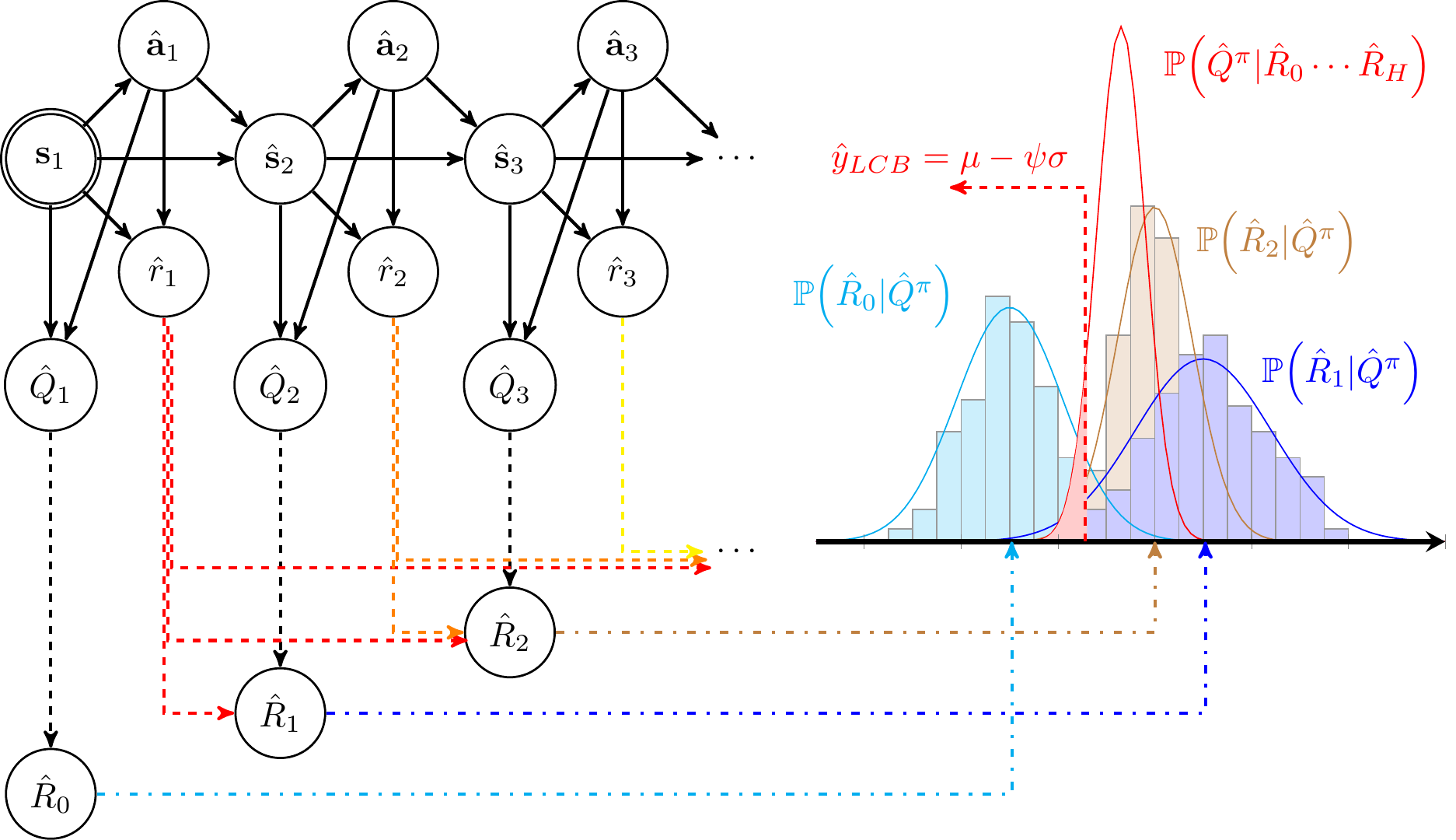}
  \vspace*{-6mm}
  \caption{The graphical model representation of \acronym} 
  \label{fig:pgm-mve}
\end{wrapfigure}
The posterior \eqref{eq:cond-indep-bayes} is a Gaussian with mean $\mu$ and variance $\sigma^2$, defined as follows:
{\small
\begin{align}
    \rho = \sum_{h=0}^{H} \rho_h,~~ \mu &= \sum_{h=0}^{H}\bigg(\frac{\rho_h}{\sum_{h=0}^{H}\rho_h}\bigg) \mu_h, \label{eq:posterior}
\end{align}
}%
where $\rho=\nicefrac{1}{\sigma^2}$ and $\rho_h=\nicefrac{1}{\sigma_h^2}$ are the precisions of the posterior and the likelihood of $\Rhat{h}$, respectively. 
The posterior mean $\mu$ corresponds to the MAP estimation of $\qm{\pi}$. Note that $\mu$ has the form of a weighted sum, $\sum_h w_h \mu_h$, with $w_h=\nicefrac{\rho_h}{\sum_{h=0}^{H}\rho_h}\in(0,1)$ being the weight allocated to $\Rhat{h}$.
If the variance of $\Rhat{h}$ for some $h$ is relatively large, we give a smaller weight to that observation.  If, on the other hand, $\Rhat{h}$ all have the same variance (e.g. $\rho_0 = \dots = \rho_H$), we recover the usual $H$-step return estimate.
Recall that the quality of $\Rhat{h}$ is determined by that of the model rollout return and the bootstrapped terminal value. Thus intuitively speaking, the adaptive weight $w_h$ given by the Bayesian posterior allows the trade-off between the epistemic uncertainty of the model with that of the $Q$ function.

Figure \ref{fig:pgm-mve} illustrates the overall posterior estimation procedure.  Given a transition tuple $(\state{}, \action{},r, \state{}')$, we start the model rollout from $\state{1}=\state{}'$.  At each rollout horizon $h$, the cumulative discounted reward $\sum_{t=0}^h \gamma^t \rhat{t}$ is sampled by the dynamics model and the terminal value $\qm{}_h$ is sampled by the $Q$ function (the sampling procedure is described in Section \ref{sec:ensemble-sampling}).  We then get $\Rhat{h}$ by adding the $h$-step MB return samples and the terminal values $\gamma^{h+1}\qm{}_{h+1}$, which we deem as sampled from the distribution $\mathbb{P}(\Rhat{h}|\qm{\pi})$ parameterized by $\mu_h$, $\sigma_h^2$ (we use the sample mean and variance).
These individual $h$-step \textit{observations} are then combined through the Bayesian inference to give us the posterior distribution over $\qm{\pi}$. 

It is worth noting that the MAP estimator can also be derived from the perspective of variance optimization \citep{buckman2018steve} over the target values.  However, we have provided much evidence in Section \ref{sec:experiments} and Appendix \ref{appendix:ablations} that the point estimate does not work in the offline setting due to value overestimation.  Hence, it is imperative that we should have the full posterior distribution over the target value, such that we can make a conservative estimation rather than the MAP estimation.

To further understand the impact of using the MAP estimator for the value estimation, consider an estimator $\tilde{Q}$ of $\qm{\pi}$ and its squared loss: $L(\qm{\pi}, \tilde{Q})=(\qm{\pi} - \tilde{Q})^2$. It is known that the posterior mean of $\qm{\pi}$ minimizes the \emph{Bayes risk} w.r.t. $L(\qm{\pi}, \tilde{Q})$ \citep{wasserman2010statistics}, meaning that the posterior risk $\int L(\qm{\pi}, \tilde{Q})\mathbb{P}(\qm{\pi}|\Rhat{0},\dots,\Rhat{H})d\qm{\pi}$ is minimized at $\tilde{Q}=\mu$. In this context, $\mu$ is also called the (generalized) \emph{Bayes estimator} of $\qm{\pi}$, which is an admissible estimator \citep{robert2007bayesian}. 
Despite seemingly advantageous, this result has a negative implication in offline RL. That is, the MAP estimator minimizes the squared loss from $\qmodel{\pi}{\state{},\action{}}$ over the entire support of the posterior, weighted by the posterior distribution. Now, the distribution shift of $\pi$ from $\pi_\beta$ can lead to significantly biased $\qm{\pi}$ compared to the true $Q^{\pi}$. In this case, the quality of the MAP estimator when evaluated in the real MDP would be poor. Especially, the overestimation bias in the MAP estimation can quickly propagate to the $Q_\phi$ function and thereby exacerbate the distribution shift.

\subsection{Ensembles of Dynamics and Q Functions for Sampling h-Step MVE Targets} \label{sec:ensemble-sampling}

In this section, we discuss how we estimate the parameters $\mu_h, \sigma^2_h$ of $\mathbb{P}(\Rhat{h}|\qm{\pi})$ from the ensemble of dynamics models and that of $Q$ functions.  

Assume we have a bootstrapped dynamics ensemble model $\hat{f}$ consisting of $K$ different models ($\hat{f}_1,\dots,\hat{f}_K$) trained with different sequences of mini-batches of $\mathcal{D}$ \citep{chua2018pets,janner2019mbpo}. Similarly, we assume a $Q$ ensemble of size $M$.  Given a state $\mstate{t}$ and an action $\maction{t}$, we can construct the probability over the next state $\mstate{t+1}$ and reward $\rhat{t}$ by the ensemble as follows:
\begin{equation*}
    \prob{\mstate{t+1}, \rhat{t}|\mstate{t}, \action{t}} = \sum_{k=1}^K \prob{\hat{f}_k} \cdot \prob{\mstate{t+1},\rhat{t}|\mstate{t}, \action{t}, \hat{f}_k}
\end{equation*}
where $\mathbb{P}(\hat{f}_k)$ is the probability of selecting the $k$th model from the ensemble, which is $1/K$ when all models are weighted equally. Now, the sampling method that exactly follows the probabilistic graphical model shown in Figure \ref{fig:pgm-mve} would first sample a model from the ensemble at each time step, followed by sampling the next state transition (and reward) from the model, which should then be repeated $K$ times per state to generate a single sample.  Then, we evaluate the resulting state $\mstate{t+1}$ and action $\maction{t+1}\sim\pi_\theta(\mstate{t+1})$ with the $Q$ ensemble to obtain $M$ samples. To obtain $N$ trajectories from a single initial state to estimate $\mu_h$ and $\sigma^2_h$ for $h=1,\dots,H$, the overall procedure requires $\mathcal{O}(NKH)$ computation, which can quickly become infeasible for moderately large $K$ and $N$ values.

To reduce the computational complexity, we follow \citet{chua2018pets} where each \textit{particle} is propagated by a single model of the ensemble for $H$ steps.  With this, we can obtain $N$ trajectories of length $H$ from one state with $\mathcal{O}(NH)$ instead of $\mathcal{O}(NKH)$ (below we use $N=K$, i.e., we generate one particle per model).  
Concretely, given a single transition $\tau=(\state{0}, \action{0},r_0, \state{1})$, we create $K$ numbers of \textit{particles} by replicating $\state{1}$ $K$ times, denoted as $\mstate{1}^{(k)}\:\forall k$. 
The $k$th particle is propagated by a fixed model $\hat{f}_k$ and the policy $\pi_\theta$ for $H$ steps, where $(\mstate{t}^{(k)}, \rhat{t-1}^{(k)})= \hat{f}_k(\mstate{t-1}^{(k)},\maction{t-1}^{(k)})$ and $\maction{t}^{(k)}\sim \pi_\theta(\mstate{t}^{(k)})$. At each imagined timestep $t\in[0,H+1]$, $M$ number of terminal values are sampled by the $Q_{\phi'}$ ensemble at $(\mstate{t}^{(k)},\maction{t}^{(k)})$.

Despite the computational benefit, an implication of this sampling method is that it no longer directly follows the graphical model representation in Figure \ref{fig:pgm-mve}.  
However, we can still correctly estimate $\mu_h$ and $\sigma_h^2$ by turning to the law of total expectation and the law of total variance. That is, 
\begin{align}
    \mu_h = \Egiven{\pi_\theta}{\Rhat{h}}{\tau} =&\: \E{\hat{f}_k}{\Egiven{\pi_\theta}{\Rhat{h}}{\tau, \hat{f}_k}} \label{eq:total-expectation}
\end{align}
where the outer expectation is w.r.t. the dynamics ensemble sampling probability $\mathbb{P}(\hat{f}_k)=\nicefrac{1}{K}$.
Hence, given a fixed dynamics model $\hat{f}_k$, we sample $\Rhat{h}$ by following $\pi_\theta$ and compute the average of the $h$-step return, which is then averaged across different ensemble models. In fact, the resulting $\mu_h$ is the mean of all aggregated $M\times K$ samples of $\Rhat{h}$.

The $h$-step return variance $\mathrm{Var}_{\pi_\theta}(\Rhat{h}|\tau)$ decomposes via the law of total variance as following:
\begin{equation}
    \sigma_h^2 = \var{\pi_\theta}{\Rhat{h}| \tau} = \underbrace{\E{\hat{f}_k}{\var{\pi_\theta}{\Rhat{h}| \tau, \hat{f}_k}}}_{A} + \underbrace{\var{\hat{f}_k}{\Egiven{\pi_\theta}{\Rhat{h}}{\tau, \hat{f}_k}}}_{B}. \label{eq:total-variance}
\end{equation}
Here, $A$ is related to the epistemic uncertainty of the $Q_{\phi'}$ ensemble; while $B$ is associated with the epistemic uncertainty of the dynamics ensemble. The total variance $\mathrm{Var}_{\pi_\theta}(\Rhat{h}|\tau)$ captures both uncertainties.  This way, even though we use a different sampling scheme than presented in the graphical model of Figure \ref{fig:pgm-mve}, we can compute the unbiased estimators of the Gaussian parameters. 

Once we obtain $\mu_h$ and $\sigma_h^2$, we plug them into \eqref{eq:posterior} to compute the posterior mean and the variance. A conservative value estimation can be made by $\hat{y}_{LCB} = \mu - \psi \sigma$ with some coefficient $\psi>0$ \citep{jin2021pessimism,rashidinejad2021lcb}. Under the Gaussian assumption, this corresponds to the worst-case return estimate in a Bayesian credible interval for $\hat{Q}^\pi$. We summarize \acronym~in Algorithm \ref{alg:cbop} in Appendix \ref{appendix:algo-summary}.

\section{Experiments} \label{sec:experiments}
\begin{table*}[t!]
\captionsetup{font=footnotesize}
\tiny
\caption{Normalized scores on D4RL MuJoCo Gym environments. Experiments ran with $5$ seeds.}
\begin{center}
\begin{tabular}{cccccccccc}
\toprule
 & & MOPO & MOReL & COMBO & CQL & TD3+BC & EDAC & IQL & \acronym \\
\midrule
\midrule
\multirow{3}{*}{\rotatebox[origin=c]{90}{random}} & 
   halfcheetah  & $35.4\pm 2.5$ & $25.6$ & $\mathbf{38.8}$   & $35.4$ & $10.2\pm 1.3$ & $28.4\pm 1.0$          & -                & $32.8\pm 0.4$ \\
 & hopper       & $11.7\pm0.4$  & $\mathbf{53.6}$ & $17.9$            & $10.8$ & $11.0\pm 0.1$ & $31.3\pm 0.0$ & - & $31.4\pm 0.0$\\
 & walker2d     & $13.6\pm 2.6$ & $37.3$ & $7.0$             & $7.0$  & $1.4\pm 1.6$  & $\mathbf{21.7\pm 0.0}$ & - & $17.8\pm 0.4$ \\
\midrule
\multirow{3}{*}{\rotatebox[origin=c]{90}{medium}} & 
   halfcheetah  & $42.3\pm 1.6$  & $42.1$ & $54.2$   & $44.4$ & $42.8\pm 0.3$ & $67.5\pm 1.2$ & $47.4$ & $\mathbf{74.3\pm 0.2}$\\
 & hopper       & $28.0\pm 12.4$ & $95.4$ & $94.9$   & $79.2$ & $99.5\pm 1.0$ & $101.6\pm 0.6$& $66.2$ & $\mathbf{102.6\pm 0.1}$\\
 & walker2d     & $17.8\pm 19.3$ & $77.8$ & $75.5$   & $58.0$ & $79.7\pm 1.8$ & $92.5\pm 0.8$ & $78.3$ & $\mathbf{95.5\pm 0.4}$ \\
\midrule
\multirow{3}{*}{\rotatebox[origin=c]{90}{medium} \rotatebox[origin=c]{90}{replay}} & 
   halfcheetah  & $53.1\pm 2.0$  & $40.2$ & $55.1$   & $46.2$ & $43.3\pm 0.5$ & $63.9\pm 0.8$ & $44.2$ & $\mathbf{66.4\pm 0.3}$ \\
 & hopper       & $67.5\pm 24.7$ & $93.6$ & $73.1$   & $48.6$ & $31.4\pm 3.0$ & $101.8\pm 0.5$ & $94.7$ & $\mathbf{104.3\pm 0.4}$ \\
 & walker2d     & $39.0\pm 9.6$  & $49.8$ & $56.0$   & $26.7$ & $25.2\pm 5.1$ & $87.1\pm 2.3$ & $73.8$ & $\mathbf{92.7\pm 0.9}$\\
\midrule
\multirow{3}{*}{\rotatebox[origin=c]{90}{medium} \rotatebox[origin=c]{90}{expert}} & 
   halfcheetah  & $63.3\pm 38.0$ & $53.3$ & $90.0$   & $62.4$ & $97.9\pm 4.4$    & $\mathbf{107.1\pm 2.0}$ & $86.7$ & $105.4\pm 1.6$\\
 & hopper       & $23.7\pm 6.0$  & $108.7$& $111.1$  & $98.7$ & $\mathbf{112.2\pm 0.2}$   & $110.7\pm 0.1$ & $91.5$ & $111.6\pm 0.2$\\
 & walker2d     & $44.6\pm 12.9$ & $95.6$ & $96.1$   & $111.0$ & $101.1\pm 9.3$  & $114.7\pm 0.9$ & $109.6$ & $\mathbf{117.2\pm 0.5}$\\
\midrule
\multirow{3}{*}{\rotatebox[origin=c]{90}{expert}} & 
   halfcheetah  & - & - & - & -                       & $105.7\pm 1.9$ & $\mathbf{106.8\pm 3.4}$ & - & $100.4\pm 0.9$\\
 & hopper       & - & - & - & -                       & $\mathbf{112.2\pm 0.2}$ & $110.3\pm 0.3$ & - & $111.4\pm 0.2$ \\
 & walker2d     & - & - & - & -                       & $105.7\pm 2.7$ & $115.1\pm 1.9$ & - & $\mathbf{122.7\pm 0.8}$\\
\midrule
\multirow{3}{*}{\rotatebox[origin=c]{90}{full} \rotatebox[origin=c]{90}{replay}} & 
   halfcheetah  & - & - & - & - & - & $84.6\pm 0.9$  & - & $\mathbf{85.5\pm 0.3}$\\
 & hopper       & - & - & - & - & - & $105.4\pm 0.7$ & - & $\mathbf{108.1\pm 0.3}$\\
 & walker2d     & - & - & - & - & - & $99.8\pm 0.7$  & - & $\mathbf{107.8\pm 0.2}$\\
\bottomrule
\end{tabular}
\end{center}
\label{table:mujoco}
\end{table*}

We have designed the experiments to answer the following research questions: 
\textbf{(\hypertarget{rq1}{RQ1})} Is \acronym~able to adaptively determine the weights assigned to different $h$-step returns according to the relative uncertainty of the learned model and that of the $Q$ function?
\textbf{(\hypertarget{rq2}{RQ2})} How does \acronym~perform in the offline RL benchmark? 
\textbf{(\hypertarget{rq3}{RQ3})} Does \acronym~with LCB provide conservative target $Q$ estimation? 
\textbf{(\hypertarget{rq4}{RQ4})} How does having the full posterior over the target values  compare against using the MAP estimation in performance?
\textbf{(\hypertarget{rq5}{RQ5})} How much better is it to adaptively control the weights to $h$-step returns during training as opposed to using a fixed set of weights throughout training?

We evaluate these RQs on the standard \emph{D4RL} offline RL benchmark \citep{fu2020d4rl}.  In particular, we use the D4RL MuJoCo Gym dataset that contains three environments: \textit{halfcheetah}, \textit{hopper}, and \textit{walker2d}.  For each environment, there are six different behavior policy configurations: \textit{random} (\textit{r}), \textit{medium} (\textit{m}), \textit{medium-replay} (\textit{mr}), \textit{medium-expert} (\textit{me}), \textit{expert} (\textit{e}), and \textit{full-replay} (\textit{fr}). We release our code at \url{https://github.com/jihwan-jeong/CBOP}.

\subsection{\acronym~can Automatically Adjust Reliance on the Learned Model}
\label{subsec:expected-horizon}

\begin{wrapfigure}[14]{r}{0.45\textwidth}
    \captionsetup{font=small}
    \vspace{-1em}
    \centering
    \includegraphics[width=0.45\textwidth]{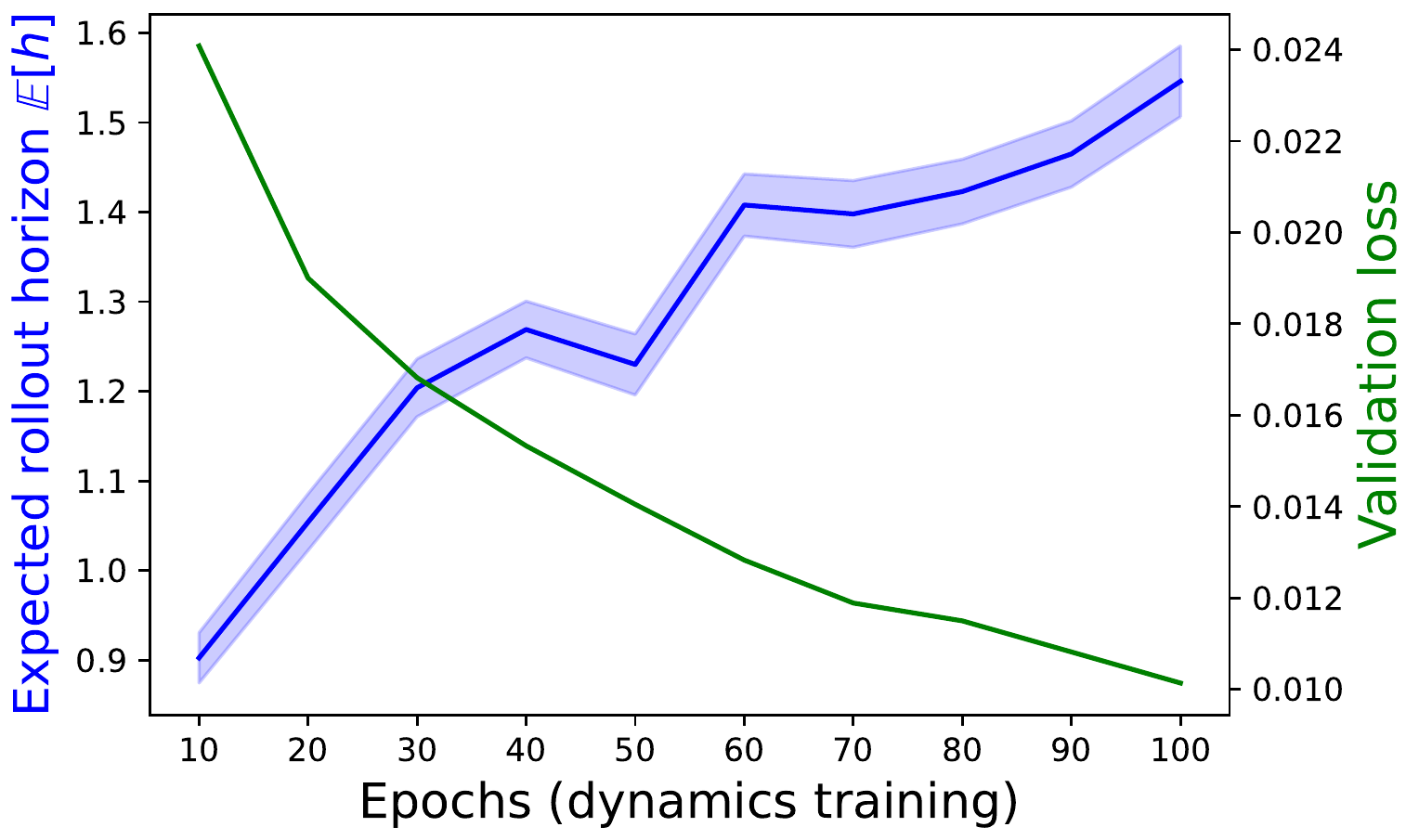}
    \caption{$\eh$ during \acronym~training with the dynamics model trained for different numbers of epochs. \acronym~can place larger weights to longer-horizon rollouts as the dynamics model becomes more accurate.}
  \label{fig:expected-horizon}
\end{wrapfigure}%

To investigate \hyperlink{rq1}{RQ1}, we use the notion of the \emph{expected rollout horizon}, which we define as $\mathbb{E}[h]=\sum_{h=0}^H w_h \cdot h$.
Here, $w_h$ is the weight given to the mean of $\Rhat{h}$ as defined in \eqref{eq:posterior}, which sums to $1$.  A larger $\eh$ indicates that more weights are assigned to longer-horizon model-based rollouts.

Figure \ref{fig:intro-fig} already shows that $\eh$ decreases as the $Q$ function becomes better over time.
On the other hand, Figure \ref{fig:expected-horizon} shows how the quality of the learned model affects $\eh$. 
Specifically, we trained the dynamics model on \textit{halfcheetah-m} for different numbers of epochs ($10,\dots,100$); then, we trained the policy with \acronym~for $150$ epochs. 

\subsection{Performance Comparison} \label{subsec:experiments-mujoco}

To investigate \hyperlink{rq2}{RQ2}, we select baselines covering both model-based and model-free approaches: (\emph{model-free}) \textbf{CQL} \citep{kumar2020cql}, \textbf{IQL} \citep{kostrikov2022iql}, \textbf{TD3+BC} \citep{fujimoto2021td3+bc}, \textbf{EDAC} \citep{an2021edac}; (\emph{model-based}) \textbf{MOPO} \citep{yu2020mopo}, \textbf{MOReL} \citep{kidambi2020morel}, and \textbf{COMBO} \citep{yu2021combo}.  Details of experiments are provided in Appendix \ref{appendix:experiment-settings}. 



Table \ref{table:mujoco} shows the experimental results. 
Comparing across all baselines, \acronym~presents new state-of-the-art performance in \textbf{\emph{11 tasks out of 18}} while performing similar in the remaining configurations.
Notably, \acronym~outperforms prior works in \textit{medium}, \textit{medium-replay}, and \textit{full-replay} configurations with large margins.  We maintain that these are the datasets of greater interest than, e.g., \textit{random} or \textit{expert} datasets because the learned policy needs to be much different than the behavior policy in order to perform well.  Furthermore, the improvement compared to previous model-based arts is substantial: \acronym~outperforms MOPO, MOReL, and COMBO by 
$116.4$\%, $23.2$\% and $23.7$\% (respectively) on average across four behavior policy configurations. 

\begin{wraptable}[9]{r}{0.45\textwidth}
\vspace{-1.2em}
\captionsetup{font=footnotesize}
\scriptsize
\caption{Difference between the values predicted by the learned $Q$ functions and the true discounted returns from the environment.}
\centering
\begin{tabular}{lcc c cc}
\toprule
      & \multicolumn{2}{c}{CQL} && \multicolumn{2}{c}{\acronym} \\
\cmidrule(lr{.75em}){2-3} \cmidrule(lr{.75em}){5-6}
    Task name & Mean & Max && Mean & Max \\
\midrule
    hopper-m     & -61.84 & -3.20 && -55.83 & -16.21 \\
    hopper-mr    & -142.89 & -28.73 && -172.45 & -39.45 \\
    hopper-me    & -79.67 & -5.16 && -114.39 & -11.24 \\
\bottomrule
\end{tabular}
\label{tab:pe}
\end{wraptable}

\subsection{\acronym~Learns Conservative Values} \label{subsec:conservatism}

To answer \hyperlink{rq3}{RQ3}, we have selected $3$ configurations (\textit{m}, \textit{me}, and \textit{mr}) from the \textit{hopper} environment and evaluated the value function at the states randomly sampled from the datasets, i.e., $\mathbb{E}_{\state{}\sim \mathcal{D}}[\hat{V}^\pi(\state{})]$ (nb. a similar analysis is given in CQL).
Then, we compared these estimates with the Monte Carlo estimations from the true environment by rolling out the learned policy until termination. 

Table \ref{tab:pe} reports how large are the value predictions compared to the true returns.  Notice that not only the mean predictions are negative but also the maximum values are, which affirms that \acronym~indeed has learned conservative value functions.  Despite the predictions by \acronym~being smaller than those of CQL in \textit{hopper-mr} and \textit{me}, we can see that \acronym~significantly outperforms CQL in these settings. See Appendix \ref{appendix:conservatism-analysis} for more details.
 

\subsection{Ablation Studies} \label{subsec:experiments-ablations}


\paragraph{LCB vs. MAP in the offline setting} To answer \hyperlink{rq4}{RQ4}, we compare \acronym~with STEVE \citep{buckman2018steve} which is equivalent to using the MAP estimation for target $Q$ predictions. 
Figure \ref{fig:intro-fig} (left) shows the case where the value function learned by STEVE blows up (orange).  Further, we include the performance of STEVE in all configurations in Appendix \ref{appendix:ablations}.  To summarize the results, STEVE fails to learn useful policies for $11$ out of $18$ tasks.  Especially, except for the \textit{fr} datasets, using the MAP estimation has led to considerable drops in the performances in the \textit{hopper} and \textit{walker2d} environments, which reaffirms that it is critical to have the full posterior distribution over the target values such that we can make conservative target predictions.

\paragraph{Adaptive weighting}
For \hyperlink{rq5}{RQ5}, we also considered an alternative way of combining $\Rhat{h}$ $\forall h$ by explicitly assigning a fixed set of weights: uniform or geometric.
We call the latter $\lambda$-weighting, in reference to the idea of TD$(\lambda)$ \citep{sutton1988tdlambda}.  We evaluated the performance of the fixed weighting scheme with various $\lambda \in(0,1)$ values, and report the full results in Appendix \ref{appendix:ablations}. In summary, there are some $\lambda$ values that work well in a specific task. However, it is hard to pick a single $\lambda$ that works across all environments, and thus $\lambda$ should be tuned as a hyperparameter. In contrast, \acronym~can avoid this problem by automatically adapting the rollout horizon. 

\paragraph{Benefits of full posterior estimation}
To ablate the benefits of using the full posterior distribution in conservative policy optimization, we have compared \acronym~to a quantile-based approach that calculates the conservative estimate through the $\alpha$-quantile of the sampled returns $\hat{y}(\state{},\action{},\state{}')$ (\ref{eq:h-return}) from the ensemble. The experimental details and results are reported in Appendix \ref{appendix:ablations}. In summary, we have found that \acronym~consistently outperformed this baseline on all tasks considered, and \acronym~was more stable during training, showing the effectiveness of the Bayesian formulation.

\section{Related Work} \label{sec:related-work}



In the pure offline RL setting, it is known that the direct application of off-policy algorithms fails due to value overestimation and the resulting policy distribution shift \citep{kumar2019bear, kumar2020cql, fujimoto2021td3+bc, yu2021combo}. 
Hence, it is critical to strike the balance between \textit{conservatism} and \textit{generalization} such that we mitigate the extent of policy distribution shift while ensuring that the learned policy $\pi_\theta$ performs better than behavior policy $\pi_\beta$.
Below, we discuss how existing model-free and model-based methods address these problems in practice.

\paragraph{Model-free offline RL}
\textit{Policy constraint} methods directly constrain the deviation of the learned policy from the behavior policy.
For example, BRAC \citep{wu2019brac} and BEAR \citep{kumar2019bear} regularize the policy by minimizing some divergence measure between these policies (e.g., MMD or KL divergence). Alternatively, BCQ \citep{fujimoto2019bcq} learns a generative model of the behavior policy and uses it to sample perturbed actions during policy optimization. On the other hand, \textit{value regularization} methods such as CQL \citep{kumar2020cql} add regularization terms to the value loss in order to implicitly regulate the distribution shift \citep{kostrikov2021fisherbrc,wang2020crr}.
Recently, some simple yet effective methods have been proposed. For example, TD3+BC \citep{fujimoto2021td3+bc} adds a behavioral cloning regularization term to the policy objective of an off-policy algorithm (TD3) \citep{fujimoto2018td3} and achieves SOTA performances across a variety of tasks. Also, by extending Clipped Double Q-learning \citep{fujimoto2018td3} to an ensemble of $N$ $Q$ functions, EDAC \citep{an2021edac} achieves good benchmark performances. 

\paragraph{Model-based offline RL}
Arguably, the learning paradigm of offline RL strongly advocates the use of a dynamics model, trained in a supervised way with a fixed offline dataset. Although a learned model can help generalize to unseen states or new tasks, model bias poses a significant challenge.  Hence, it is critical to know when to trust the model and when not to. MOPO \citep{yu2020mopo} and MOReL \citep{kidambi2020morel} address this issue by constructing and learning from a pessimistic MDP whose reward is penalized by the uncertainty of the state prediction. On the other hand, COMBO \citep{yu2021combo} extends CQL within the model-based regime
by regularizing the value function on OOD samples generated via model rollouts. \citet{rigter2022rambo} also takes an adversarial approach by optimizing the policy with respect to a worst-case dynamics model. In contrast to these, \acronym~estimates a full Bayesian posterior over values by using ensembles of models and value functions during policy evaluation of an actor-critic algorithm.
In principle, having the full distribution that \acronym~provides could also facilitate the use of other risk-informed statistics and epistemic risk measures to address value overestimation (see, e.g., \citet{eriksson2020epistemic}).


\paragraph{Model-based value expansion}
Unlike Dyna-style methods that augment the dataset with model-generated rollouts \citep{sutton1990dyna, janner2019mbpo}, MVE \citep{feinberg2018mve} uses them for better estimating TD targets during policy evaluation. While equally weighted $h$-step model returns were used in MVE, STEVE \citep{buckman2018steve} introduced an adaptive weighting scheme from the optimization perspective by approximately minimizing the variance of the MSBE loss, while ignoring the bias. Interestingly, the Bayesian posterior mean (i.e., the MAP estimator) we derive in \eqref{eq:posterior} matches the weighting scheme proposed in STEVE. 
However as we show in Figure \ref{fig:intro-fig} and \ref{fig:ablation-steve}, using the MAP estimator as value prediction in the offline setting often results in largely overestimated $Q$ values, which immensely hampers policy learning. See Section \ref{sec:bayesian} for the related discussion.

\section{Conclusion} \label{sec:conclusion}

In this paper, we present \acronym: conservative Bayesian model-based value expansion (MVE) for offline policy optimization. \acronym~is a model-based offline RL algorithm that trades off model-free and model-based value estimates according to their respective epistemic uncertainty during policy evaluation while facilitating conservatism by taking a lower bound on the Bayesian posterior value estimate. 
Viewing each $h$-step MVE target as a conditionally independent noisy observation of the \textit{true} target value under the learned MDP, we derive the Bayesian posterior distribution over the target value.
For a practical implementation of \acronym, we use the ensemble of dynamics and that of $Q$ function to sample MVE targets to estimate the Gaussian parameters, which in turn are used to compute the posterior distribution. 
Through empirical and analytical analysis, we find that 
the MAP estimator of the posterior distribution could easily lead to value overestimation when the learned MDP is not accurate under the current policy. 
In contrast, \acronym~constructs the LCB from the Bayesian posterior as a conservative estimation of the target value to successfully mitigate the issue while achieving state-of-the-art performance on several benchmark datasets. 



\pagebreak

\bibliography{references}
\bibliographystyle{iclr2023_conference}

\clearpage

\appendix

\section{Assumptions} \label{sec:appendix-assumptions}

In this part, we discuss and analyze the core assumptions that we have made in the derivation and implementation of \acronym.  First, recall that we view different $h$-step MVE returns $\Rhat{h}$ for all $h=0,\dots, H$ as \textit{conditionally independent} observations of the true underlying parameter $\qm{\pi}$.  Second, we have modeled the likelihood of the observations with the Gaussian distribution with mean $\mu_h$ and standard deviation $\sigma_h$, which we estimate via sampling from the ensemble of dynamics and that of $Q$ function.  Third, we use the improper prior, which still provides us a proper posterior distribution that is also Gaussian. Below, we describe in more detail about each of these assumptions.

\subsection{The Conditional Independence Assumption} \label{subsec:assumption-conditional-independence}
In order to meet the conditional independence assumption between $\Rhat{h}$, we need to estimate each $\Rhat{h}$ with samples that are independently sampled.  One way of achieving this is to generate samples per each $h$, resulting in an algorithm that requires $\mathcal{O}(NH^2)$ samples (and computation).  However, we have found that \textit{there is no specific benefit in this computational intensive sampling procedure in terms of the final performance}.  Hence, our practical implementation only performs the forward sampling once, reducing the computational cost down to $\mathcal{O}(NH)$.


\subsection{The Bayesian Posterior Estimation}

\paragraph{The improper prior assumption}

We have used the improper (or uninformative) prior in deriving \acronym~in Section \ref{sec:bayesian}.  Not to mention that the improper priors have been widely used in literature \citep{wasserman2010statistics,berger1985statistical,christensen2011bayesian}, we further argue that it is quite natural (and sometimes necessary) not to assume any prior information if we are to apply our algorithm to general environments/tasks that have different dynamics.
When some prior information is available, however, it is possible to incorporate it as long as we can use a conjugate prior that leads to a closed-form posterior update.  It is critical to keep the posterior in closed-form since otherwise we have to resort to, e.g., posterior sampling, which will substantially (and unnecessarily) increase the computational footprint.  


\paragraph{Empirical evidence supporting the Gaussian assumption over $\prob{\Rhat{h}~|~\qm{\pi}}$}

First, note that the true return distribution should have a single peak in the locomotion environments we consider due to their deterministic nature, as long as the policy is deterministic.  However, model-generated returns can have bimodality in their distributions since different models in the dynamics ensemble can lead to different trajectories, some of which can early terminate with low returns, while others continue to receive larger returns.  Hence, it is interesting to examine whether it is reasonable to assume the Gaussian distribution over the $h$-step returns. 

\begin{figure}[t!]
    \centering
      \subfigure[\textit{halfcheetah-mr}] {\label{fig:5-halfcheetah} \includegraphics[width=0.4\linewidth]{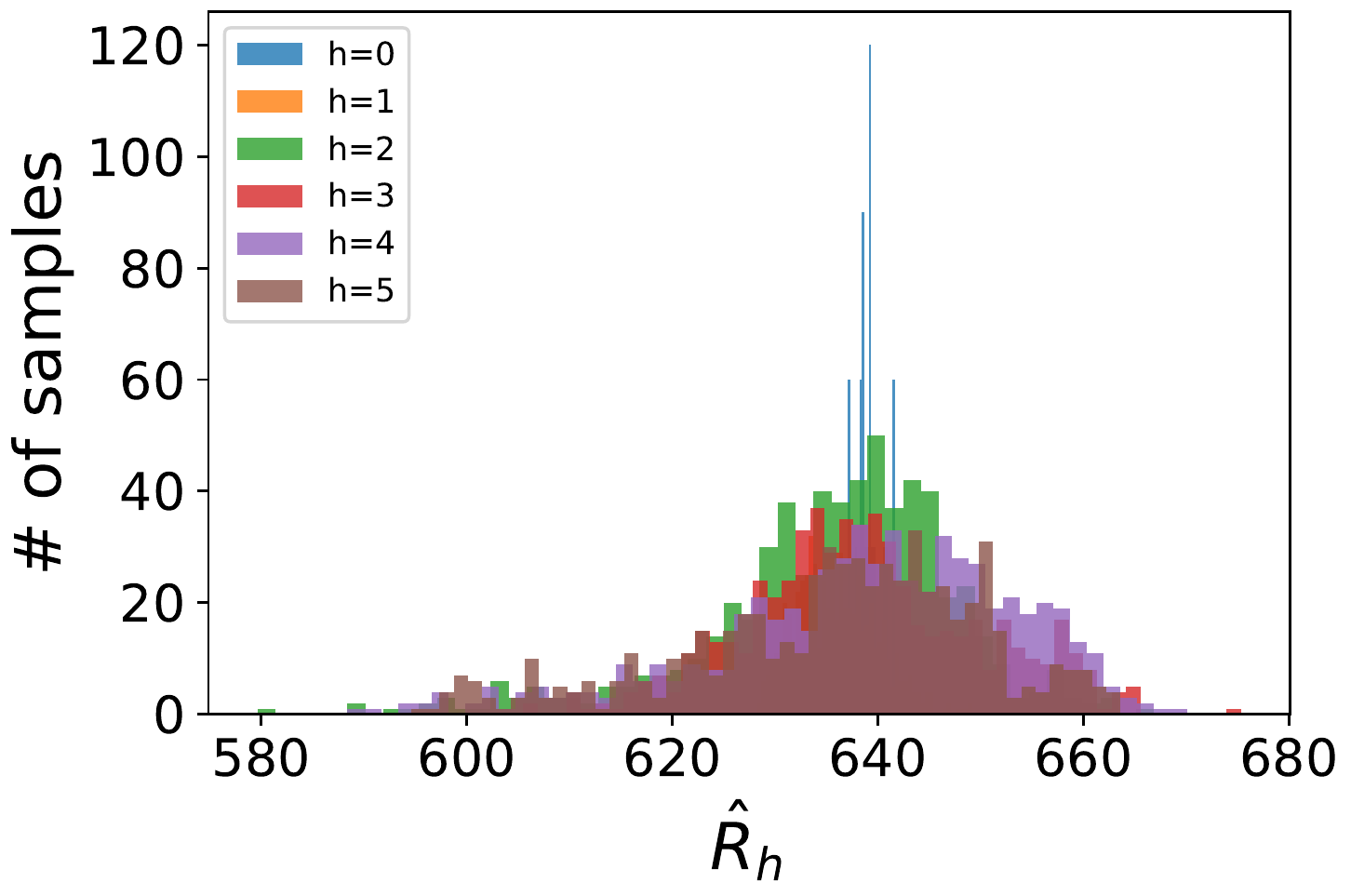}}
      \subfigure[\textit{hopper-mr}] {\label{fig:5-hopper} \includegraphics[width=0.4\linewidth]{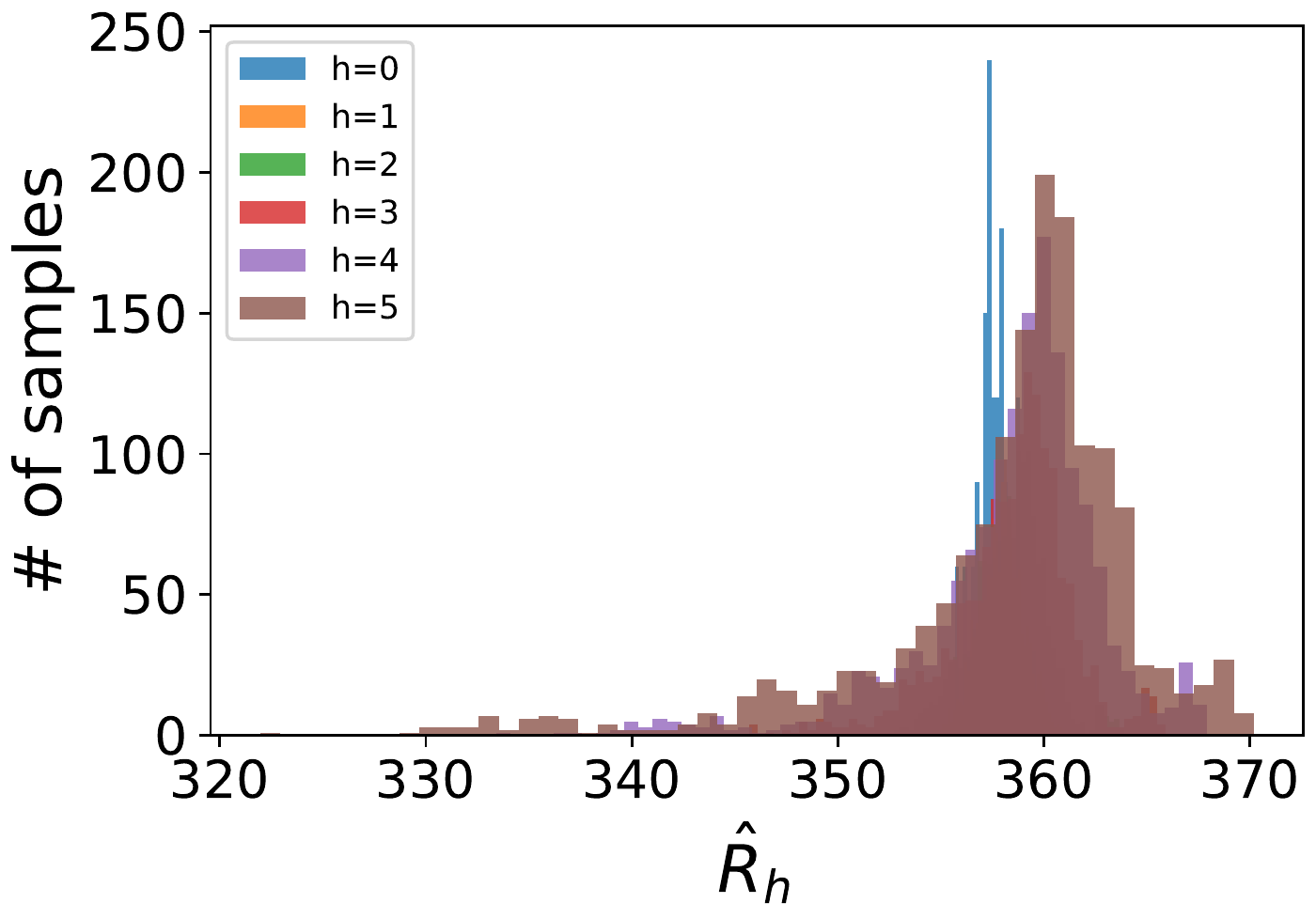}}
      \subfigure[\textit{walker2d-mr}] {\label{fig:5-walker2d} \includegraphics[width=0.4\linewidth]{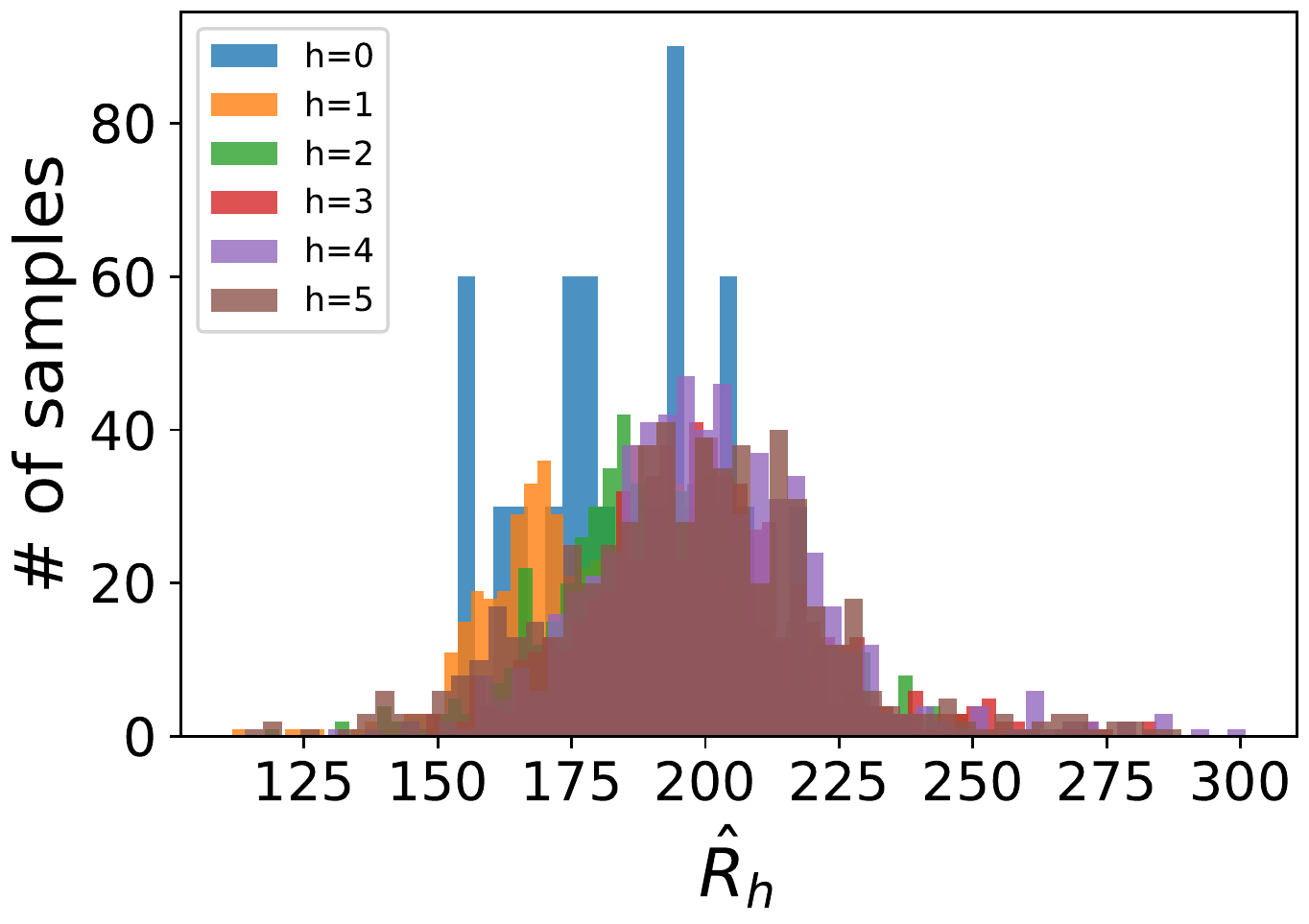}}
      \subfigure[A state in \textit{walker2d} showing the bimodality feature] {\label{fig:5-walker2d-bimodel} \includegraphics[width=0.4\linewidth]{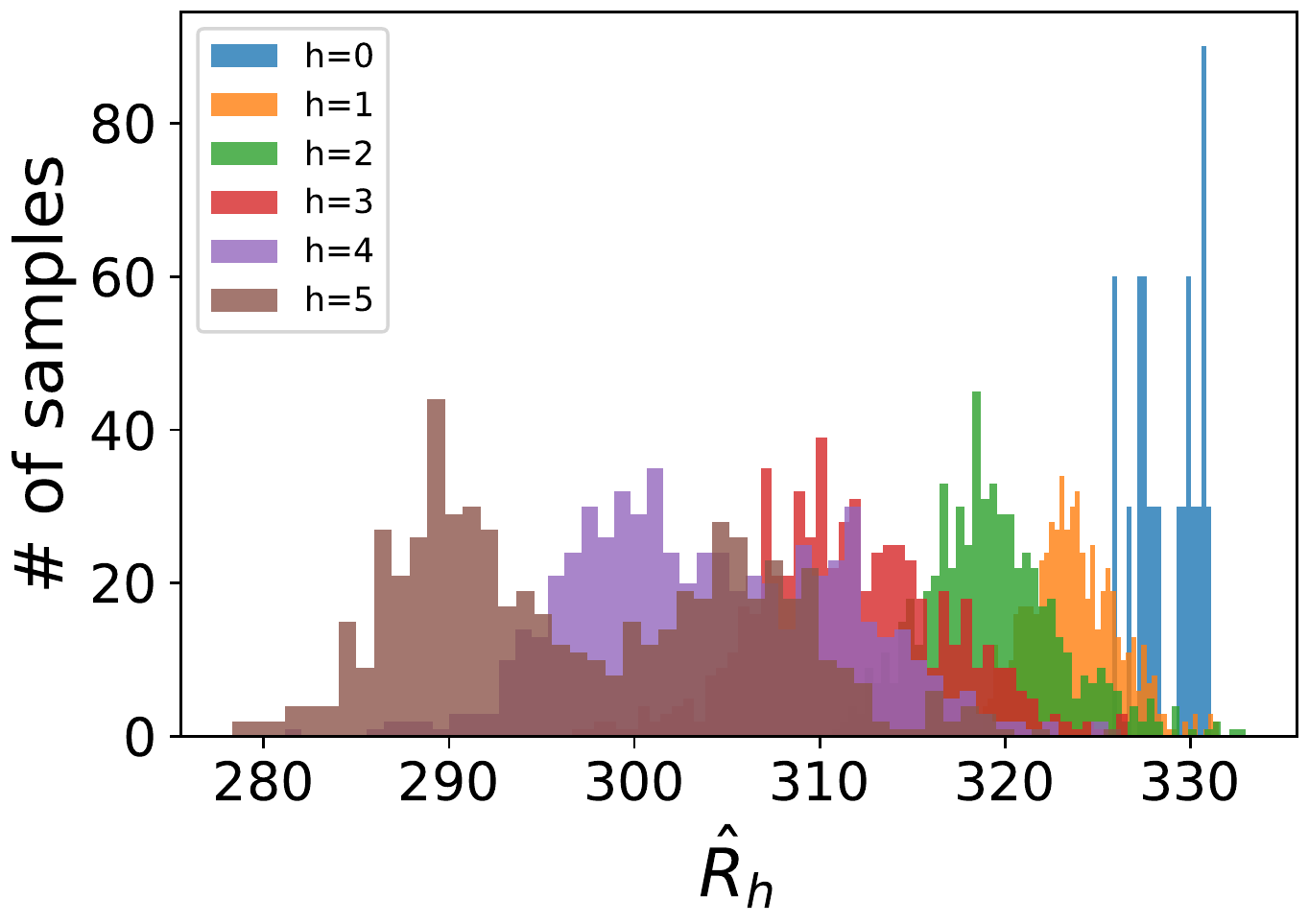}}
    \caption{The histogram of $\Rhat{h}$ $\forall h\in[0,5]$ of a randomly selected state during training, evaluated across three locomotion environments with the \textit{medium-replay-v2} configuration.}
    \label{fig:histograms}
\end{figure}

To answer this question, we have plotted the histograms of $h$-step returns for different $h$ values in three tasks: \textit{halfcheetah-mr}, \textit{hopper-mr}, and \textit{walker-mr}. Figure \ref{fig:histograms} (a)-(c) show that it is reasonable to assume $\Rhat{h}$ are normally distributed.  We have also observed that the empirical distribution of $\Rhat{h}$ sampled from certain states can have bimodality (Figure \ref{fig:histograms}d).  Notice that the histograms are more spread out as $h$ increases, which is due to compounded model errors.  However, we note that the Gaussian distribution can still capture the support of the return distribution reasonably well.  

\paragraph{The Gaussian likelihood assumption}
As discussed above and shown in Figure \ref{fig:histograms}, the Gaussian assumption captures the actual return distributions reasonably well. 
Although it is possible to derive a closed-form posterior update in Student t distribution by making an additional assumption in the variance of $\Rhat{h}$ likelihood (nb. we omit the actual derivation as it is not the contribution of this paper), we have observed that this does not lead to meaningful performance improvements compared to the much simpler Gaussian posterior that we derive in Section \ref{sec:bayesian}.  


\section{Algorithm Details}
\subsection{Algorithm Summary}\label{appendix:algo-summary}

\begin{algorithm}[t!]
\caption{\acronym: Conservative Bayesian MVE for Offline Policy Optimization} \label{alg:cbop}
\begin{algorithmic}[1]
    \State {\bfseries Input:} Data $\mathcal{D}$, discount factor $\gamma$, 
    rollout horizon $H$, LCB coefficient $\psi$
    \State Initialize actor $\pi_\theta$, $Q$ ensemble $Q_\phi$ and target $Q_{\phi'}$, dynamics ensemble $\hat{f}_k = (\hat{T}_k, \hat{r}_k) \forall k$
    \State Pretrain $\hat{f}_\xi$ on $\mathcal{D}$ till convergence
    \State Pretrain $\pi_\theta$ and $Q_\phi$ on $\mathcal{D}$ with BC and FQE respectively (Appendix \ref{appendix:bc+pe})
    
    \While{$\pi_\theta$ not converged}
        \State Sample a batch of transitions $B=\{\tau_i: ~\tau_i=(\state{}, \action{}, r, \state{}')_i \}_{i=1}^{|B|}\subset\mathcal{D}$
        
        \For{$\tau_i\in B$}  \Comment{this step happens in parallel for all $\tau_i\in B$}
            \State $\mstate{0}^k\leftarrow \state{},\mstate{1}^k\leftarrow \state{}',\maction{0}^k\leftarrow \action{}, \rhat{0}^k\leftarrow r,\quad\forall k\in[1,K]$
            \For{$h=0$ to $H$}
                \If{$h\ge 1$}
                    \State Sample an action $\maction{h}^k \sim\pi_\theta(\mstate{h}^k)~\forall k$
                    \State Sample next state transition and reward  $(\mstate{h+1}^k, \rhat{h}^k)\leftarrow \hat{f}_k(\mstate{h}^k,\maction{h}^k)~\forall k$
                \EndIf
                \State $\Rhat{h}^{k, m}\leftarrow \sum_{t=0}^h \gamma^t \rhat{t}^k + \gamma^{h+1}\qm{m}_{\phi'}(\mstate{h+1}^k,\maction{h+1}^k)~~\forall m$
            \EndFor   
            \State Compute $\mu_h$ and $\sigma_h$ by \eqref{eq:total-expectation} and \eqref{eq:total-variance}, respectively
            \State Estimate $\mu, \sigma^2$ of $\prob{\qm{}|\Rhat{0}, \ldots, \Rhat{H}} \sim \mathcal{N}(\mu, \sigma^2)$ by \eqref{eq:posterior}
            \State Compute target $Q$ value: $y_i(\state{},\action{},\state{}') \leftarrow \mu - \psi\sigma$
        \EndFor
        \State Update $\pi_\theta$ and $Q_\phi$ following an off-policy actor-critic algorithm (e.g., SAC \cite{haarnoja2018sac})
        \State Update the target network $Q_{\phi'}$
    \EndWhile
\end{algorithmic}
\end{algorithm}


Algorithm \ref{alg:cbop} summarizes \acronym. In lines $20$-$21$, we can use any off-policy actor-critic algorithm as the backbone of our approach, since the only part that changes is the computation of the target value $y(\state{}, \action{}, \state{}')$. In this work, we follow EDAC \citep{an2021edac} --- which builds on SAC \citep{haarnoja2018sac} --- because it also employs $Q$ ensembles. 
As discussed in Appendix \ref{appendix:bc+pe}, a large discrepancy in the scale of the terminal $Q_{\phi'}$ predictions and that of the model-based rollout returns $\sum \gamma^t \rhat{t}$ in the initial iterations greatly hampers policy learning. Hence, we pretrain the policy $\pi_\theta$ and $Q_\phi$ with with behavioral cloning (BC) and policy evaluation (PE) as elaborated in Appendix \ref{appendix:bc+pe}.

\subsection{Dynamics Model Architecture}
In this work, we approximate the true dynamics with a probabilistic ensemble model introduced by PETS \citep{chua2018pets}.
We follow the common configurations used in the literature, e.g., MBPO \citep{janner2019mbpo} and MOPO \citep{yu2020mopo}.
Each model in the ensemble has $4$ fully-connected layers with $200$ neurons.
Specifically, we train the ensemble of $30$ models, from which we select $20$ models (often called `elite') with smaller validation errors.
For next state predictions, we train the ensemble model to predict the \textit{delta} states, or $\Delta=\state{}'-\state{}$ for $(\state{}, \state{}')\in\mathcal{D}$.  We normalize the inputs and outputs of the model for training and evaluation. 

The approach for training the dynamics ensemble closely follows previous work on Bayesian ensemble estimation \citep{chua2018pets,janner2019mbpo}.  To reduce the effect of correlation, we follow the existing work by using independent initialization for each ensemble member and by training each of them using different mini-batches sampled from the dataset.  Although in practice some correlation may be inevitable, there are several key advantages to estimating uncertainty in this way. Firstly, bootstrapped uncertainty estimates have been shown to have strong theoretical properties --- see, e.g. \cite{efron1982jackknife} or \cite{breiman1996bagging}.  Secondly, bootstrapping avoids the computational challenges associated with estimating the uncertainty of model predictions directly, and our experiments have shown that the uncertainty we obtained was indeed well-calibrated. For further details, please see the expected horizon analysis shown in Figure \ref{fig:expected-horizon} and Section \ref{subsec:expected-horizon}, which demonstrates the effectiveness of \acronym~subject to different qualities of the learned dynamics ensemble.

\subsection{Pretraining} \label{appendix:bc+pe}

\begin{algorithm}[t!]
\caption{FQE: Fitted Q-Evaluation \citep{le2019batch}} \label{alg:fqe}
\begin{algorithmic}[1]
    \State {\bfseries Input:} Dataset $\mathcal{D}=\{\state{i}, \action{i}, r_i, \state{i}'\}_{i=1}^n$, policy $\pi$ to be evaluated
    \State Initialize the parameters of $Q_{\phi^{(0)}}$ randomly
    \For{$t=1,\dots,T$}
        \State Compute the targets $y_i=r_i+\gamma Q_{\phi^{(t-1)}}(\state{i}', \pi(\state{i}'))~\forall i$
        \State Build the training set $\mathcal{D}^{(t)}=\{(\state{i}, a_i), y_i\}_{i=1}^n$
        \State Solve a supervised learning problem:\\
        \qquad\quad $\phi^{(t)} = \arg\min_{\phi} \E{\{(\state{i}, \action{i}), y_i\}\sim \mathcal{D}^{(t)}}{ \left(Q_\phi(\state{i}, \action{i}) - y_i\right)^2}$
    \EndFor
    \State $\phi\leftarrow \phi^{(T)}$
    \State \textbf{return} $Q_\phi$
\end{algorithmic}
\end{algorithm}

\begin{figure}[t!]
    \centering
      \subfigure[Random initialization] {
      \captionsetup{font=small}
      \label{fig:scale-a} \includegraphics[width=0.35\textwidth]{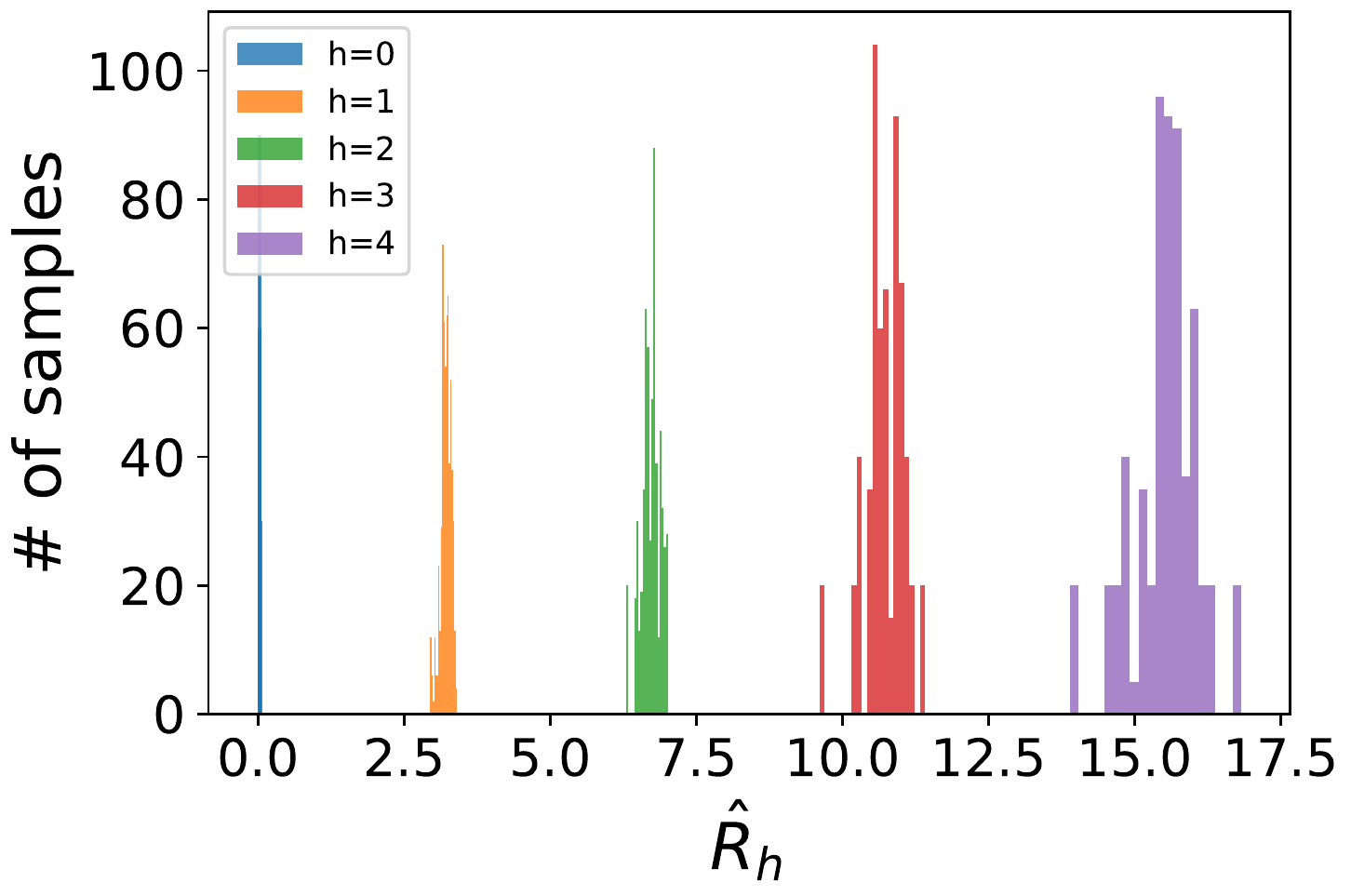}}
      \subfigure[Pretrained by BC+PE] {
      \captionsetup{font=small}
      \label{fig:scale-b}  \includegraphics[width=0.35\textwidth]{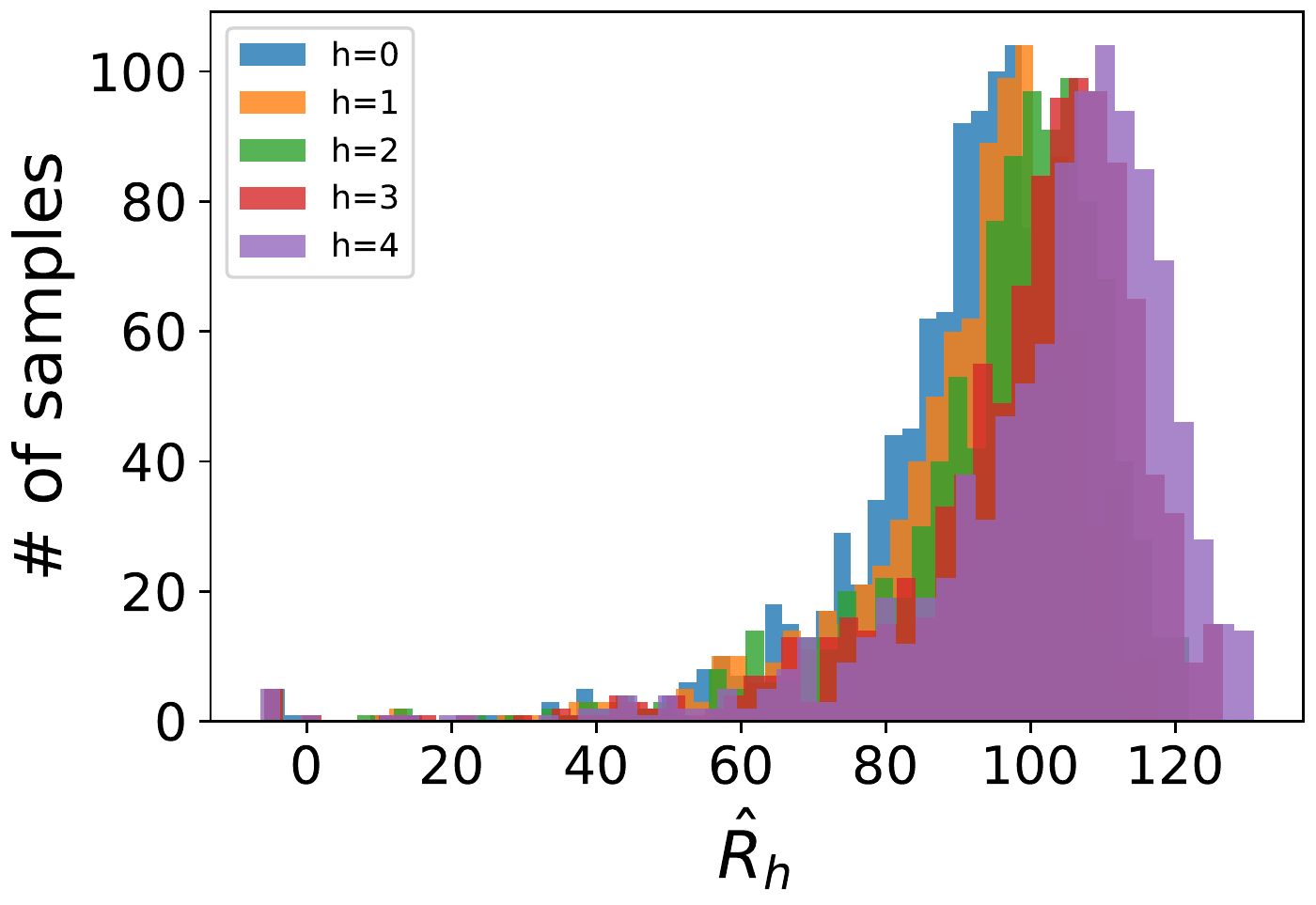}}
    \vspace*{-5pt}
    \caption{The histogram of $\Rhat{h}$ $\forall h\in[0,4]$ evaluated on \textit{halfcheetah-medium-v2}}
    \vspace{-0pt}
  \label{fig:scale-diff}
\end{figure}%

In some environments, we notice that training $Q_\phi$ and $\pi_\theta$ from scratch could be challenging, and Figure \ref{fig:scale-diff} illustrates the reason.
Remember that we pretrain the dynamics ensmeble with the offline data $\mathcal{D}$ before starting the policy optimization.  This means that the reward predictions made by the learned model would have the proper scale. 
On the other hand, the $Q_{\phi'}$ ensemble is initialized with small random values.  Hence, in the early iterations of policy learning, even though the $Q_\phi$ ensemble has not been trained yet, its predictions have a very small variance compared to the model-based rollout returns given by the learned dynamics ensemble (Figure \ref{fig:scale-a}). 
This will then lead to all weights being concentrated on $\Rhat{0}$, effectively MF; the MB rollouts would only slow down learning without contributing anything in this case. 
Besides, the variance of $Q_{\phi'}$ ensemble would be negligible, suggesting that taking the LCB would not introduce a sufficient level of conservatism into learning, which can hurt the performance.

Therefore in the experiments, we pretrain $Q_\phi$ and $\pi_\theta$ with the offline data.
Specifically, we use behavior cloning (BC) for the policy network $\pi_\theta$.  In BC, we minimize the mean squared loss $\mathcal{L}_{\mathrm{BC}}(\theta)=\mathbb{E}_{(\state{}, \action{})\sim\mathcal{D}}[(\action{} - \pi_\theta(\state{}))^2]$.~
For the value network $Q_\phi$, we perform policy evaluation (PE) using Fitted Q-Evaluation (FQE) \citep{le2019batch}, which is schematically explained in the pseudocode in Algorithm \ref{alg:fqe}.  In line 4, when the policy to be evaluated is the behavior policy $\pi_\beta$, we can take the recorded next action $\action{i+1}$ from $\mathcal{D}$ in place of $\pi(\state{i}')$.

More concretely, at each iteration $t$ of FQE, a supervised learning dataset $\mathcal{D}^{(t)}=\{(\state{i}, \action{i}), y_i\}_{i=1}^n$ is constructed by estimating the target value $y_i$ for each $(\state{i}, \action{i})\sim\mathcal{D}$ with the current $Q$ approximation $Q_{\phi^{(t-1)}}$ and the associated transition tuple $(\state{i}, \action{i}, r_i, \state{i}')$ via $y_i=r_i+\gamma Q_{\phi^{(t-1)}}(\state{i}', \pi(\state{i}'))$.
We then update the $Q$ function parameters $\phi$ by minimizing the MSE loss.  That is, $\phi^{(t)}\leftarrow \arg\min_\phi \frac{1}{n}\sum_{i=1}^n [Q_{\phi^{(t-1)}}(\state{i}, \action{i}) - y_i]^2$.
FQE repeats the two steps (i.e., constructing the dataset and minimizing the MSE loss) to learn the $Q_\phi$ ensemble model.


\section{Experiment Details}\label{appendix:configurations}

\subsection{Experimental settings} \label{appendix:experiment-settings}
\paragraph{D4RL MuJoCo Gym} We use the \textit{v2} version for each dataset as provided by the D4RL library \citep{fu2020d4rl}.  Following Algorithm \ref{alg:cbop}, we pretrain $\pi_\theta$ and $Q_\phi$ with BC and FQE, respectively.  The resulting policy and the $Q$ ensemble are trained for $1,000$ more epochs using \acronym. Table \ref{table:mujoco} reports the mean and standard deviation obtained from $5$ random seeds.


\paragraph{Comparision of target $Q$ values of \textbf{MAP} and \textbf{\acronym}~(Figure \ref{fig:intro-a})} In Figure \ref{fig:intro-a}, we compare the MAP estimation with the LCB in the \textit{hopper-random} dataset. We have plotted the mean and $\pm$ one standard error over the course of training. The MAP estimation simply uses the mean $\mu$ in \eqref{eq:posterior} as the target $y(\state{},\action{},\state{}')$, where as the LCB utilizes the variance of the posterior distribution to compute $y(\state{},\action{},\state{}')=\mu-\psi \cdot \sigma$.  Note that we can also use other conservative estimate of the target using the posterior distribution; for example, we can use value-at-risk (VaR), conditional value-at-risk (CVaR) or other quantiles.

\paragraph{Expected rollout horizon of \acronym~(Figure \ref{fig:intro-b} and Figure \ref{fig:expected-horizon})} In Figure \ref{fig:intro-fig} and \ref{fig:expected-horizon}, we report the expected rollout horizon values. The expected rollout horizon can be computed per each sample in the batch during policy training, and we have reported the average value across all samples in a batch.

\subsection{Hyperparameters}
\begin{table*}[t!]
\footnotesize
\caption{The LCB coefficient $\psi$ used in the D4RL MuJoCo Gym experiments.}
\begin{center}
\begin{tabular}{lccc}
\toprule
            & \multicolumn{3}{c}{$\psi$} \\
            \cmidrule(lr{.75em}){2-4}
Task Name   & halfcheetah   & hopper    & walker2d \\
\midrule
\midrule
random          & $3.0$       & $5.0$ & $5.0$ \\ 
medium          & $0.5$       & $3.0$ & $3.0$ \\
medium-replay   & $0.5$       & $2.0$ & $2.0$ \\
medium-expert   & $3.0$       & $3.0$ & $3.0$ \\
expert          & $5.0$       & $3.0$ & $3.0$ \\
full-replay     & $2.0$ & $3.0$ & $2.0$ \\
\bottomrule
\end{tabular}
\end{center}
\label{table:hyperparameters}
\end{table*}

Table \ref{table:hyperparameters} summarizes the \acronym~hyperparameters we use in the experiments presented in Section \ref{sec:experiments}.  The only hyperparameter that we have tuned is the LCB coefficient $\psi$ through the grid search over the set $\{0.5, 2.0, 3.0, 5.0\}$.  We have used $H=10$, $K=20$, $M=20$, and $lr=3\times 10^{-4}$ for all experiments, except for the \textit{hopper} environment where we used $M=50$.\footnote{
In the early stage of algorithm development, we selected the \textit{medium} configuration from the three environments in the D4RL benchmark and used $M=20$ for all experiments when testing the performance of \acronym.  It turned out that \acronym~works well in the HalfCheetah and Walker2d environments without tuning, but we found that we needed to have a larger value ensemble to get reasonable performance in the Hopper environment.  We chose $M=50$. since it worked well and this choice is also supported by previous work \citep{an2021edac}.  Accordingly during hyperparameter tuning, we used $M=50$ for Hopper and $M=20$ for the other two environments.
} The LCB parameters reported in Table \ref{table:hyperparameters} are tuned based on the final online evaluation performance from corresponding environments.

\paragraph{Offline Hyperparameter Selection via FQE \citep{paine2020hyperparameter}}

When strictly adhering to the offline paradigm of policy learning, it is crucial to restrict access to online interactions at all stages of learning including the hyperparameter selection.However, many existing works still use the online evaluation for hyperparameter selection \citep{an2021edac,wang2021romi,fujimoto2021td3+bc,chen2021decision} and we followed the same evaluation protocol for tuning the hyperparameters of our method.  We believe there is a dire need for standardizing the evaluation protocol in the offline RL, but this work should be addressed by the offline RL research community as a whole, which is beyond the scope of our paper.
One important way to reduce the amount of online interactions used for hyperparameter selection is to minimize the number of hyperparameters to tune.  In this regard, \acronym~is particularly advantageous since we need only to tune the LCB coefficient $\psi$.

To further validate the choice of $\psi$ values in Table \ref{table:hyperparameters}, we performed a post hoc analysis following the hyperparameter selection work proposed in \cite{paine2020hyperparameter}.  To this end, we considered three data configurations (\textit{m}, \textit{mr}, \textit{fr}) and two environments (\textit{halfcheetah}, \textit{walker2d}), and we retrieved the model checkpoints of the learned policy networks for all seeds.  Then, we evaluated each policy $\pi_\theta$ with the following metric:
\begin{equation}
    \E{\state{0}\sim\mathcal{D}}{Q_\zeta(\state{0}, \pi_\theta(\state{0})} \label{eq:fqe-hyperparam}
\end{equation}
Here, $\state{0}$ are the initial states stored in the offline dataset and $Q_\zeta$ is the value function associated with the policy $\pi_\theta$, which is obtained by running FQE (Algorithm \ref{alg:fqe}).  This $Q_\zeta$ is different from the learned value function $Q_\phi$, and \cite{paine2020hyperparameter} noted that using $Q_\zeta$ is better than using $Q_\phi$ for the purpose of hyperparameter selection. The candidate $\psi$ values are sorted based on the scores from \eqref{eq:fqe-hyperparam}, and we can use $\psi$ with the highest score.

Table \ref{table:fqe-vs-online-ranking} compares the rankings of the four $\psi$ values we considered in the experiments from FQE and the online evaluation. 
The rightmost column shows the Spearman's rank correlation coefficient ($\rho$) which is the correlation coefficient between the two sets of rankings.  Notably, the $\psi$ values selected via FQE match the values we obtained from the online evaluation for $4$ out of $6$ tasks. 
In \textit{halfcheetah-m}, $\psi=0.5$ has the online performance of $74.3$ (as reported in Table \ref{table:mujoco}), while the performance from $\psi=2$ is $72.4$ which is only slightly worse.  For \textit{walker2d-fr}, $\psi=2$ is at $107.8$ (reported in Table \ref{table:mujoco}) and $\psi=3$ gives $89.3$ when evaluated in the true environment.  Even if $\psi=3$ was chosen based on FQE, we can easily see that this is still a substantial improvement compared to the data-logging policy which has the average normalized score of $39.8$.

Overall, the Spearman's rank correlation values are always greater than or equal to $0.8$, suggesting that the rankings from FQE align very well with those from the online evaluation.  This suggests that (1) \acronym~can be reliably tuned solely with an offline dataset via FQE and that (2), with the benefit of hindsight, our selection of $\psi$ values in Table \ref{table:hyperparameters} is a valid one.

\begin{table*}[t!]
\scriptsize
\caption{Comparing the rankings of the LCB coefficient $\psi$ based on the online evaluation and FQE \citep{paine2020hyperparameter}}
\begin{center}
\begin{tabular}{lcccccc}
\toprule
                                &            &  \multicolumn{4}{c}{$\psi$}                                  & Rank correlation \\
            \cmidrule(lr{.75em}){3-6}
Task Name                       &   Ranking  &  $0.5$       &   $2.0$           &   $3.0$           & $5.0$ & $(\rho)$\\
\midrule
\midrule
\multirow{2}{*}{halfcheetah-m}  & FQE    &    $2$           & $\boldsymbol{1}$  &    $3$            & $4$ & \multirow{2}{*}{$0.8$} \\ \Tstrut{}
                                & Online & $\boldsymbol{1}$ &   $2$             &    $3$            & $4$ & \\\midrule \Tstrut{}
\multirow{2}{*}{halfcheetah-mr} & FQE    & $\boldsymbol{1}$ &   $2$             &    $4$            & $3$ & \multirow{2}{*}{$0.8$}\\ \Tstrut{}
                                & Online & $\boldsymbol{1}$ &   $2$             &    $3$            & $4$ & \\\midrule \Tstrut{}
\multirow{2}{*}{halfcheetah-fr} & FQE    & $3$              & $\boldsymbol{1}$  &    $2$            & $4$ & \multirow{2}{*}{$0.8$}\\ \Tstrut{}
                                & Online & $2$              & $\boldsymbol{1}$  &    $3$            & $4$ & \\\midrule \Tstrut{}
\multirow{2}{*}{walker2d-m}     & FQE    &    $4$           &   $2$             & $\boldsymbol{1}$  & $3$ & \multirow{2}{*}{$1.0$}\\ \Tstrut{}
                                & Online &    $4$           &   $2$             & $\boldsymbol{1}$  & $3$ & \\\midrule \Tstrut{}
\multirow{2}{*}{walker2d-mr}    & FQE    &    $4$           & $\boldsymbol{1}$  &   $2$             & $3$ & \multirow{2}{*}{$1.0$}\\ \Tstrut{}
                                & Online &    $4$           & $\boldsymbol{1}$  &   $2$             & $3$ & \\\midrule \Tstrut{}
\multirow{2}{*}{walker2d-fr}    & FQE    & $3$              &   $2$             & $\boldsymbol{1}$  & $4$ & \multirow{2}{*}{$0.8$}\\ \Tstrut{}
                                & Online & $3$              & $\boldsymbol{1}$  &   $2$             & $4$ & \\\bottomrule
\end{tabular}
\end{center}
\label{table:fqe-vs-online-ranking}
\end{table*}

\paragraph{Other considerations}

\acronym~trades off the uncertainty of the learned dynamics model with that of the learned $Q$ ensemble.  In practice, we use the ensemble models to implicitly capture the respective epistemic uncertainty. Hence, it is critical that the models we use indeed exhibit well-calibrated uncertainty in their predictions.  In this regard, we found that it is useful to incorporate the gradient diversification loss for the $Q$ ensemble as introduced in \citet{an2021edac}, which helps prevent the uncertainty in predictions from collapsing.  Instead of tuning the hyperparameter $\eta$ that controls the level of gradient diversification loss, we use a fixed number $\eta=1$ across all experiments.

\begin{figure}[t!]
    \centering
    \includegraphics[width=0.8\textwidth]{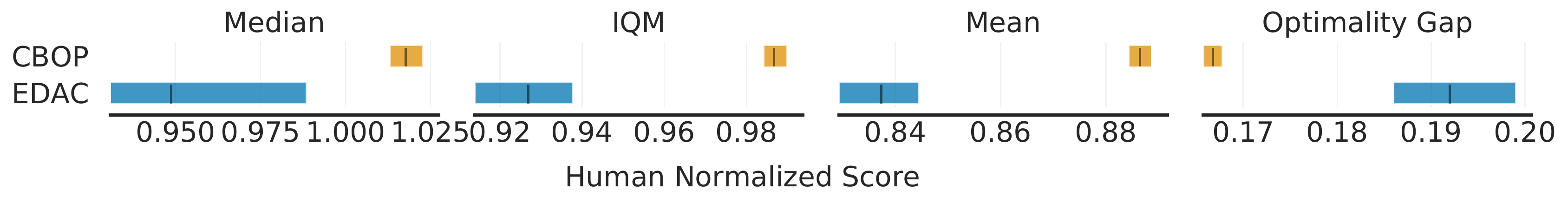}
    \caption{RLiable results across all $18$ locomotion tasks. Shaded regions show $95$\% CIs.  We refer readers to \citep{agarwal2021deep} for detailed explanation of the metrics considered.}
  \label{fig:rliable}
\end{figure}

Please note that the use of the ensemble diversification trick is orthogonal to our contributions in this work.  Furthermore, we provide a reliable performance comparison between \acronym~and EDAC to validate that \acronym~outperforms EDAC.  To this end, we use RLiable \citep{agarwal2021deep} which provides various metrics other than the simple average to more reliably determine the relative performances of compared methods.
Specifically, we have reproduced EDAC and compared its performance against \acronym~using the Median, IQM (interquartile mean), Mean, and Optimality Gap (Figure \ref{fig:rliable}).  In all metrics considered, \acronym~exhibits substantially better performance without overlapping $95$\% confidence intervals (CI).  In fact, another important performance metric, called the probability of improvement, of \acronym~against EDAC is $88.27$\%, which strongly indicates the superiority of \acronym.

\section{Additional Experiments}


\subsection{Conservatism Analysis} \label{appendix:conservatism-analysis}

\begin{table}[t!]
\footnotesize
\caption{A full comparison across three environments showing the difference between the values predicted by the learned $Q$ functions and the true discounted returns from the environment.}
\centering
\begin{tabular}{lcc c cc}
\toprule
      & \multicolumn{2}{c}{CQL} && \multicolumn{2}{c}{\acronym} \\
\cmidrule(lr{.75em}){2-3} \cmidrule(lr{.75em}){5-6}
    Task name & Mean & Max && Mean & Max \\
\midrule
\midrule
    hopper-m    & -61.84   & -3.20     && -55.83   & -16.21    \\
    hopper-mr   & -142.89  & -28.73    && -172.45  & -39.45    \\
    hopper-me   & -79.67   & -5.16     && -114.39  & -11.24    \\
\midrule
    halfcheetah-m   & -222.43  & -180.97    && -106.24  & -66.97    \\
    halfcheetah-mr  & -363.00  & -198.42    && -84.42   & -8.48     \\
    halfcheetah-me  & -310.95  & -23.74     && -210.51  & -54.58    \\
\midrule
    walker2d-m  & -167.36  & -8.88     && -84.70   & -15.00    \\
    walker2d-mr & -285.02  & -25.44    && -80.31   & -14.06    \\
    walker2d-me & -156.71  & -64.64    && -75.89   & -42.30    \\
\bottomrule
\end{tabular}
\label{tab:pe-full}
\end{table}

\begin{figure}[t!]
    \centering
      \subfigure[\textit{hopper-m}] {\label{fig:pe-m} \includegraphics[width=0.32\linewidth]{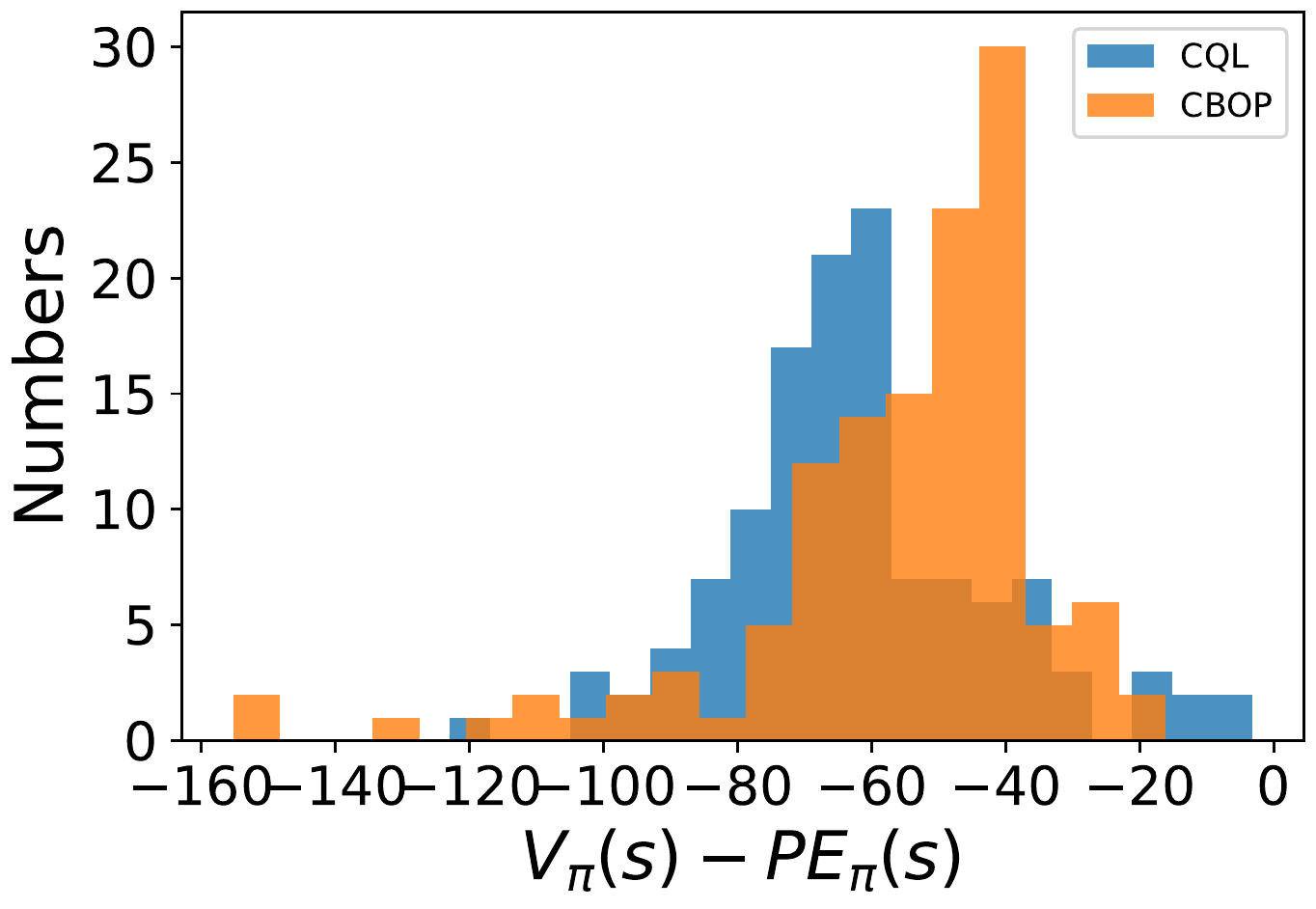}}
      \subfigure[\textit{hopper-me}] {\label{fig:pe-me}  \includegraphics[width=0.32\linewidth]{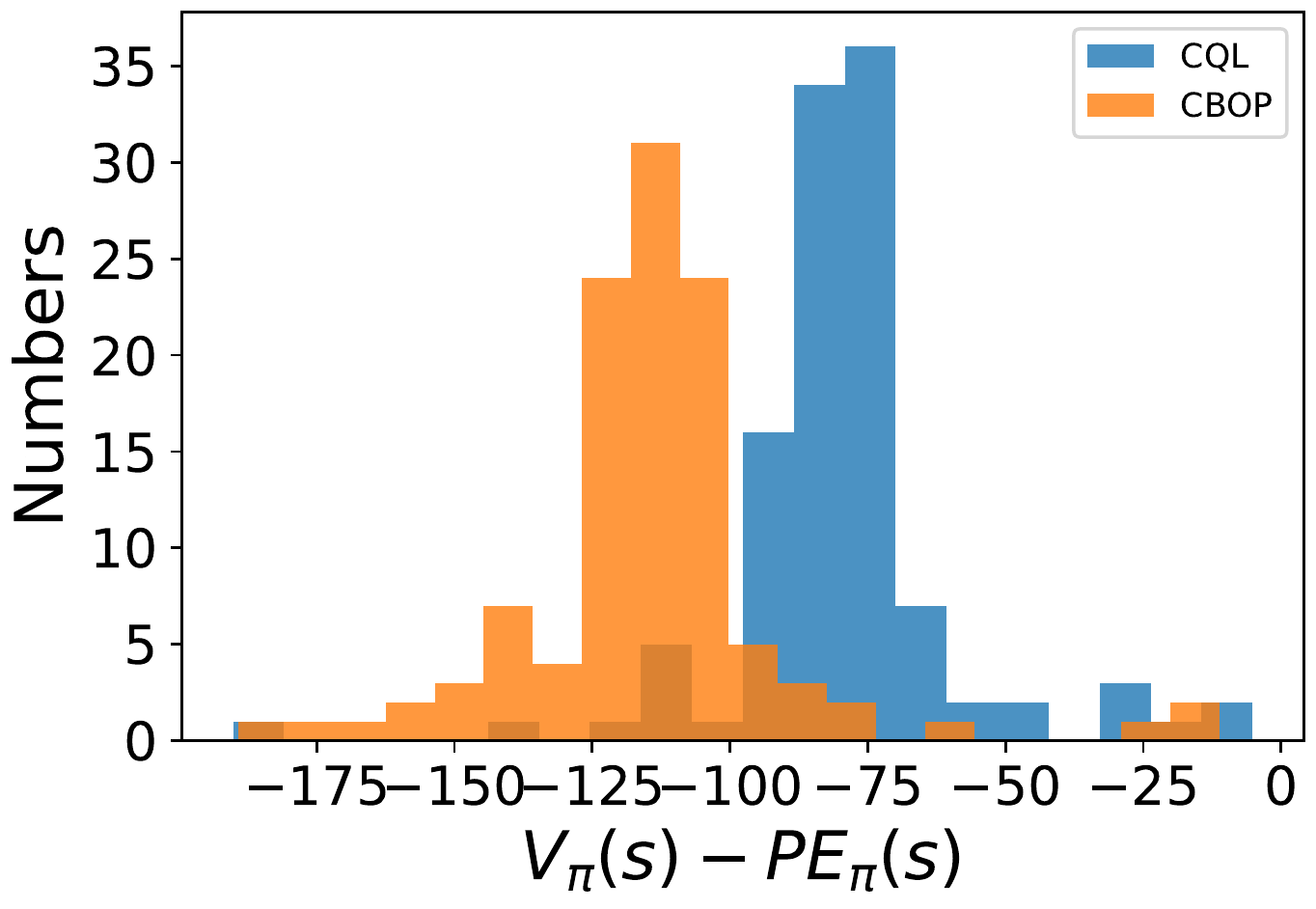}}
      \subfigure[\textit{hopper-mr}] {\label{fig:pe-mr}  \includegraphics[width=0.32\linewidth]{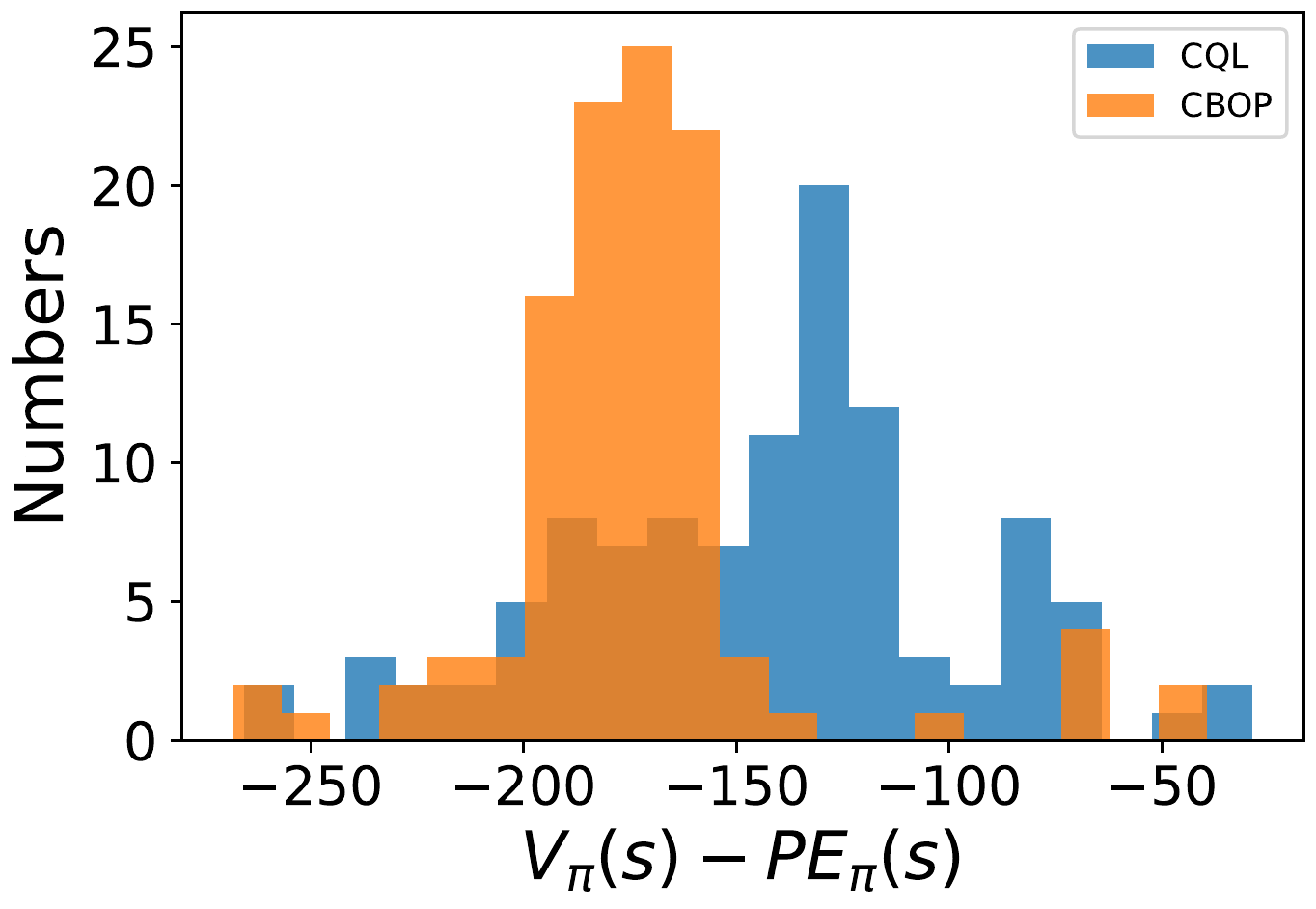}}
    \caption{The distribution of difference between policy values predicted by algorithms and Monte Carlo policy evaluation results in the true environment. Here, $s\sim\mathcal{D}, a=\pi(s)$.}
    \label{fig:pe-full-dist}
\end{figure}

In Section \ref{subsec:conservatism}, we have empirically verified that \acronym~indeed learns a conservative value function.  Specifically, given the offline dataset $\mathcal{D}$, we compute the following value difference:
\begin{equation}
    \mathbb{E}_{\state{}\sim \mathcal{D}}\big[\hat{V}^\pi(\state{}) - \mathbb{E}[V^\pi(\state{})]\big] \label{eq:conservatism}
\end{equation}
where we compute the true value $\mathbb{E}[V^\pi]$ via the Monte Carlo estimation in the true environment.  We have provided the comparison of CQL and \acronym~evaluated in the \textit{hopper} environment in Table \ref{tab:pe}, and Figure \ref{fig:pe-full-dist} shows the full histograms of \eqref{eq:conservatism} for in this environment. Furthermore, Table \ref{tab:pe-full} includes the results from all three MuJoCo locomotion environments. We can clearly see that \acronym~has learned a conservative value function in these tasks.

\subsection{Decomposition of $h$-Step Return Variance}

In Section \ref{sec:ensemble-sampling}, we have shown that the variance of $h$-step returns can be decomposed into $A$ and $B$ terms according to the law of total variance, which we restate here for ease of exposition:
\begin{equation}
    \sigma_h^2 = \var{\pi_\theta}{\Rhat{h}| \tau} = \underbrace{\E{\hat{f}_k}{\var{\pi_\theta}{\Rhat{h}| \tau, \hat{f}_k}}}_{A} + \underbrace{\var{\hat{f}_k}{\Egiven{\pi_\theta}{\Rhat{h}}{\tau, \hat{f}_k}}}_{B}. \label{eq:total-variance-re}
\end{equation}
Here, $A$ reflects the epistemic uncertainty from the $Q_{\phi'}$ ensemble, while $B$ accounts for the uncertainty derived from the learned dynamics ensemble. 
The beauty of \acronym~is that it can capture both uncertainties by sampling through the dynamics and value ensembles and subsequently compute the value target in a conservative way through the Bayesian posterior formulation.
A natural question may be whether $A$ would vanish and become unnecessary when the policy and value function have converged?

\begin{figure}[t!]
    \centering
      \subfigure[\textit{hopper-expert}] {\label{fig:A-hopper-r-eval-e}  \includegraphics[width=0.48\linewidth]{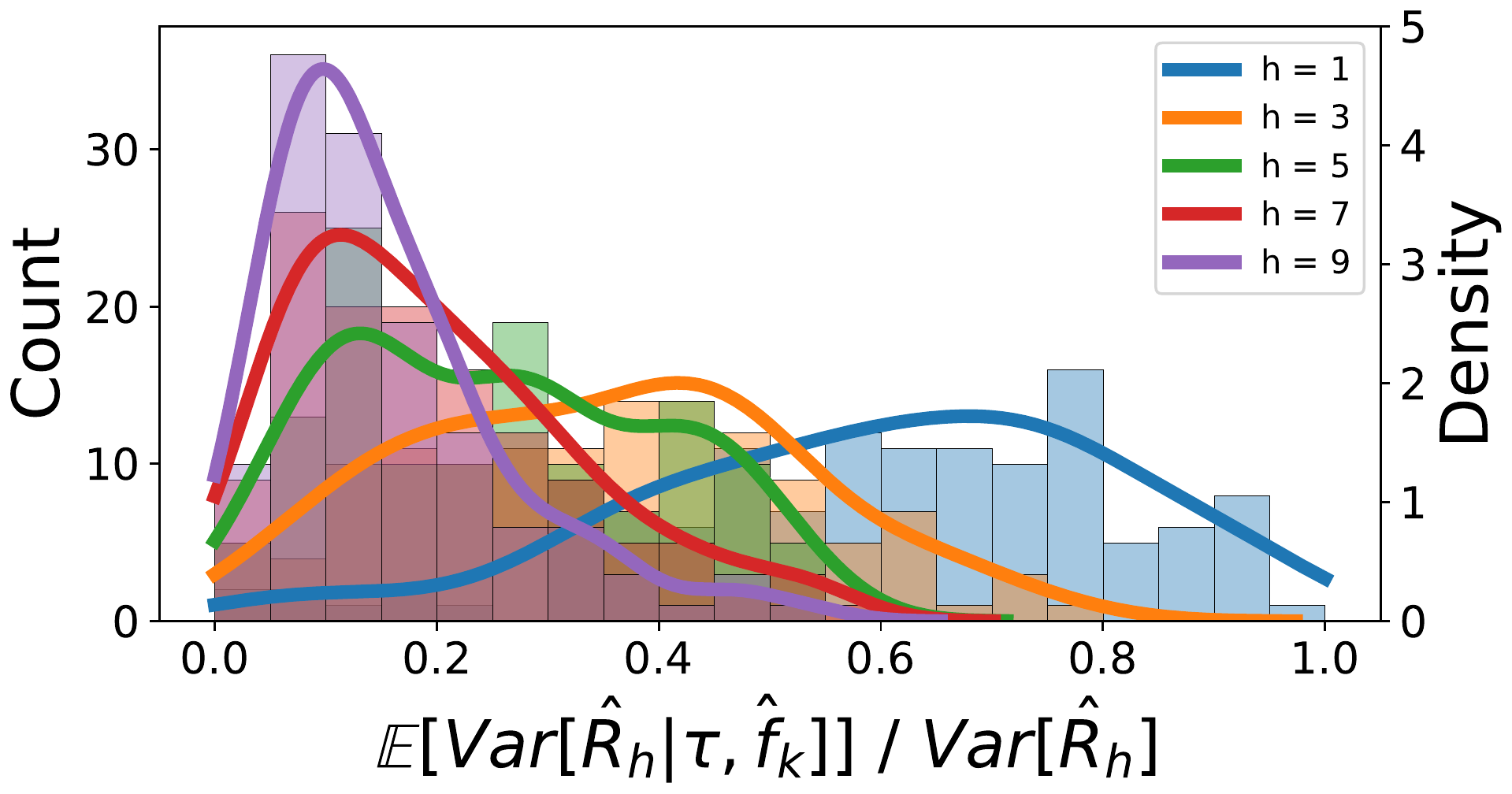}}
      \subfigure[\textit{hopper-random}] {\label{fig:A-hopper-r} \includegraphics[width=0.48\linewidth]{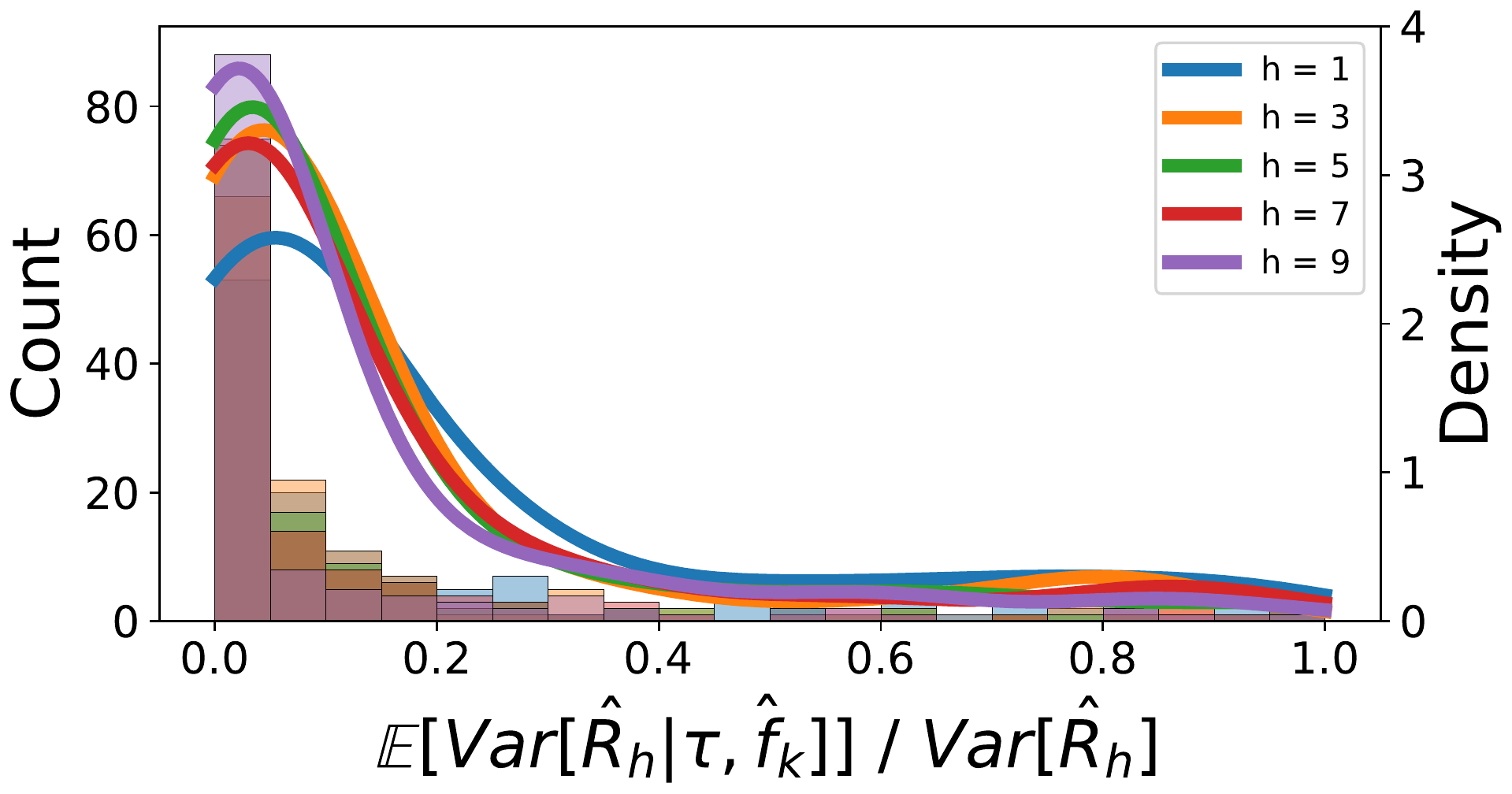}}
    \caption{
    The distribution of the ratio, $\frac{A}{A+B}=\E{\hat{f}_k}{\var{\pi_\theta}{\Rhat{h}| \tau, \hat{f}_k}} / \sigma^2_h$, from \eqref{eq:total-variance-re} when $\pi_\theta$ and $Q_\phi$ are trained with the \textit{hopper-r} dataset. (a) evaluates $\pi_\theta$ and $Q_\phi$ with $(\state{}, \action{})$ sampled from the \textit{hopper-e} dataset; (b) is the result from evaluating with the \textit{hopper-r} dataset.
    The histogram shows the empirical distribution based on a batch of samples. Probability density functions are the kernel density estimation results corresponding to each histogram with the same color.
    }
    \label{fig:a-vanish-OOD}
\end{figure}

To answer this question, recall that in the offline setting, the logged data will typically only cover a subset of the state-action space.  Hence, when we use the learned dynamics ensemble to forward sample rollout trajectories during the target value estimation procedure in \acronym, some of the trajectories will inevitably visit unseen states.  Even after the policy and the value have sufficiently converged, the rolled out trajectories will still visit OOD states (in fact, as the learned policy has shifted from the behavior policy, it is more likely that it visits more OOD states during the rollouts).  Thus, we can say that the $A$ term will not (and should not) vanish at these OOD state/actions such that \acronym~can account for the epistemic uncertainty in the value and act conservatively against it.

We have further empirically verified the relative contributions of the $A$ and $B$ terms, respectively, after the policy/value have converged.  Firstly, we considered the case when a policy and value ensemble learned with the \textit{hopper-r} dataset is used for sampling the $h$-step returns $\Rhat{h}$ starting from a set of initial states randomly selected from the \textit{hopper-e} dataset.  
Roughly speaking, this setup would ensure that we evaluate the total variance at states and actions that the policy/value have not been trained with.  Thus, we expect a relatively large amount of epistemic uncertainty still left in the $A$ term.  On the other hand, we also evaluated the learned policy/value from the states sampled from the same dataset they were trained with (i.e., \textit{hopper-r}).  In this case, we would like to see relatively little epistemic uncertainty left in $A$ since the policy and value were repeatedly trained with those states and actions.

To this end, we retrieved the policy and value ensemble checkpoints trained with the \textit{hopper-r} dataset.  Then, we calculated the proportion of $A$ with respect to the total variance, $\frac{A}{A+B}$, per each $h$-step return per each $(\state{},\action{})$ sample, which was sampled randomly from either the \textit{hopper-e} or \textit{hopper-r} dataset. 

As expected, Figure \ref{fig:A-hopper-r-eval-e} shows that there is a significant amount of variance left in the $A$ term even though we have evaluated the converged policy and value function since they were evaluated with OOD states/actions.  Especially when $h$ is small, the $A$ term contributes more to the total variance than when $h$ is large.  As $h$ increases, we can see that the weight shifts gradually towards $B$, which indicates there is more uncertainty in the model-based estimates of the returns for longer horizon rollouts.
In contrast, Figure \ref{fig:A-hopper-r} shows much less contributions from $A$ compared to $B$ even for smaller $h$.

\begin{figure}[t!]
    \centering
      \subfigure[\textit{hopper-medium}] {\label{fig:A-hopper-m} \includegraphics[width=0.48\linewidth]{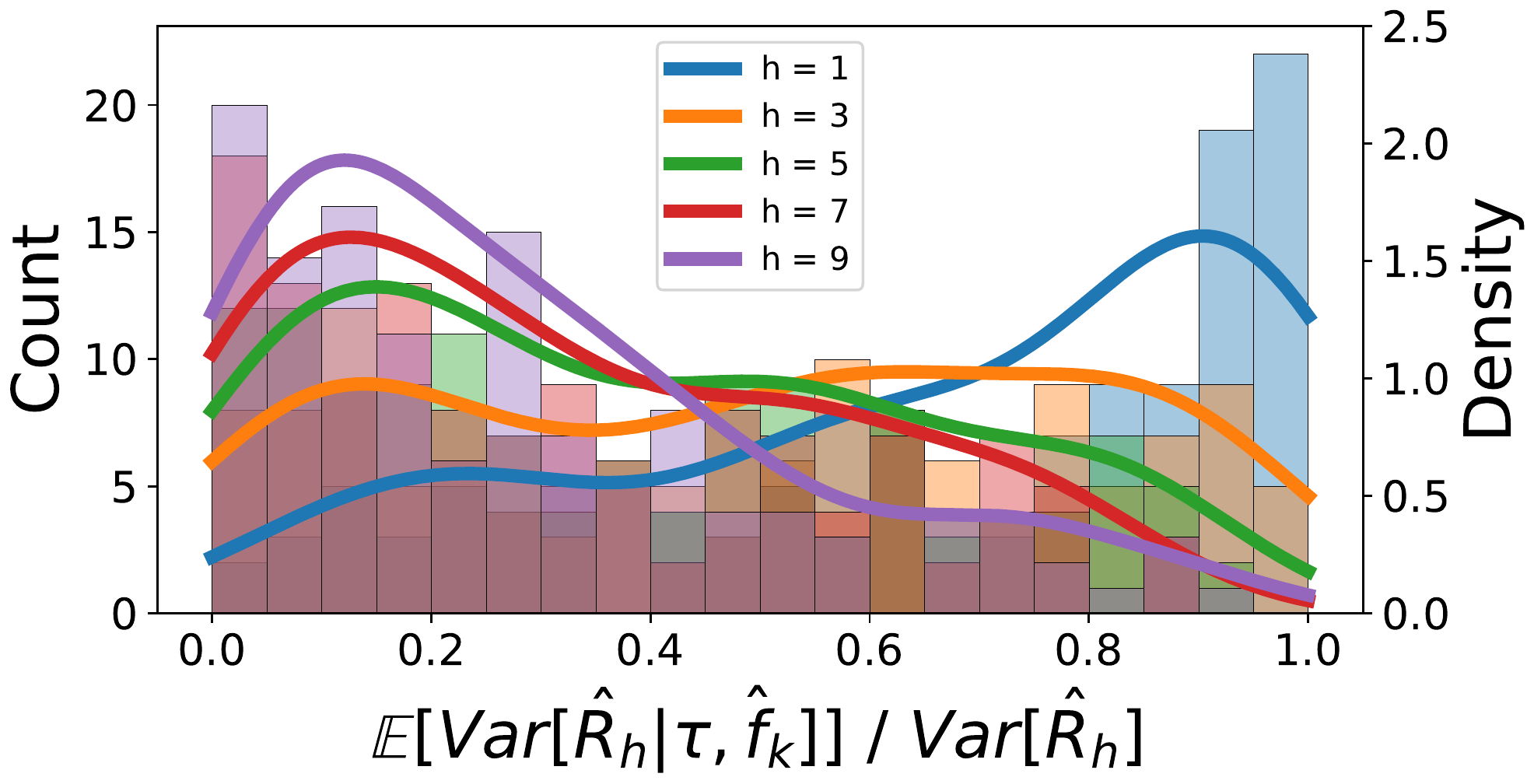}}
      \subfigure[\textit{hopper-full-replay}] {\label{fig:A-hopper-fr}  \includegraphics[width=0.48\linewidth]{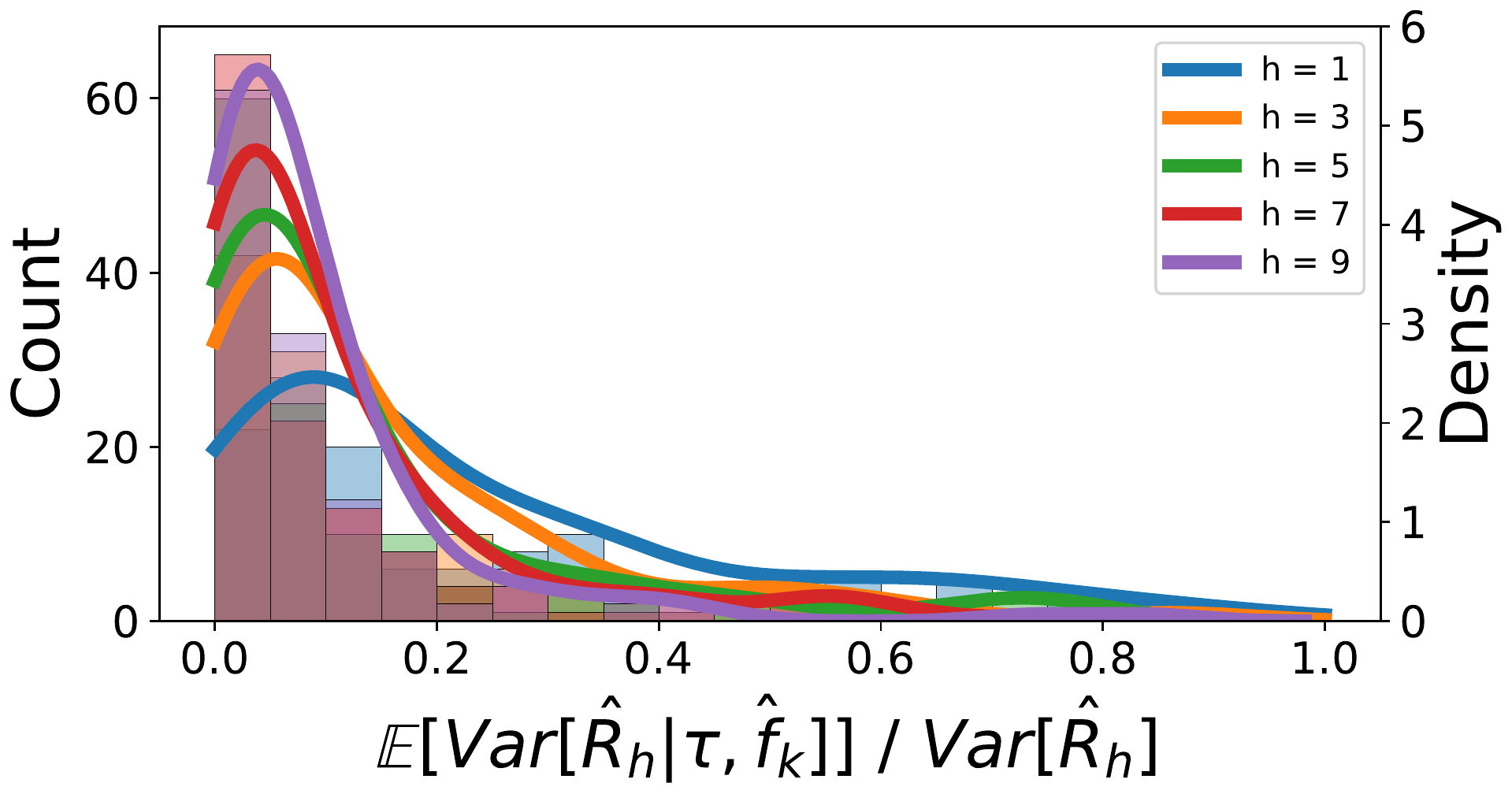}}
      \subfigure[\textit{halfcheetah-medium}] {\label{fig:A-halfcheetah-m}  \includegraphics[width=0.48\linewidth]{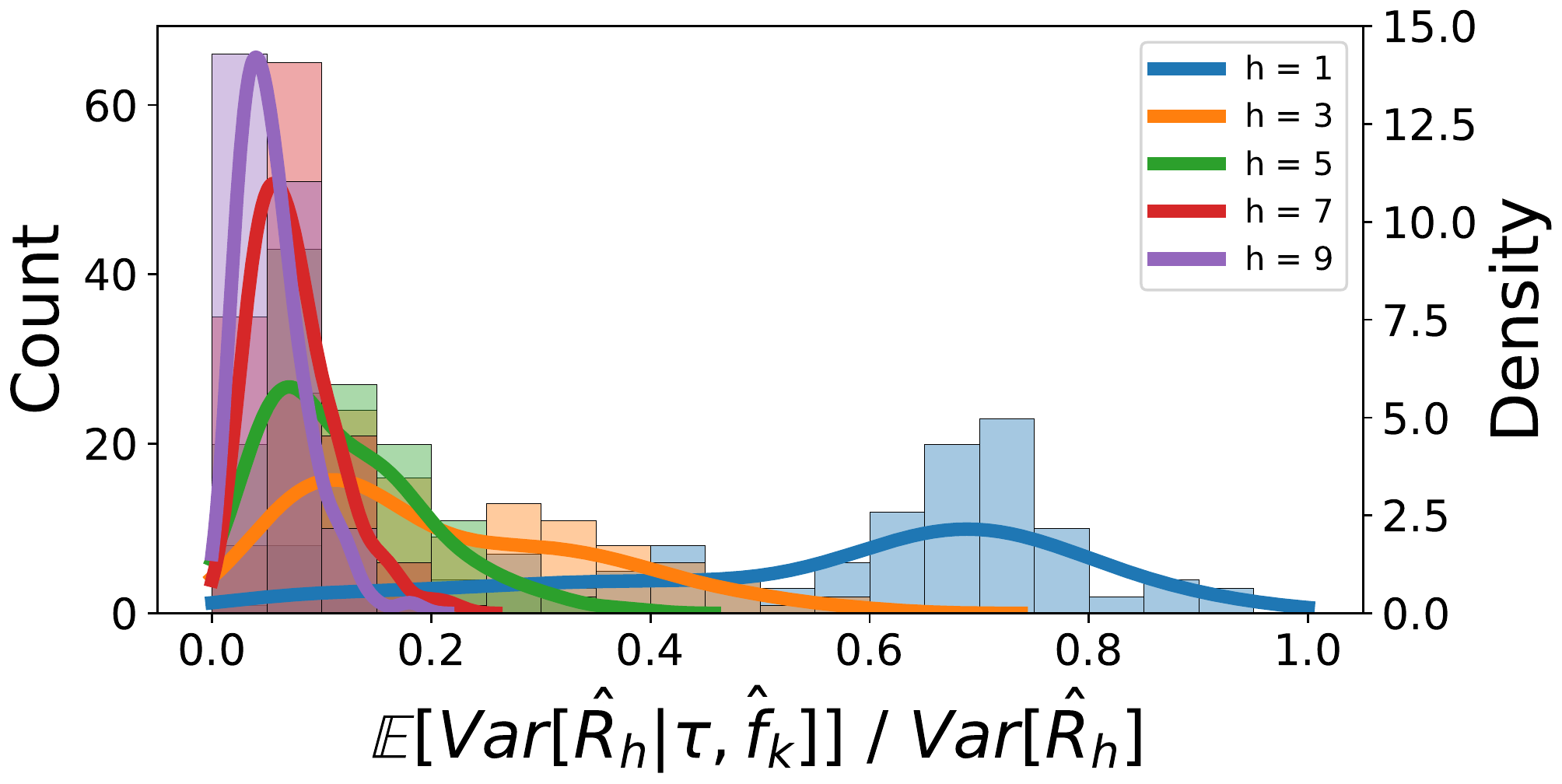}}
      \subfigure[\textit{halfcheetah-full-replay}] {\label{fig:A-halfcheetah-fr}  \includegraphics[width=0.48\linewidth]{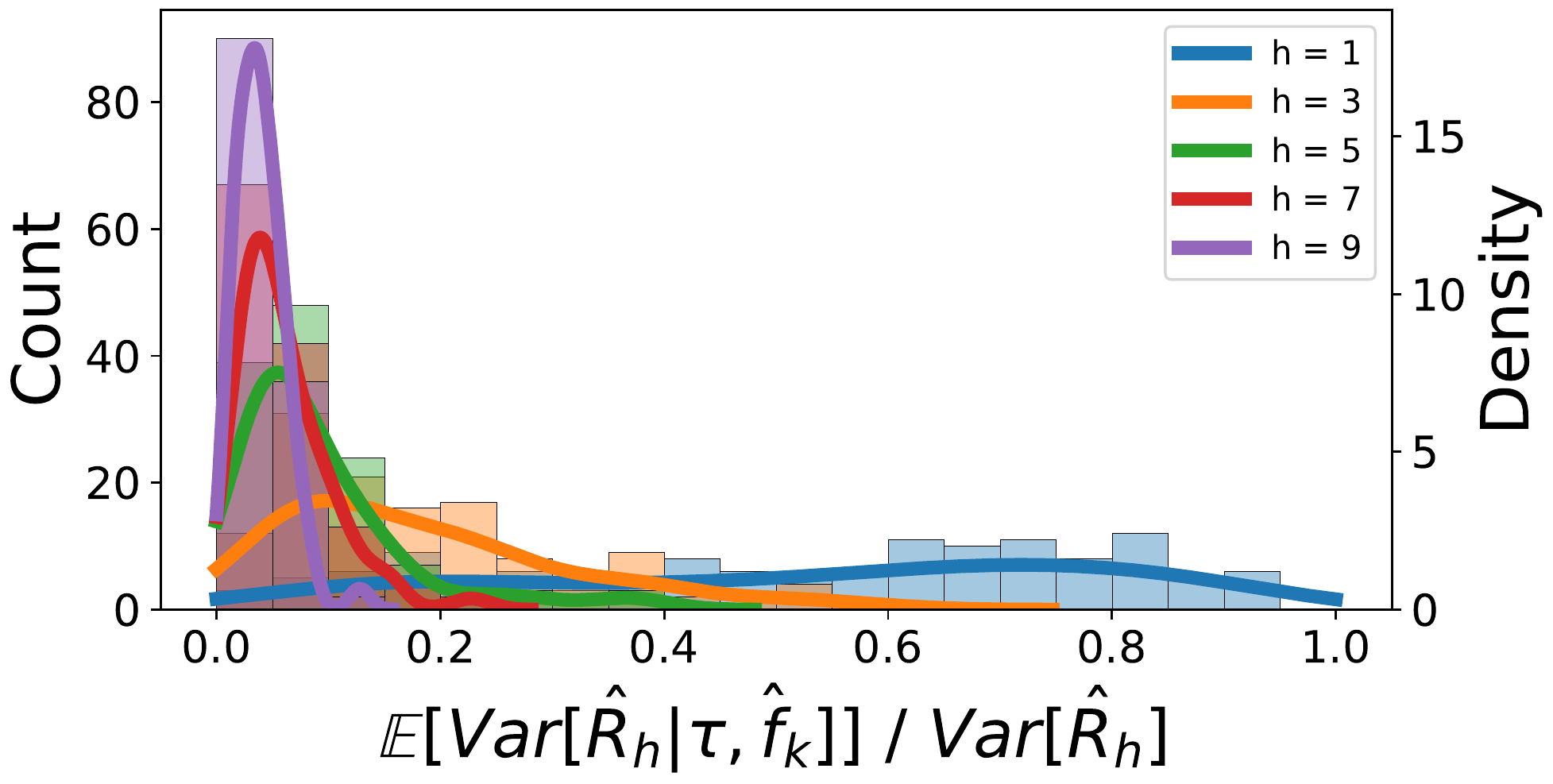}}
      \subfigure[\textit{walker2d-medium}] {\label{fig:A-walker2d-m}  \includegraphics[width=0.48\linewidth]{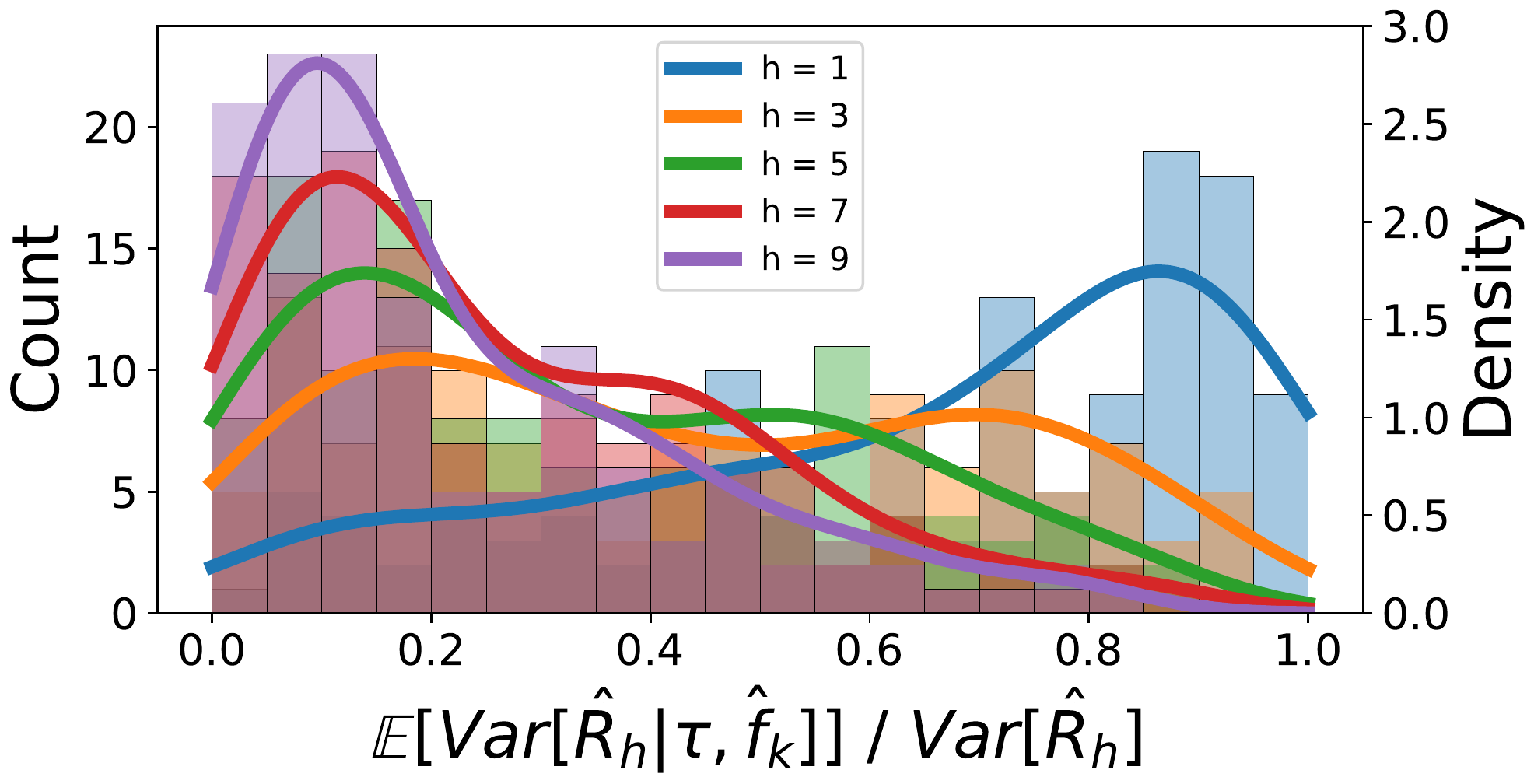}}
      \subfigure[\textit{walker2d-full-replay}] {\label{fig:A-walker2d-fr}  \includegraphics[width=0.48\linewidth]{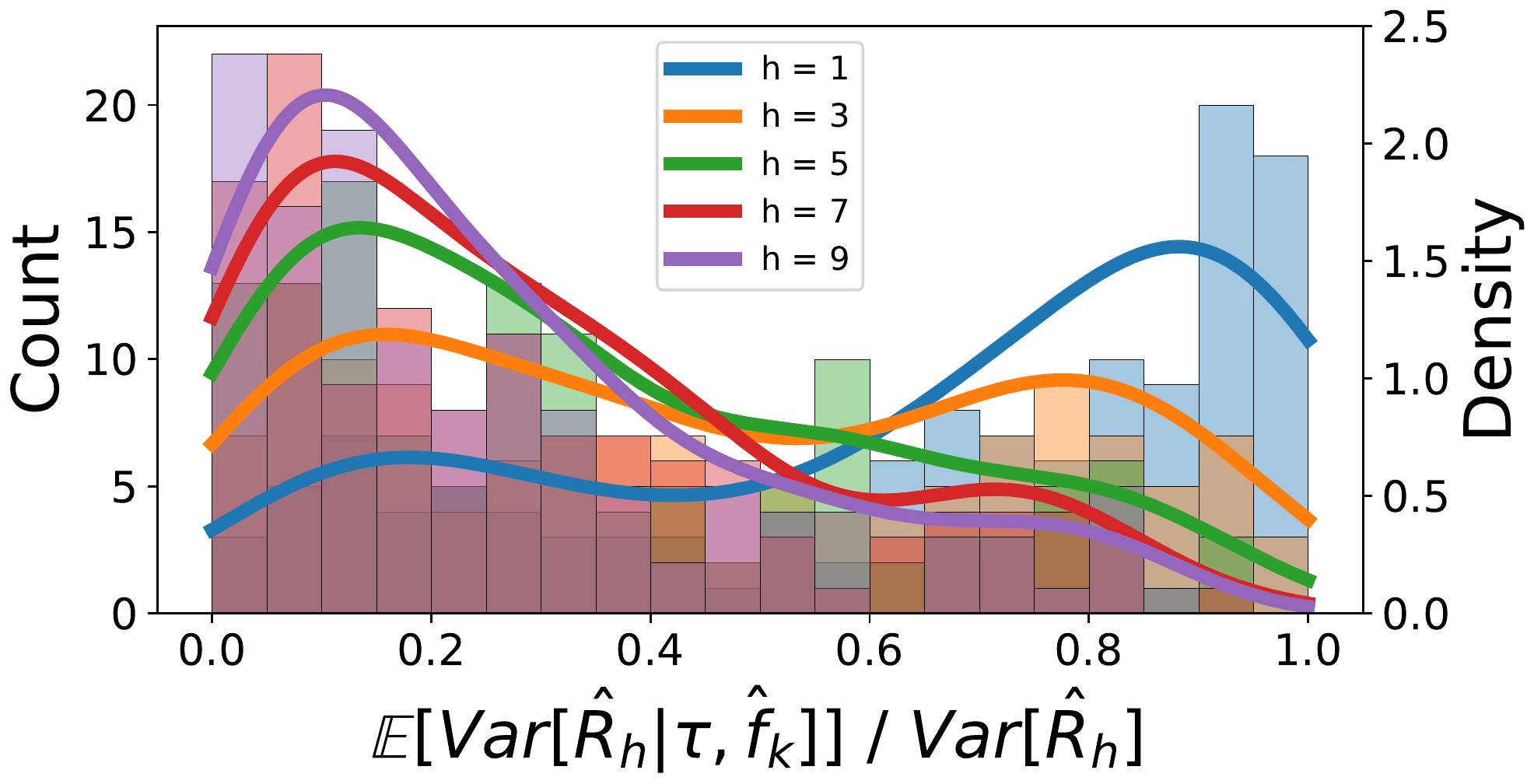}}
    \caption{The distribution of the ratio, $\frac{A}{A+B}=\E{\hat{f}_k}{\var{\pi_\theta}{\Rhat{h}| \tau, \hat{f}_k}} / \sigma^2_h$, from \eqref{eq:total-variance-re}. The histogram shows the empirical distribution based on a batch of samples. Probability density functions are the kernel density estimation results corresponding to each histogram with the same color.}
    \label{fig:a-vanish}
\end{figure}

We studied the trends from other tasks as well.  Specifically, we picked the \textit{m} and \textit{fr} D4RL configurations from the three MuJoCo environments and performed the same evaluations as discussed above. This time, the policy/value function trained with a certain dataset were evaluated with the same dataset to see if there is still a meaningful epistemic uncertainty left in $A$ term after convergence.
Figure \ref{fig:a-vanish} clearly shows that, in most of the cases, the contribution from $A$ to the total variance is not negligible, despite the policy/value being already converged. Similar to the \textit{hopper-r} case, $A$ generally contributes more than $B$ does for small $h$ values. 
As discussed, this is an intuitive result since the learned model would typically be quite accurate for single-step predictions, hence smaller $B$ compared to $A$.

It is also notable that in the \textit{fr} tasks of the \textit{hopper} and \textit{halfcheetah} environments shown in Figure \ref{fig:A-halfcheetah-fr} and \ref{fig:A-hopper-fr}, much more contribution is coming from $B$ even for small $h$ (however, $A$ still has noticeable contribution). 
Note that (1) the \textit{fr} (full-replay) dataset was curated such that it covers all transition samples encountered by various policies, starting from a random policy all the way to an expert policy. Now, also note that (2) since we pre-train the dynamics model and fix it during policy training, the epistemic uncertainty baked in the dynamics ensemble is kept fixed, whereas the uncertainty in the value ensemble can diminish as training continues.  These two factors combined can explain why we would see more contributions in the total variance from $B$ rather than $A$ in the \textit{fr} datasets.

\subsection{Ablations} \label{appendix:ablations}

In this part, we provide additional ablations that complement the results presented in the main text. 

\paragraph{The effectiveness of conservatism via LCB compared to MAP} STEVE \citep{buckman2018steve} introduced an adaptive weighting scheme for MVE, which corresponds to the MAP estimation of the posterior we get in \eqref{eq:posterior}.  In this part, we provide the complete ablations comparing \acronym~and STEVE in all tasks.

\begin{figure}[t!]
    \centering
      \subfigure[halfcheetah-v2] {\label{fig:steve-cheetah} \includegraphics[width=0.45\linewidth]{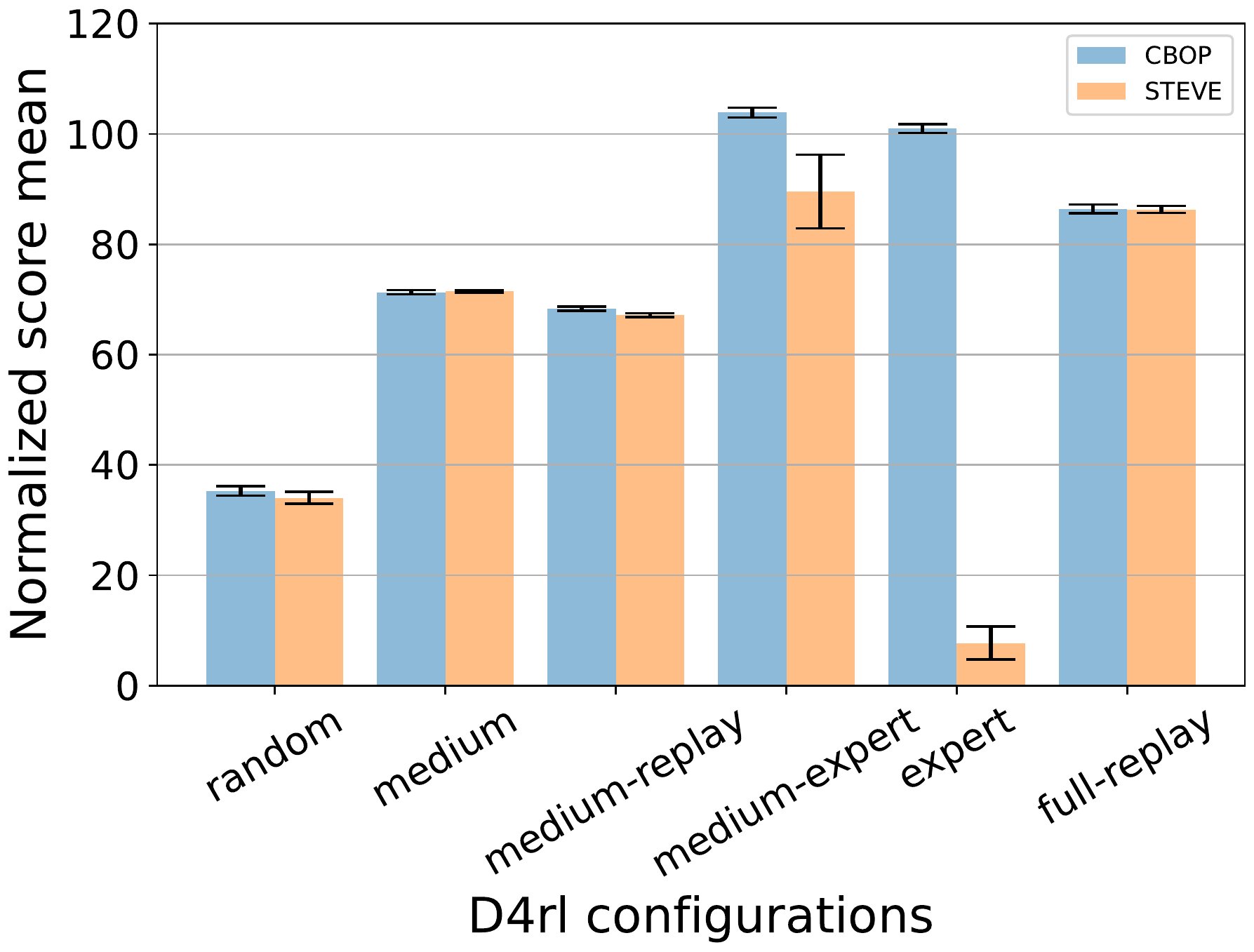}}
      \hspace{5pt}
      \subfigure[hopper-v2] {\label{fig:steve-hopper}  \includegraphics[width=0.45\linewidth]{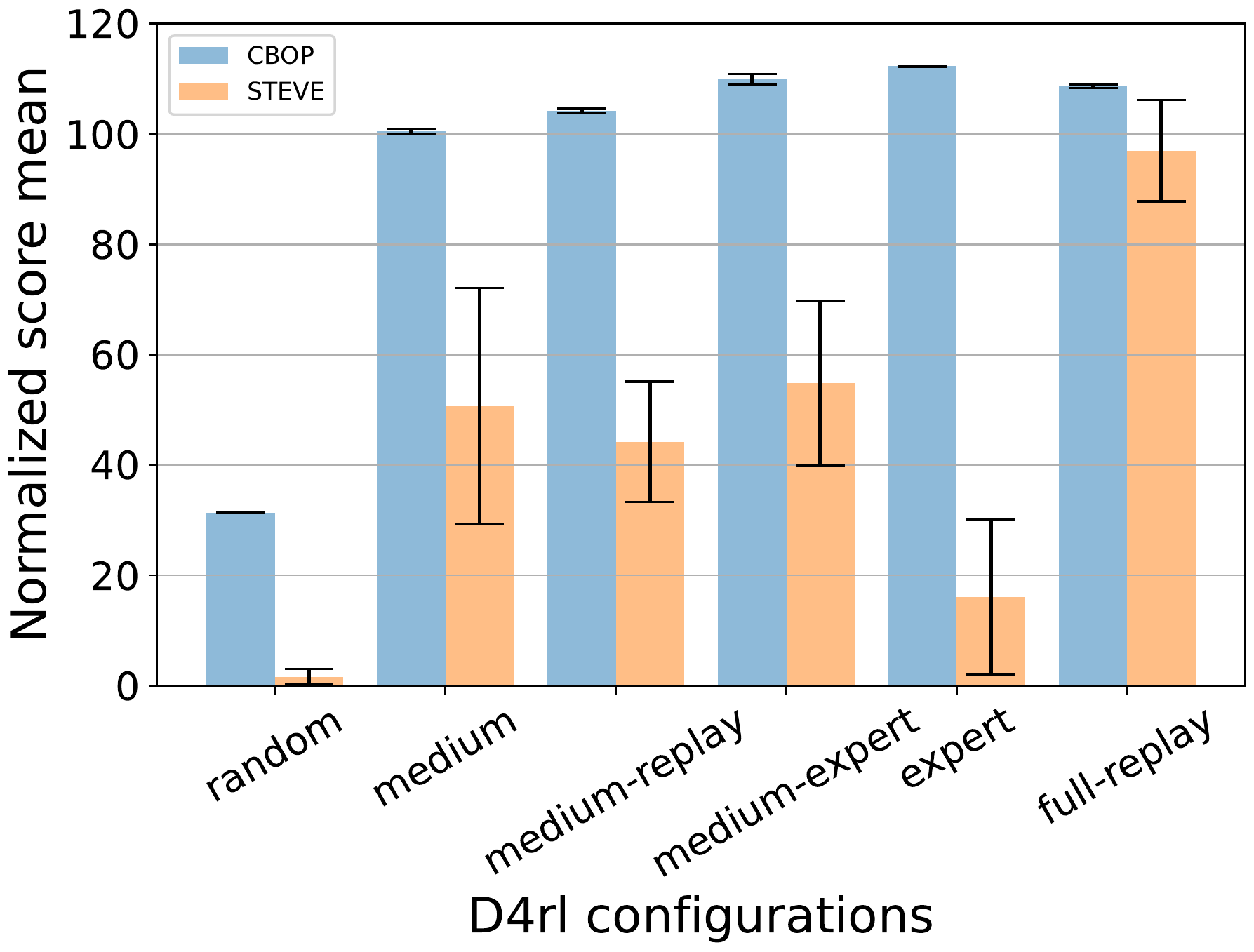}}
      \hspace{5pt}
      \subfigure[walker2d-v2] {\label{fig:steve-walker}  \includegraphics[width=0.45\linewidth]{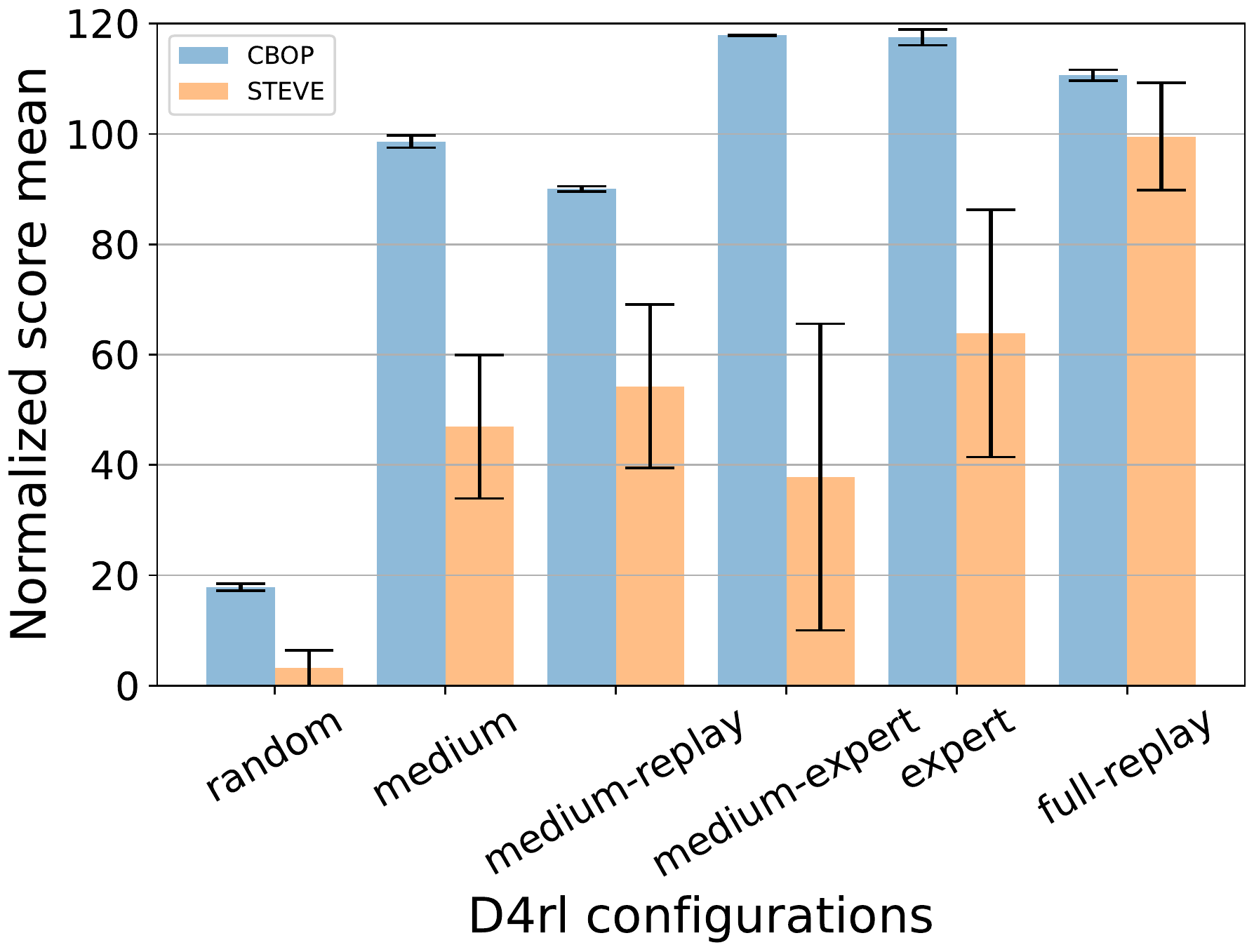}}
    \caption{Comparison of the MAP estimation and the LCB estimation in the D4RL MuJoCo benchmark tasks. Experiments are run with 3 random seeds.}
    \label{fig:ablation-steve}
\end{figure}

In Figure \ref{fig:steve-cheetah}, we see that STEVE performs comparably to \acronym~in $4$ of the $6$ tasks, where small $\psi$ have been used in \acronym~(Table \ref{table:hyperparameters}). However, for the \textit{medium-expert} and \textit{expert} tasks --- where we have used $\psi=3$ and $5$, respectively --- \acronym~outperforms STEVE. 

The differences in the performances are even more striking in the other two environments. Figure \ref{fig:steve-hopper} and \ref{fig:steve-walker} show that \acronym~significantly outperforms STEVE, suggesting that conservatism plays a crucial role. It is worth reasserting that the original adaptive weighting scheme derived in STEVE does not lend itself to a conservative value estimation as we can do with \acronym.  

\paragraph{The effectiveness of the Bayesian weighting scheme}

\begin{figure}[t!]
    \centering
      \subfigure[\textit{halfcheetah-random}] {\label{fig:fixed-hc-r} \includegraphics[width=0.4\linewidth]{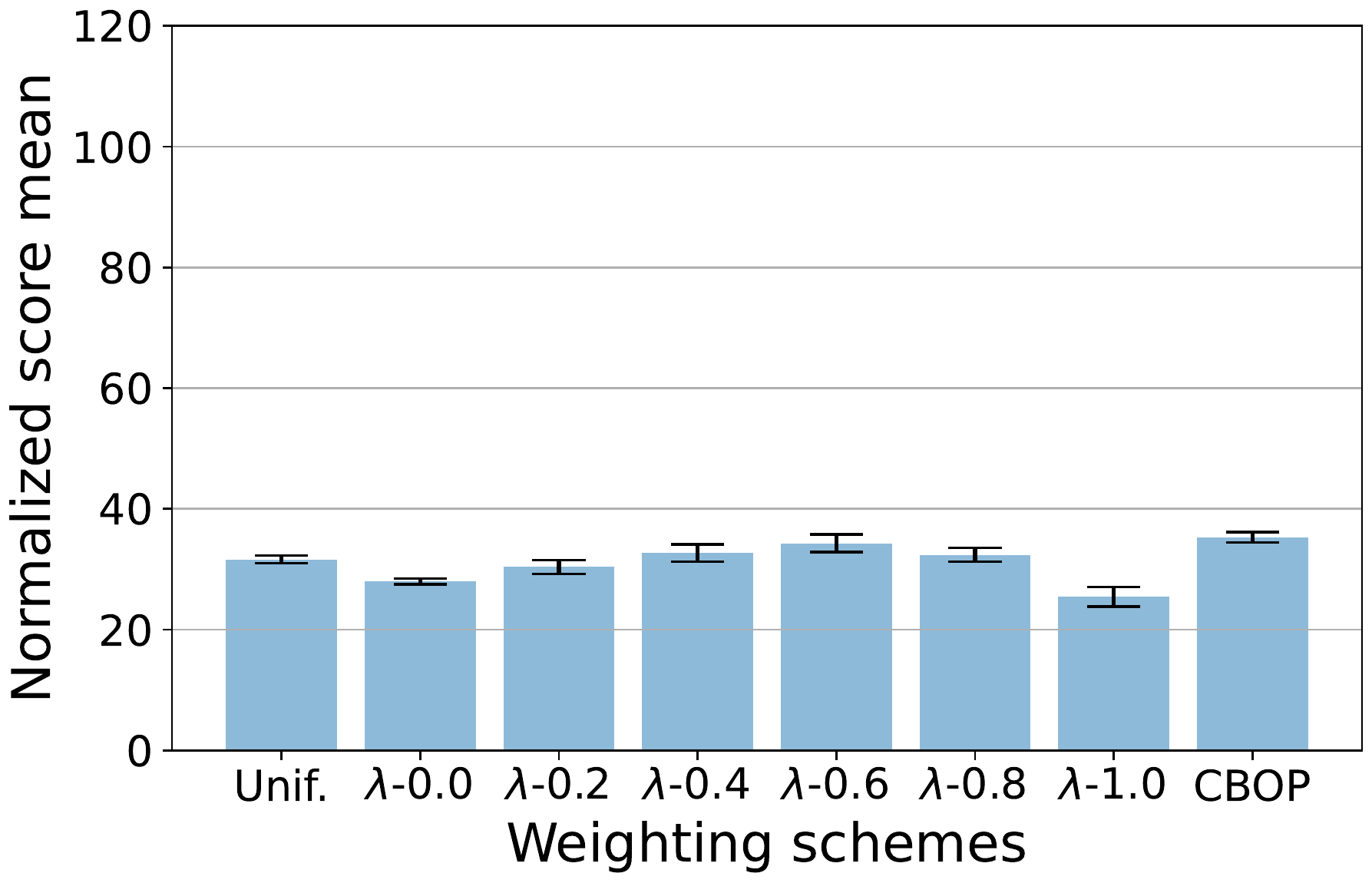}}
      \subfigure[\textit{halfcheetah-medium}] {\label{fig:fixed-hc-m}  \includegraphics[width=0.4\linewidth]{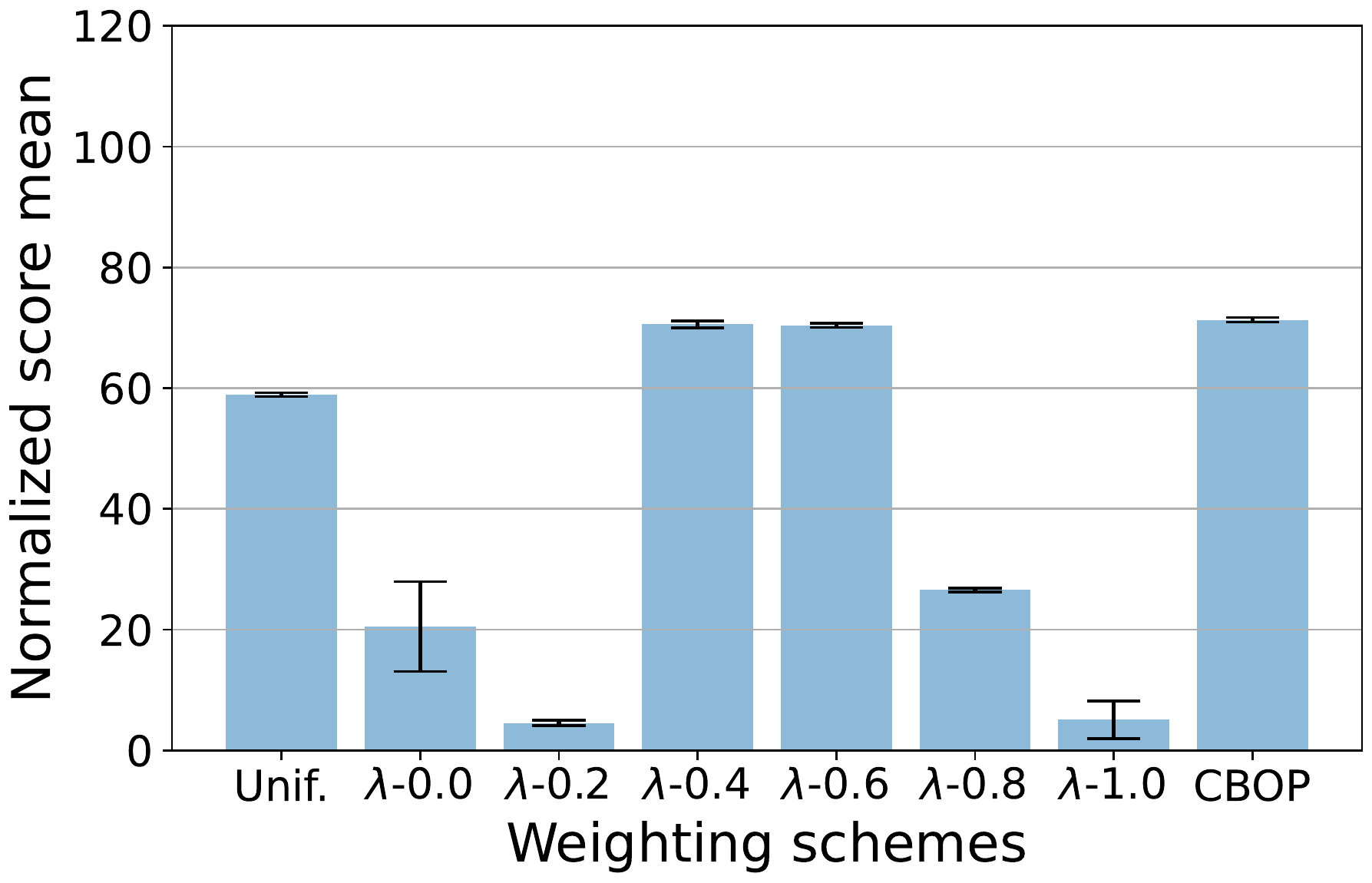}}
      \subfigure[\textit{halfcheetah-medium-replay}] {\label{fig:fixed-hc-mr}  \includegraphics[width=0.4\linewidth]{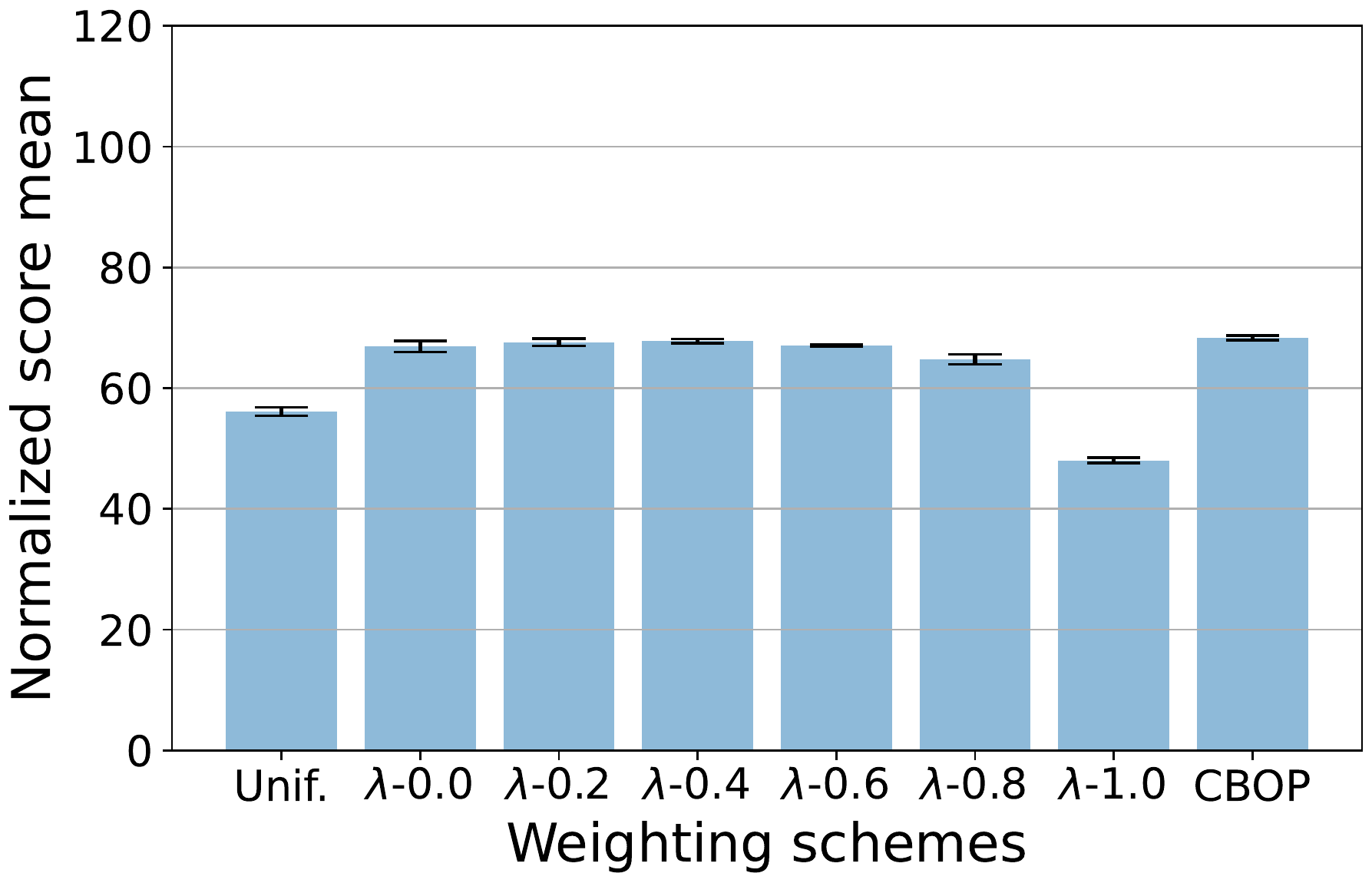}}
      \subfigure[\textit{halfcheetah-medium-expert}] {\label{fig:fixed-hc-me} \includegraphics[width=0.4\linewidth]{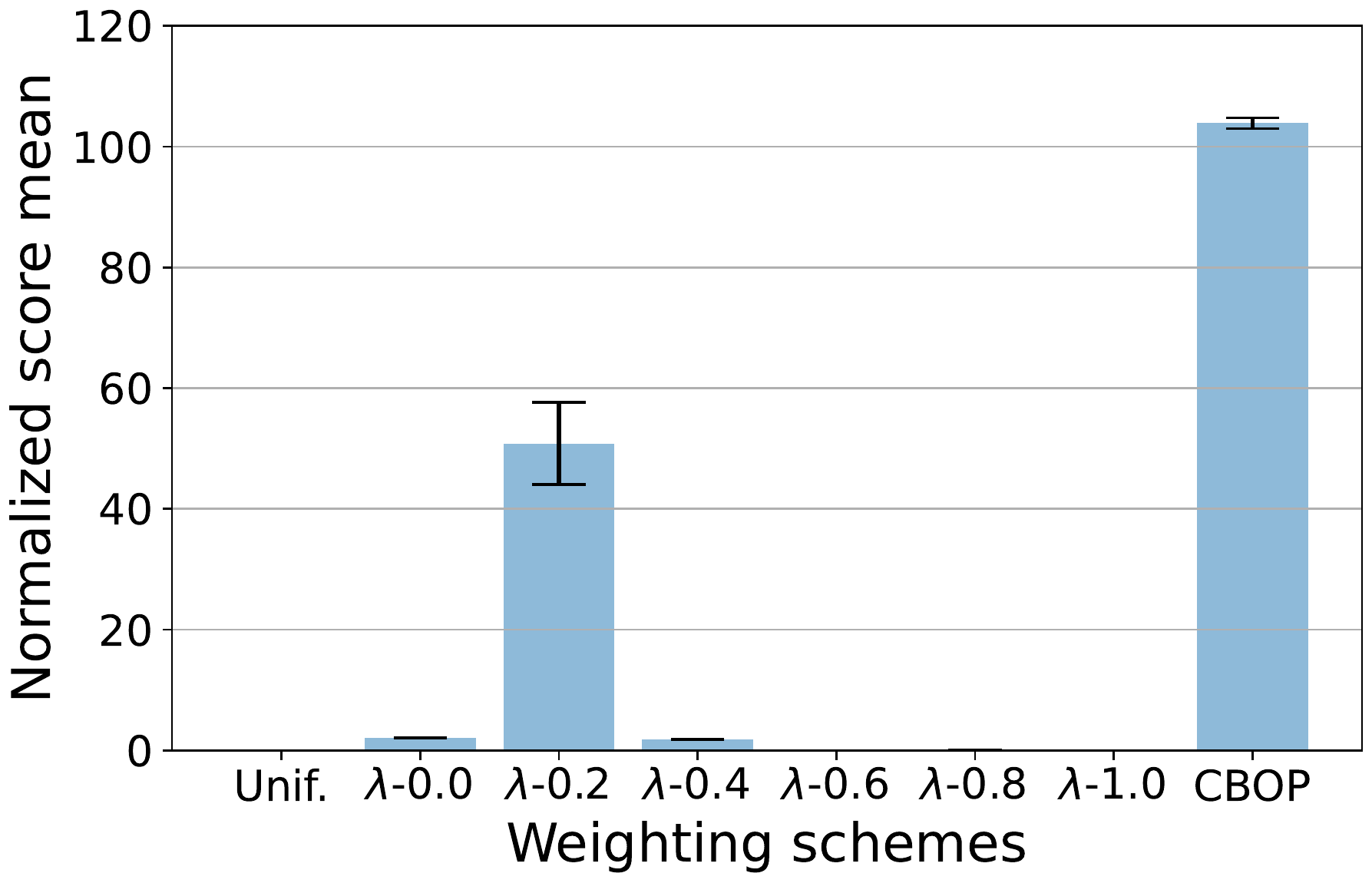}}
    \caption{Comparing the fixed weighting schemes and \acronym~on the \textit{halfcheetah} environment. Experiments are run with 3 seeds.}
    \label{fig:ablation-lambda-cheetah}
\end{figure}

\begin{figure}[tbh!]
    \centering
      \subfigure[\textit{hopper-random}] {\label{fig:fixed-h-r} \includegraphics[width=0.4\linewidth]{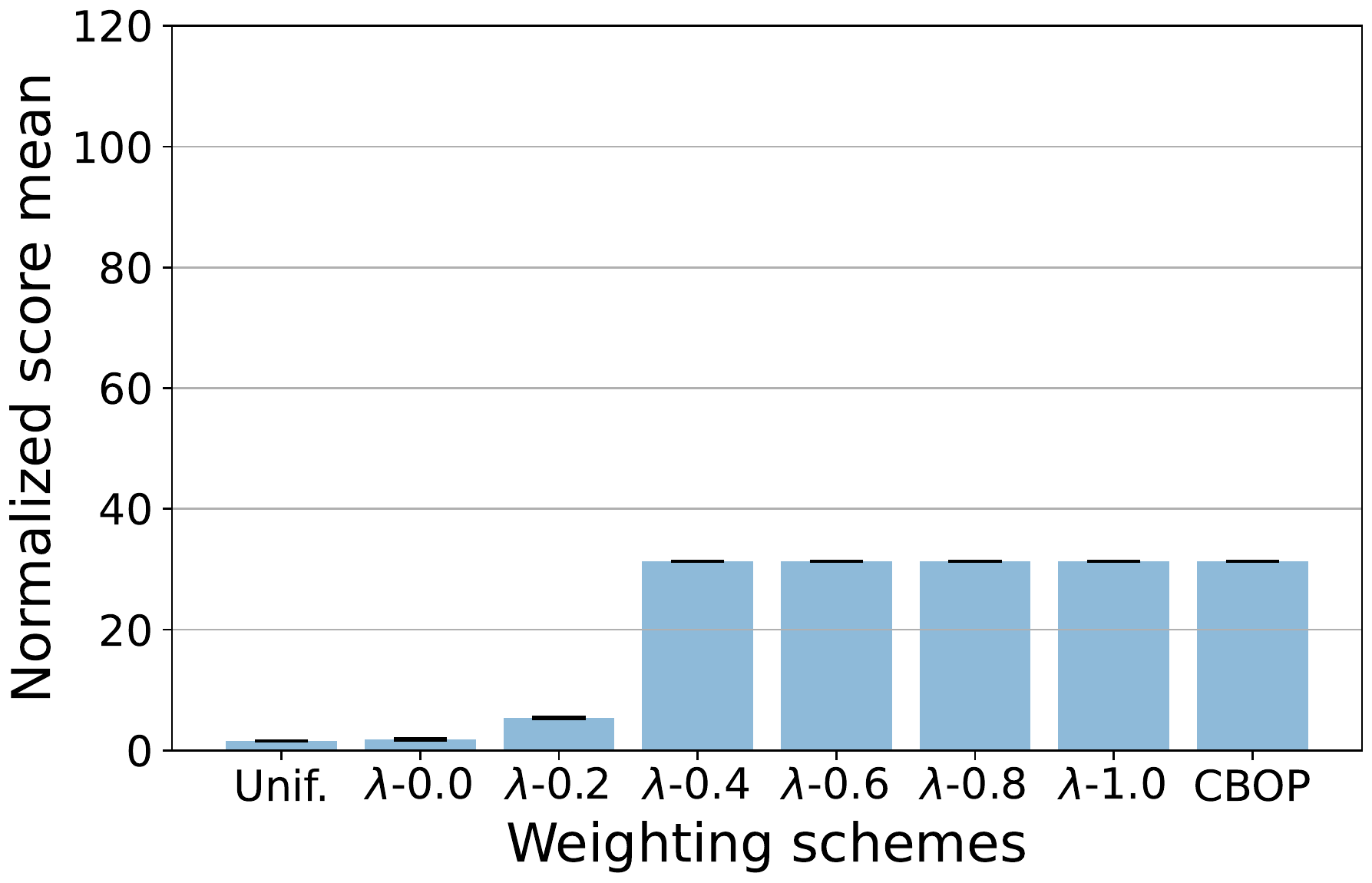}}
      \subfigure[\textit{hopper-medium}] {\label{fig:fixed-h-m}  \includegraphics[width=0.4\linewidth]{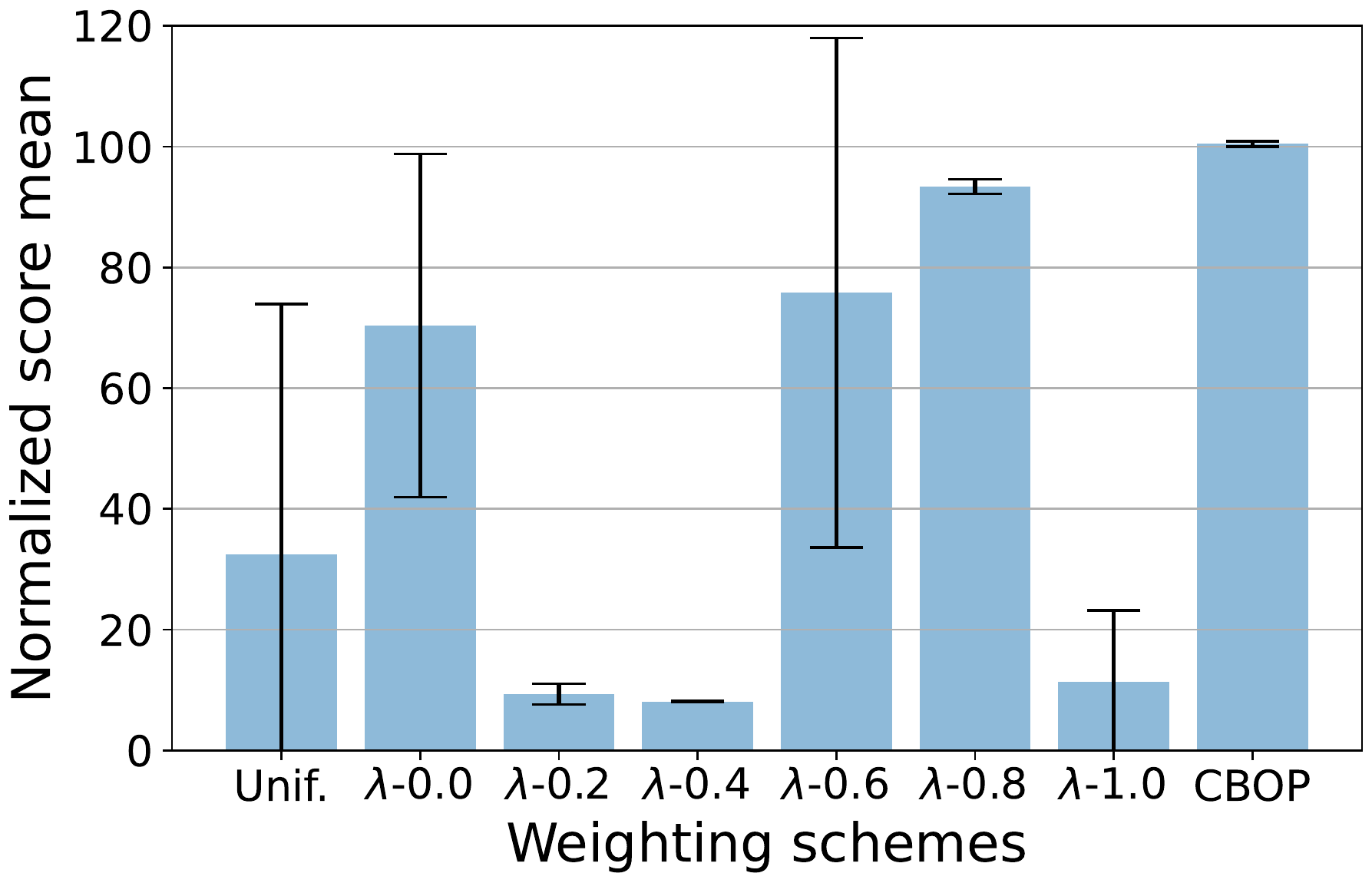}}
      \subfigure[\textit{hopper-medium-replay}] {\label{fig:fixed-h-mr}  \includegraphics[width=0.4\linewidth]{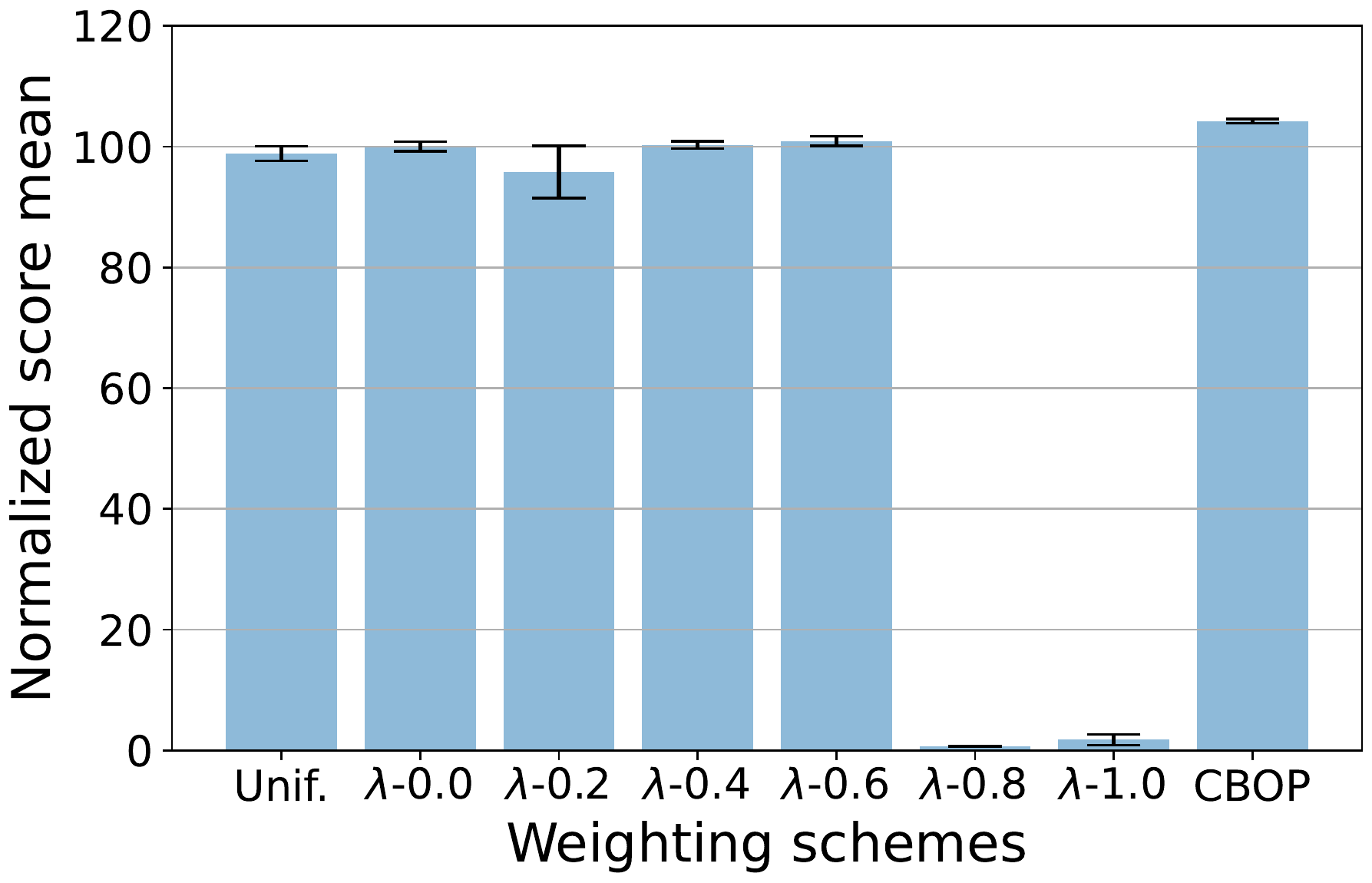}}
      \subfigure[\textit{hopper-medium-expert}] {\label{fig:fixed-h-me} \includegraphics[width=0.4\linewidth]{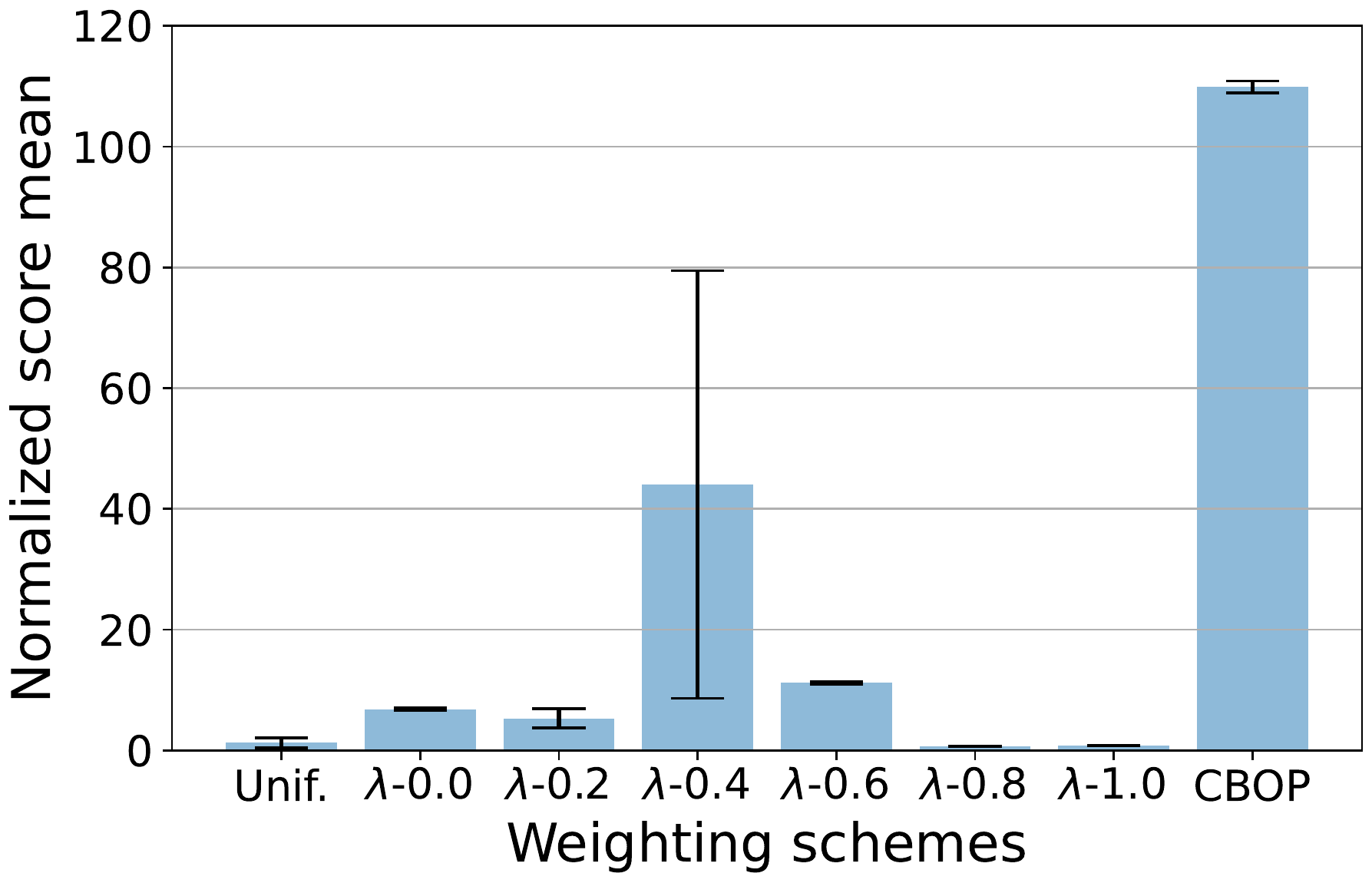}}
    \caption{Comparing the fixed weighting schemes and \acronym~on the \textit{hopper} environment. Experiments are run with 3 seeds.}
    \label{fig:ablation-lambda-hopper}
\end{figure}

In Section \ref{sec:experiments}, we have presented a part of the ablations comparing the adaptive weighting scheme of \acronym~with the fixed weighting scheme, i.e., uniform and $\lambda$ weighting. 
The weights in the uniform weighting correspond to $w_h=\frac{1}{H+1}$, while those in the $\lambda$-weighting are $w_h=\frac{1-\lambda}{1-\lambda^{H+1}}\lambda^h$.  
In the latter, the larger the $\lambda$ parameter, the more weight is allocated to longer-horizon model-based rollouts; $\lambda=1$ corresponds to solely using the $H$-step MVE target, whereas $\lambda=0$ bootstraps immediately at $\state{}'$ as in the model-free case.

In order to better isolate the impact of the different weighting schemes, we have used the conservative value estimation for these two fixed weighting schemes as well. More concretely, we have sampled $M\times K$ $\Rhat{h}$ samples for $h=0,\dots, H$ and computed the weighted sums ($\sum_{h=0}^{H}w_h\Rhat{h}$) to get $MK$ samples of target values. With these samples, we have computed the empirical mean and the variance, from which we have taken the LCB $\mu-\psi \cdot \sigma$ as the target values. 

\begin{figure}[t!]
    \centering
      \subfigure[\textit{walker2d-random}] {\label{fig:fixed-w-r} \includegraphics[width=0.4\linewidth]{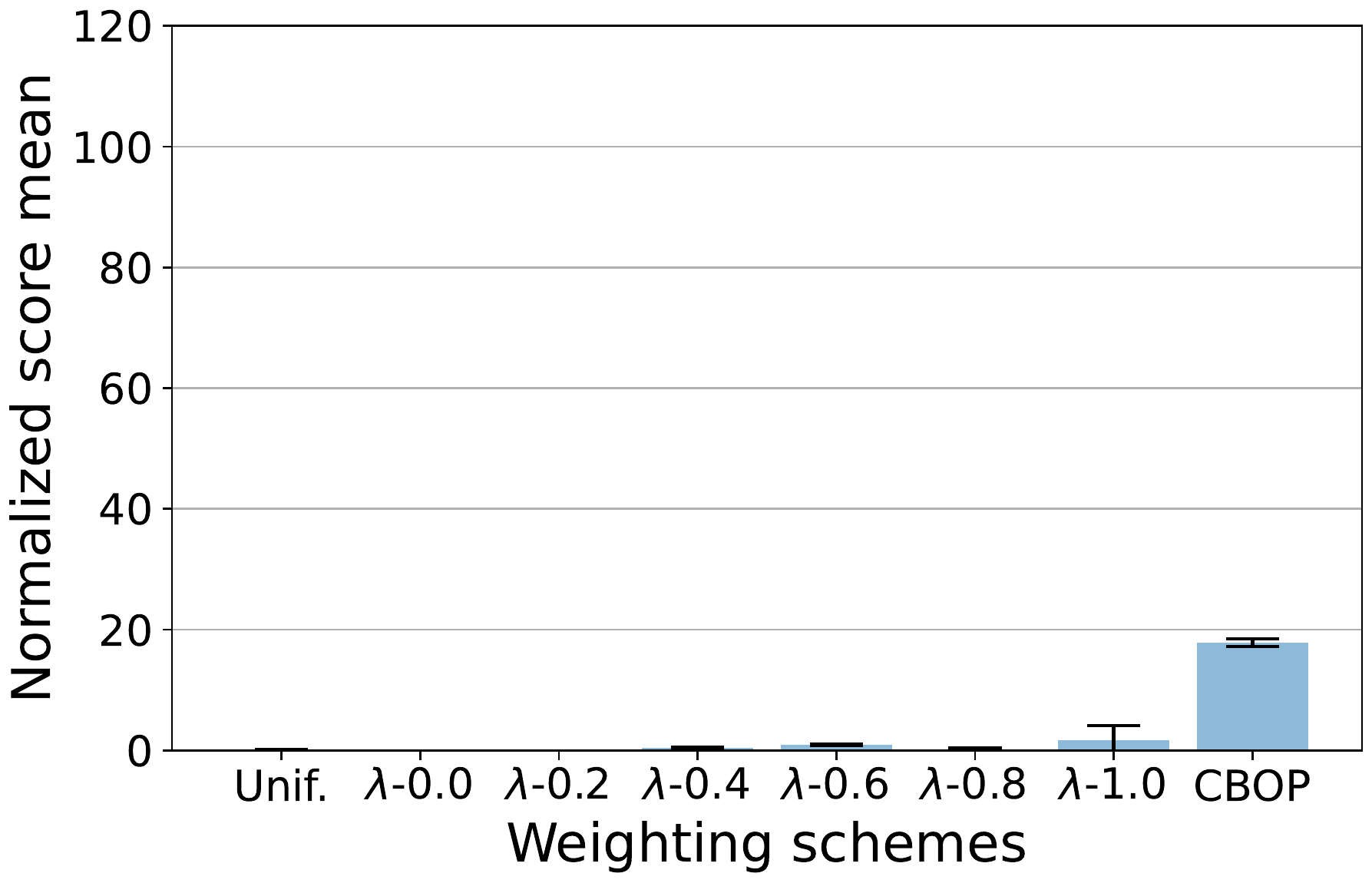}}
      \subfigure[\textit{walker2d-medium}] {\label{fig:fixed-w-m}  \includegraphics[width=0.4\linewidth]{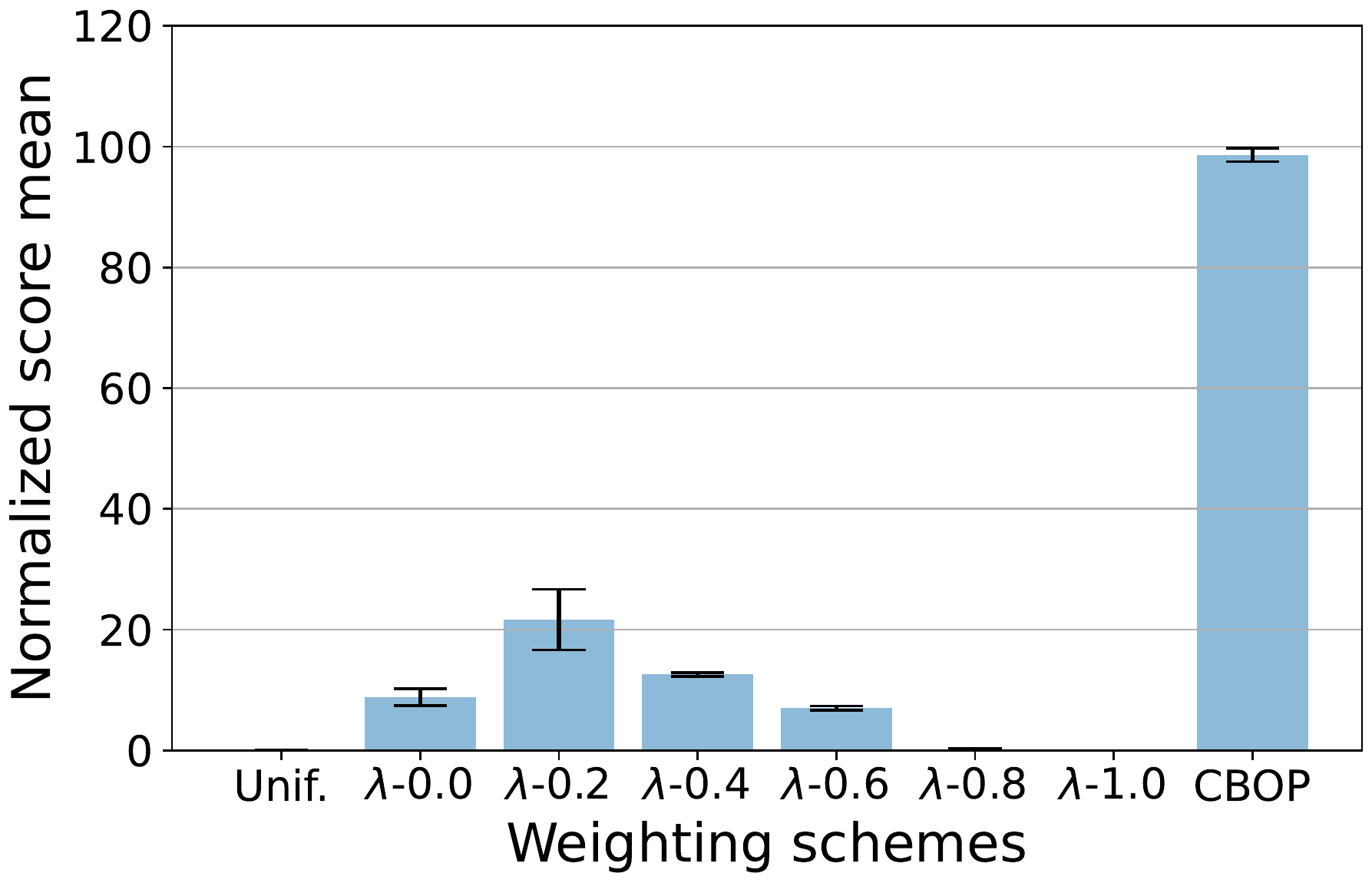}}
      \subfigure[\textit{walker2d-medium-replay}] {\label{fig:fixed-w-mr}  \includegraphics[width=0.4\linewidth]{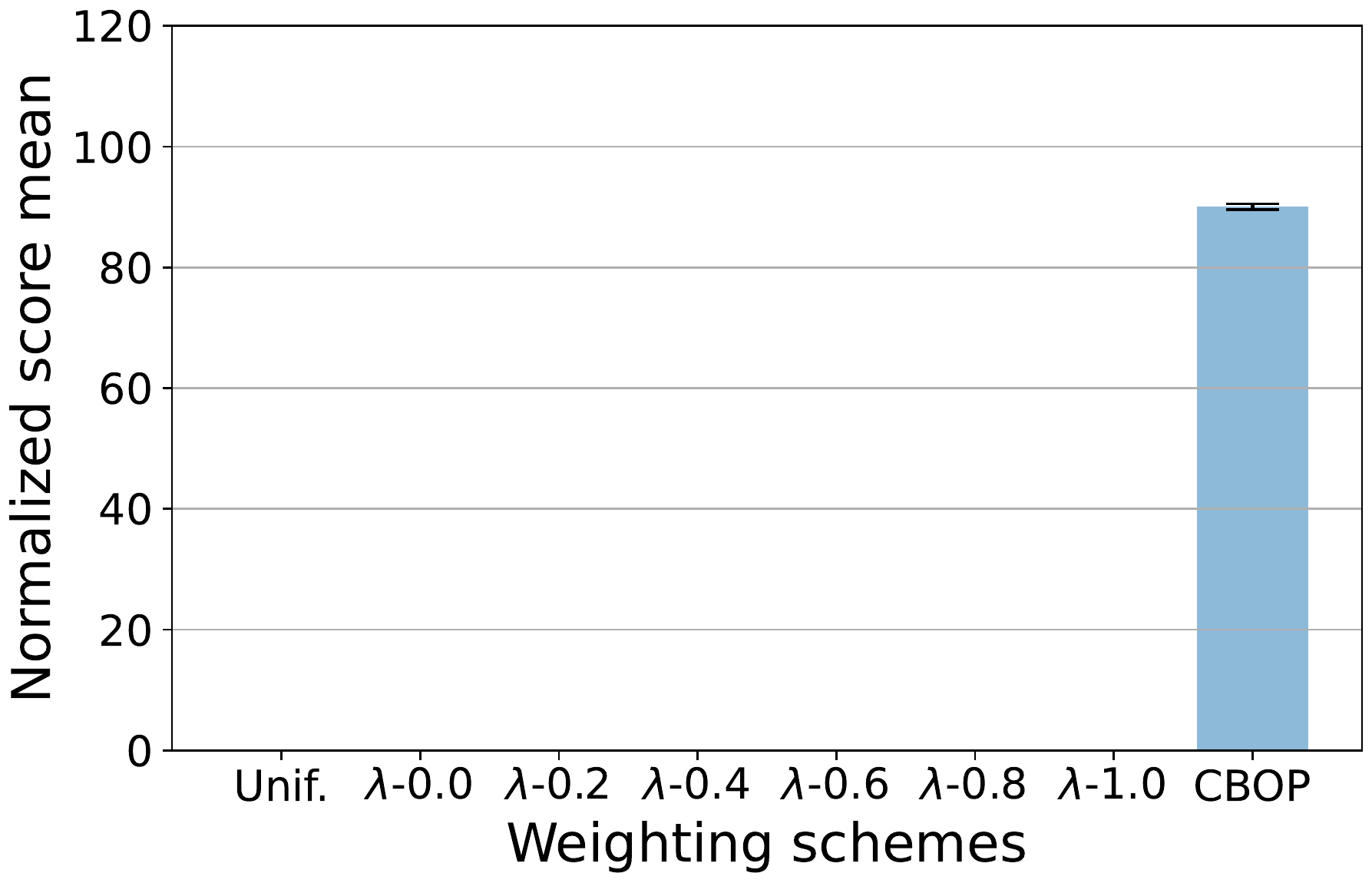}}
      \subfigure[\textit{walker2d-medium-expert}] {\label{fig:fixed-w-me} \includegraphics[width=0.4\linewidth]{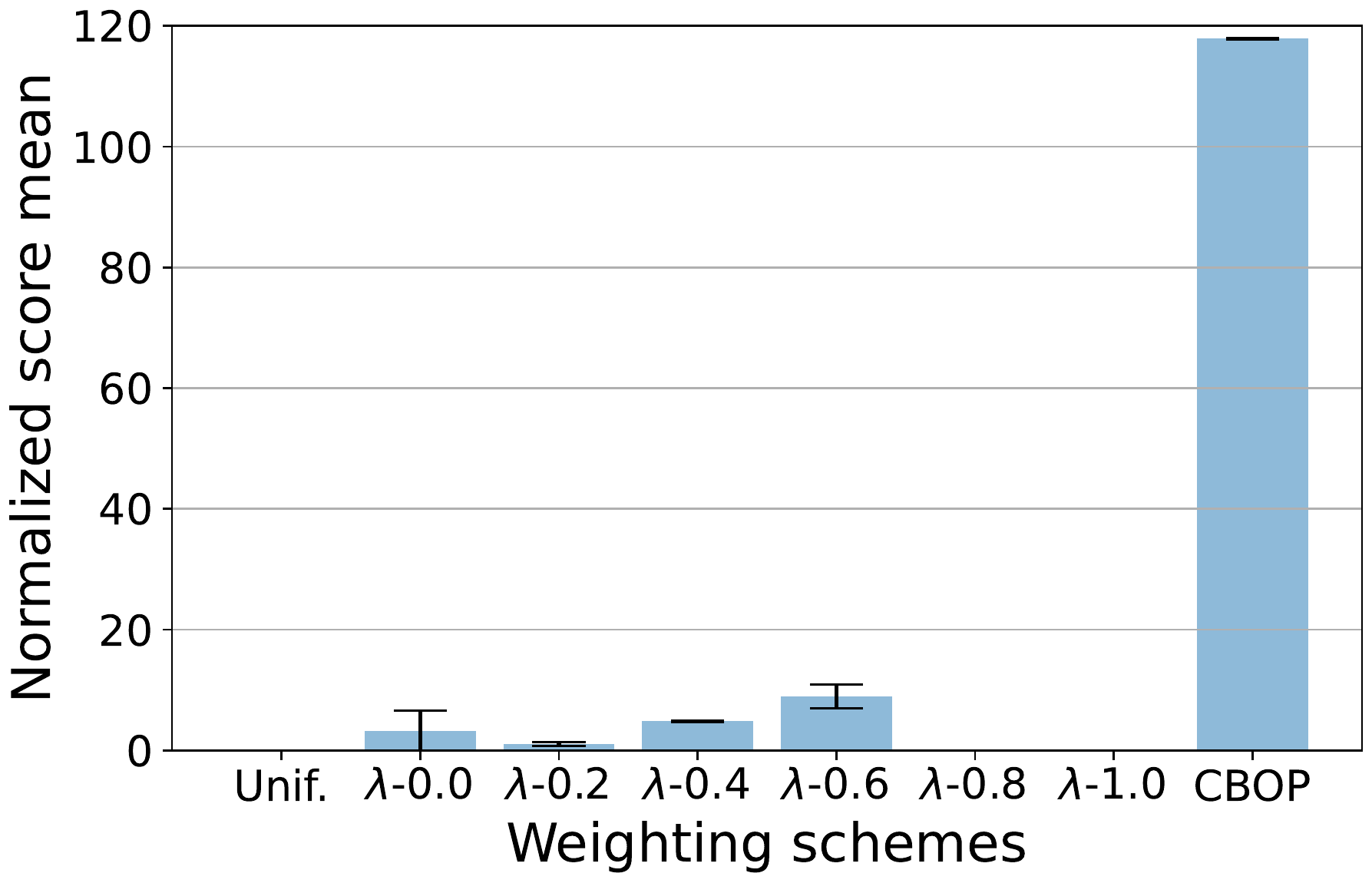}}
    \caption{Comparing the fixed weighting schemes and \acronym~on the \textit{walker2d} environment. Experiments are run with 3 seeds.}
    \label{fig:ablation-lambda-walker}
\end{figure}

Figure \ref{fig:ablation-lambda-cheetah} - \ref{fig:ablation-lambda-walker} show the results on the \textit{halfcheetah}, \textit{hopper}, and \textit{walker2d} environments, respectively. We have found that the fixed weighting does not work in the \textit{walker2d} tasks, regardless of the $\lambda$ values. Also, \acronym~has significantly outperformed the fixed weighting schemes in narrow datasets (i.e., \textit{medium-expert}) across all environments. 

In some tasks --- such as \textit{medium} and \textit{medium-replay} tasks in \textit{hopper} and \textit{halfcheetah} environments, there are some $\lambda$ values that can show similar performances as \acronym. However, large fluctuations across different $\lambda$ values as exhibited in \textit{halfcheetah-medium} and \textit{hopper-medium} suggest that finding $\lambda$ that works robustly across all tasks may be impossible.  On the contrary, the adaptive Bayesian weighting scheme of \acronym~can work reliably across all tasks considered.

\paragraph{Additional Baseline: quantile-based conservative MVE}

We have seen that \acronym~is able to adaptively regulate the reliance on model-based and model-free value estimates while acting conservatively with respect to both.  The uncertainties in the learned dynamics model and the value function are captured through the sampling procedure we detailed in Section \ref{sec:ensemble-sampling}. 
The ablation studies presented in Section \ref{subsec:experiments-ablations} show the strong merits that the Bayesian interpretation provides us through the adaptive control of the roll-out horizon and the conservative value estimates from the Bayesian posterior. 
Here, we further strengthen the case 
and ablate the benefits of being Bayesian by comparing \acronym~against another baseline that we dub \textit{Distributional MVE (DiMVE)}.

Instead of forming a Bayesian posterior over $\hat{Q}^\pi$, DiMVE simply aggregates all $MKH$ return samples that we collect from a single pass of forward sampling. Then, it performs a quantile-based conservative value estimation. Formally, let
\begin{equation*}
    \Rhat{h}^{m,k}(\state{},\action{},\state{}') := \sum_{t=0}^{h}\gamma^t \rhat{t}(\mstate{t}^{(k)}, \maction{t}^{(k)}) + \gamma^{h+1} Q_{\phi'}^m(\mstate{h+1}^{(k)}, \maction{h+1}^{(k)})
\end{equation*}
be the roll-out collected using the $k$th particle from the model ensemble and the $m$th particle from the value ensemble. The goal of DiMVE is to empirically estimate the left $\alpha$-quantile of the posterior return distribution induced by the model ensemble for $\alpha \in (0, 1]$:
\begin{equation}
    \hat{y}_{DiMVE}(\alpha) = \inf\left\lbrace y \in \mathbb{R} : \mathbb{P}(\hat{y}(\state{}, \action{},\state{}') \leq y) > \alpha \right\rbrace.
    \label{eq:distributional-mve}
\end{equation}
Let $\Rhat{1} \leq \Rhat{2} \leq  \dots \Rhat{M\times K\times H}$ be the ordering of the $\Rhat{h}^{m,k}$, in the case where the samples are unique the DiMVE estimate can be written simply as
\begin{equation*}
    \hat{y}_{DiMVE}(\alpha) \approx \hat{R}_{\lfloor  \alpha \times M\times K\times H \rfloor}.
\end{equation*}

Table \ref{tab:distributional-mve} compares the performance of \acronym~and DiMVE for the \textit{walker2d} and \textit{halfcheetah} environments with the \textit{m}, \textit{mr}, \textit{me}, and \textit{fr} dataset configurations, where $\alpha$ was tuned among $\{0.4,~0.3085,~ 0.0228,~ 0.0013, ~2.87\times 10^{-7}\}$.  Here, the last four $\alpha$ values correspond to $\psi=0.5,2.0,3.0, 5.0$, respectively, if assuming the $\Rhat{h}^{m,k}$ samples are normally distributed.  We noted that $\alpha$ value smaller than $0.3085$ resulted in value divergence towards negative infinity, and so we report the performance with the best $\alpha$ values in Table \ref{tab:distributional-mve}.  Clearly, \acronym~outperforms the baseline in all tasks, showing the effectiveness of our Bayesian formulation. Furthermore, we found DiMVE to be more unstable during training and it consistently showed larger variances in the performance.
\begin{table}[t!]
\footnotesize
\caption{Comparison of \acronym~and DiMVE}
\centering
\begin{tabular}{lcccc}
\toprule
     Task name & \multicolumn{2}{c}{\acronym} & \multicolumn{2}{c}{DiMVE (best $\alpha$)} \\
\midrule
\midrule
    halfcheetah-m   & \multicolumn{2}{c}{$\boldsymbol{74.3}\pm 0.2$}  & \multicolumn{2}{c}{$70.9\pm 0.6$ ($0.3085$)}\\\Tstrut{}
    halfcheetah-mr  & \multicolumn{2}{c}{$\boldsymbol{66.4}\pm 0.3$}  & \multicolumn{2}{c}{$65.0\pm 0.3$ ($0.3085$)}\\\Tstrut{}
    halfcheetah-me  & \multicolumn{2}{c}{$\boldsymbol{100.4}\pm 0.9$}  & \multicolumn{2}{c}{$84.4\pm 6.6$ ($0.3085$)}\\\Tstrut{}
    halfcheetah-fr  & \multicolumn{2}{c}{$\boldsymbol{85.5}\pm 0.3$}  & \multicolumn{2}{c}{$83.4\pm 0.8$ ($0.4$)}\\
\midrule
    walker2d-m  & \multicolumn{2}{c}{$\boldsymbol{95.5}\pm 0.4$}  & \multicolumn{2}{c}{$65.1\pm 3.4$ ($0.3085$)}\\\Tstrut{}
    walker2d-mr & \multicolumn{2}{c}{$\boldsymbol{92.7}\pm 0.9$}  & \multicolumn{2}{c}{$88.5\pm 0.2$ ($0.3085$)}\\\Tstrut{}
    walker2d-me & \multicolumn{2}{c}{$\boldsymbol{117.2}\pm 0.5$}  & \multicolumn{2}{c}{$113.0\pm 9.8$ ($0.3085$)}\\\Tstrut{}
    walker2d-fr & \multicolumn{2}{c}{$\boldsymbol{107.8}\pm 0.2$}  & \multicolumn{2}{c}{$104.6\pm 1.0$ ($0.3085$)}\\
\bottomrule
\end{tabular}
\label{tab:distributional-mve}
\end{table}



%

\end{document}